\documentclass[preprint,12pt,authoryear]{elsarticle}
\usepackage{geometry}
\usepackage[utf8]{inputenc} % allow utf-8 input
\usepackage[T1]{fontenc}    % use 8-bit T1 fonts
\usepackage{amsmath, amssymb, bm}	% bm for bold math
\usepackage{hyperref}
\usepackage{textcomp} 				% for textdegree
\usepackage{caption, subcaption} 	% for multiple figures
\usepackage{xcolor}
\usepackage{siunitx}[=v2]
\usepackage{lineno}

\journal{Journal of Computational Physics}

\begin{document}

\begin{frontmatter}



\title{Multiple-Input Fourier Neural Operator (MIFNO) for source-dependent 3D elastodynamics}


\author[CEA,CS]{Fanny Lehmann}
\author[CS]{Filippo Gatti}
\author[CS]{Didier Clouteau}

\affiliation[CEA]{organization={CEA, DAM, DIF},%Department and Organization
            addressline={},
            city={Arpajon},
            postcode={F-91297}, 
            state={},
            country={France}}

\affiliation[CS]{organization={LMPS-Laboratoire de Mécanique Paris-Saclay, Université Paris-Saclay, ENS Paris-Saclay, CentraleSupélec, CNRS},%Department and Organization
	addressline={},
	city={Gif-sur-Yvette},
	postcode={91190}, 
	state={},
	country={France}}

\begin{abstract}
	Numerical simulations are essential tools to evaluate the solution of the wave equation in complex settings, such as three-dimensional (3D) domains with heterogeneous properties. However, their application is limited by high computational costs and existing surrogate models lack the flexibility of numerical solvers.
	
	This work introduces the Multiple-Input Fourier Neural Operator (MIFNO) to deal with structured 3D fields representing material properties as well as vectors describing the source characteristics. The MIFNO is applied to the problem of elastic wave propagation in the Earth's crust. It is trained on the HEMEW\textsuperscript{S}-3D database containing \SI{30 000}{} earthquake simulations in different heterogeneous domains with random source positions and orientations. Outputs are time- and space-dependent surface wavefields.
	
	The MIFNO predictions are assessed as good to excellent based on Goodness-Of-Fit (GOF) criteria. Wave arrival times and wave fronts propagation are very accurate since \SI{80}{\%} of the predictions have an excellent phase GOF (larger than \SI{8}{}). The fluctuations amplitudes are good for \SI{87}{\%} of the predictions (envelope GOF larger than \SI{6}{}) and excellent for \SI{28}{\%}. The envelope score is hindered by the small-scale fluctuations that are challenging to capture due to the spectral bias inherent to neural networks and the complex physical phenomena associated with high-frequency features. Nevertheless, the MIFNO can generalize to sources located outside the training domain and it shows good generalization ability to a real complex overthrust geology. When focusing on a region of interest, transfer learning is an efficient approach to improve the accuracy with limited additional costs, since GOF scores improved by more than \SI{1}{} GOF unit with only \SI{500}{} additional specific samples.
	
	The MIFNO is the first surrogate model offering the flexibility of an earthquake simulator with varying sources and material properties. Its good accuracy and massive speed-up offer new perspectives to replace numerical simulations in many-query problems. 

\end{abstract}

\begin{keyword}
	elastic wave equation \sep computational mechanics \sep seismology \sep neural operator \sep artificial intelligence



\end{keyword}

\end{frontmatter}

%\linenumbers

\section{Introduction}
	Predicting the intensity of ground motion is a crucial challenge for earthquake early warning and seismic hazard analyses. For early warning, one should be able to estimate ground motion before seismic waves reach the area distant from the epicenter \citep{allenEarthquakeEarlyWarning2019, cremenEarthquakeEarlyWarning2020}; and seismic hazard analyses require repeated evaluations of ground motion in varying configurations \citep{bakerProbabilisticSeismicHazard2013}. Therefore, fast estimations of ground shaking intensity are key for both tasks. To this end, recorded seismograms as well as numerical simulations have been used since they provide complimentary information.
	
	On the one hand, seismograms recorded at different stations for the same earthquake provide means to predict the relationship between earthquake parameters and ground motion intensity. This type of analysis relies on the existence of earthquake catalogs that are scrutinised with machine learning. \cite{mousaviMachineLearningEarthquake2023} reviews machine learning methods such as random forests, support vector machines, artificial neural networks, etc. that have been used to design non-parametric ground motion models. The development of deep learning is also leading to early warning models able to predict ground motion intensity from the first seconds of seismograms recorded close to the epicenter, using Convolutional Neural Networks (CNN, \cite{hsuOnsiteEarlyPrediction2021, jozinovicRapidPredictionEarthquake2020}), Graph Neural Networks (GNN, \cite{bloemheuvelGraphNeuralNetworks2023}), or Recurrent Neural Networks (RNN, \cite{dattaDeepShakeShakingIntensity2022, wangPredictionPGAEarthquake2023}). One main limitation in these approaches is their need for a large number of recordings in the area under study. In regions with poor data coverage or low-to-moderate seismicity, these purely data-driven methods fall short.
	
	On the other hand, numerical simulations provide ground motion time series in any seismological context. They allow a great flexibility since parameters describing geological properties, source characteristics, stations location, etc. can be specified on demand. Physics-based simulations should be conducted in three-dimensional (3D) domains to correctly model the propagation of waves beyond the too simplistic two-dimensional (2D) setting \citep{moczoKeyStructuralParameters2018, smerziniComparison3D2D2011, zhuSeismicAggravationShallow2020}. In the meantime, numerical results should be accurate up to frequencies of a few Hz to allow meaningful engineering applications \citep{bradleyGuidanceUtilizationEarthquakeInduced2017}. Independently from the numerical method used to solve the wave equation (e.g. finite differences, finite elements, spectral elements, among others), these constraints lead to high computational costs that can become a limiting factor when quantifying uncertainties. Indeed, the computed ground motion strongly depends on the simulation parameters, such as the regional geological properties and the source characteristics. Due to the difficulty of conducting geophysical measurements and accurately inverting the source properties, these parameters are associated with large uncertainties. Therefore, with a Monte-Carlo approach, many simulations should be conducted to assess the influence of parameters uncertainties on ground motion response. The computational cost of high-fidelity simulations prohibits this approach. 
	
	With the joint development of high-performance computing and deep learning, there is a growing interest in developing surrogate models able to mimic numerical simulations. The aim is to predict space-dependent ground motion time series that give much more detailed information than ground motion models restricted to some intensity measures. Thus, neural operators \citep{liFourierNeuralOperator2021, luLearningNonlinearOperators2021, brandstetterMessagePassingNeural2022} are a suitable framework to provide solutions of the wave propagation equation from a set of input parameters describing the geological configuration (velocity of P and S waves, soil density, attenuation factors) and the source (epicenter location, source orientation, magnitude).
	
	Neural operators were shown to be efficient surrogate models for the elastic wave equation in 2D \citep{obrienImagingSeismicModelling2023, yangRapidSeismicWaveform2023, zhangLearningSolveElastic2023} and 3D \citep{kongFeasibilityUsingFourier2023, lehmannFourierNeuralOperator2023, lehmann3DElasticWave2024, zouDeepNeuralHelmholtz2023}. These works use variants of the Fourier Neural Operator \citep{liFourierNeuralOperator2021} to learn the propagation of seismic waves by the means of their frequency representation. Some studies consider both the source location and the geological parameters as input variables \citep{kongFeasibilityUsingFourier2023, obrienImagingSeismicModelling2023, yangRapidSeismicWaveform2023, zouDeepNeuralHelmholtz2023}. Yet, these works do not take into account the different nature of the geological and source parameters: a structured 2D or 3D field for the geology and a vector of coordinates for the source. The source position is encoded as a binary field where a single pixel of value 1 describes the source location. 
	
	In addition, an earthquake source cannot be described solely by the source location since its orientation also plays a crucial role, through the definition of the moment tensor. With the notable exception of \cite{obrienImagingSeismicModelling2023} that restricts their analyses to 2D and mention "significant challenges" to extend their methodology to 3D, the above-mentioned works consider an isotropic explosive source. In this situation, the moment tensor is fixed and does not need to be learnt by the surrogate model. But it prevents earthquake modeling that requires a moment tensor description of the source. The current study proposes an alternative architecture to include both the source location and moment tensor by considering their vector representation.
	
	Some neural operators have been designed specifically to account for multiple inputs, possibly with different structures. The General Neural Operator Transformer (GNOT, \cite{haoGNOTGeneralNeural2023}) uses different Multi-Layer Perceptron (MLP) blocks to encode each input before transmitting it to a transformer network. Other models were inspired by the Deep Operator Network (DeepONet, \cite{luLearningNonlinearOperators2021}) by implementing one specific \textit{branch} network for each input (MIONet, \cite{jinMIONetLearningMultipleinput2022}, Fourier-MIONet  \citep{jiangFourierMIONetFourierenhancedMultipleinput2023}). However, the complexity of these models makes 3D applications challenging.
	
	In this study, we propose a Multiple Input Fourier Neural Operator (MIFNO) that takes as inputs the 3D geological properties, the source position and the source orientation, and predicts the surface ground motion as the solution of the 3D elastic wave equation. The inherent complexity of 3D variables is handled with a Factorized Fourier Neural Operator (F-FNO, \cite{tranFactorizedFourierNeural2023}). The MIFNO involves a convolutional branch transforming information from the source into structured fields that get through the factorized Fourier layers. The MIFNO is trained on the HEMEW\textsuperscript{S}-3D database containing \SI{30 000}{} simulations of elastic wave propagation in 3D heterogeneous geologies with randomly located and oriented sources \citep{lehmannSyntheticGroundMotions2024}.
	
	In the following, Section \ref{sec:data_methods} describes the data and the neural operators used for training. Section \ref{sec:results} analyses the predictions, discusses the evaluation metrics, and the influence of inputs on the outcomes. Section \ref{sec:generalization} explores generalization to out-of-distribution data and Section \ref{sec:transfer_learning} extends it with transfer learning to a real earthquake. Finally, Section \ref{sec:conclusion} draws conclusions.

\section{Data and methods}
\label{sec:data_methods}
\subsection{Problem setting}
	We consider a cubic domain $\Omega= [0; \Lambda] \subset \mathbb{R}^3$ of size \SI{9.6 x 9.6 x 9.6}{km}. Seismic waves propagate inside this domain from the source up to the upper surface denoted $\partial \Omega_{top}$. Except the traction-free upper surface $\partial \Omega_{top}$, all other external surfaces have absorbing conditions to mimic a semi-infinite propagation domain. 
	
	The propagation domain is described by geological parameters (velocity of P and S waves, soil density, attenuation factors). For the sake of simplicity, we assume that the geology is entirely defined by the S-wave velocity $V_S$ and other parameters can be computed from deterministic relationships involving only $V_S$ (in particular, the ratio $V_P/V_S$ is fixed to \SI{1.7}{}). In the general setting, the propagation domain is characterized by a function $a: \Omega \to \mathbb{R}$ that gives the value of $V_S$ at each spatial point $\bm{x} \in \Omega$. 
	
	The propagation of waves inside the domain obeys the elastic wave equation \begin{equation}
	\rho \dfrac{\partial^2 \bm{u}}{\partial t^2} = \nabla \lambda \left( \nabla \cdot \bm{u} \right) 
	+ \nabla \mu \left[ \nabla \bm{u} + \left( \nabla \bm{u} \right)^T \right] 
	+ \left( \lambda + 2 \mu \right) \nabla \left( \nabla \cdot \bm{u} \right)
	- \mu \nabla \times \nabla \times \bm{u} + \bm{f}
	\label{eq:elastic_equation}
	\end{equation}
	where $\rho : \Omega \to \mathbb{R}$ is the material unit mass density, $\lambda : \Omega \to \mathbb{R}$, $\mu : \Omega \to \mathbb{R}$ are the Lamé parameters (all contained in our general function $a$), characterizing the thermodynamically reversible mechanical behaviour of the material, $\bm{f}$ is the body force distribution, and $\bm{u}~:~\Omega~\times~[0, T]~\to~\mathbb{R}^3$ is the displacement field. The forcing term $\bm{f}(\bm{x}, t)~=~\mathrm{div}~\:~\bm{m}(\bm{x})~\cdot~s(t)$ is the divergence of a moment tensor density $\bm{m}$, localized at a point-wise location $\bm{x_s}$. In this study, the moment tensor is given as a vector $\bm{\theta_s}$ and the source position is denoted by the vector $\bm{x_s}$. 
	Then, equation \ref{eq:elastic_equation} can be rewritten under the general form 
	\begin{equation}
	\mathcal{L}(a, \bm{u}) = \bm{f}(\bm{x_s}, \bm{\theta_s})
	\label{eq:general_equation}
	\end{equation}
	
	\begin{figure}[t]
		\centering		
		\includegraphics[width=\textwidth]{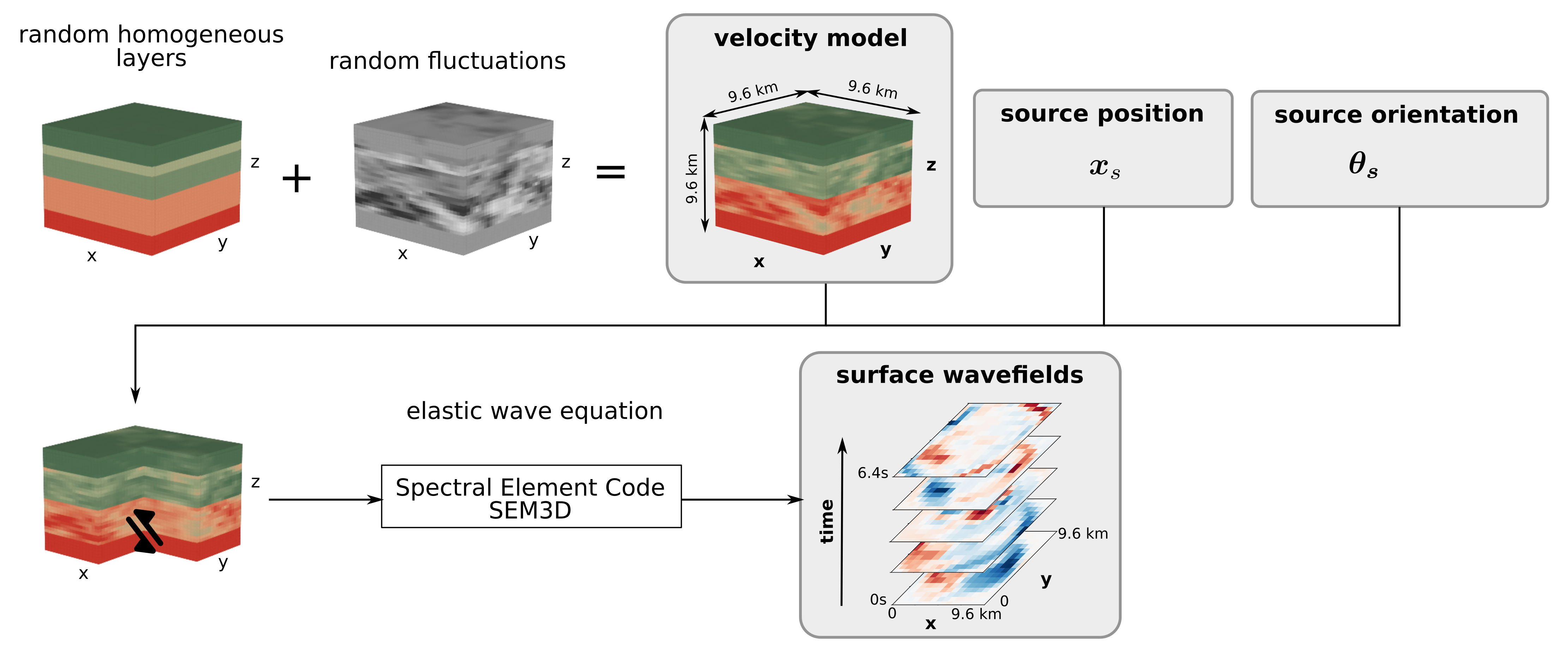}
		\caption{Composition of the HEMEW\textsuperscript{S}-3D database. Velocity models were built from the addition of randomly chosen horizontal layers and heterogeneities drawn from random fields. Combined with the source position and the source orientation, they form the inputs of the neural operator. Outputs of the Spectral Element Code SEM3D are velocity wavefields synthetized at the surface of the domain by a grid of virtual sensors.}
		\label{fig:HEMEW3D}
	\end{figure}
	
	We aim at a surrogate model $G_{\phi}(a, \bm{x_s}, \bm{\theta_s})$ predicting the velocity field $\bm{\dot{u}}$ (obtained as the time derivative of the displacement $\bm{u}$) from the sole knowledge of the geology $a$ and the source characteristics $\bm{x_s}, \bm{\theta_s}$. Reference velocity fields are obtained with physics-based numerical simulations. To reduce the memory requirements, simulation outputs are synthetized only at the surface of the propagation domain $\partial \Omega_{top}$. For the sake of clarity, $\bm{u}$ denotes surface velocity fields $\bm{\dot{u}}_{|\partial \Omega_{top}}$ in the following. The surrogate model is trained to minimize the error between the predicted surface velocity fields $\bm{\hat{u}}$ and the simulated ones $\bm{u}$.

\subsection{Data}
\label{sec:data_HEMEWS}
	Each data sample is composed of four elements: the description $a$ of S-wave velocity inside $\Omega$, the source position $\bm{x_s}$, the source orientation $\bm{\theta_s}$, and the simulated surface velocity field $\bm{u}$.
	
	S-wave velocity models are designed as random \textit{horizontal} models augmented with random heterogeneities. The \textit{horizontal} models are built as a stack of horizontal layers with random thickness and random value. They describe the baseline geophysical knowledge of the domain. Then, log-normal random fluctuations are added independently inside each layer to represent geological heterogeneities (Fig. \ref{fig:HEMEW3D}). These random fields are stationary and characterized by their coefficient of variation $\sigma$ following a folded normal distribution $|\mathcal{N}(0.1, 0.2)|$, their correlation length $\ell_c$ randomly chosen in $\{1.5, 3, 4.5, 6 \: km\}$, and their power spectral density or correlation function. After the addition of random heterogeneities, S-wave velocities are clipped between V\textsubscript{S,min}~=~\SI{1071}{m/s} and V\textsubscript{S,max}~=~\SI{4500}{m/s}. Table \ref{tab:stats_materials} summarizes the velocity models parameters and a more detailed description of the velocity models can be found in \citep{lehmannSyntheticGroundMotions2024}. More advanced random fields could be defined for practical applications accounting for V\textsubscript{P}, V\textsubscript{S}, $\rho$ variability together with controlled anisotropy \citep{taModelingRandomAnisotropic2010}. However, it is believed that the proposed model spreads a large enough set to demonstrate the applicability of our surrogate model to other probabilistic models of the geological formation. 
	
	\begin{table}[h]
		\begin{center}
			\begin{tabular}{|c|c|}
				\hline 
				Parameter & Statistical distribution \\ 
				\hline 
				Number of heterogeneous layers $N_{\ell}$ & $\mathcal{U}(\{1,2,3,4,5,6\})$ \\
				Layers thickness $h_1, \cdots, h_{N_{\ell}}$ & $\mathcal{U} \left( \{(h_1, \cdots, h_{N_{\ell}}) > 0 | h_1 + \cdots + h_{N_{\ell}} = 7.8 \ \text{km} \} \right)$ \\
				Mean $V_S$ value per layer & $\mathcal{U}([1785, 3214 \ \text{m/s}])$ \\
				Layer-wise coefficient of variation & $\left| \mathcal{N}(0.2, 0.1) \right|$ \\
				Layer-wise correlation length along x & $\mathcal{U}(\{ 1.5, 3, 4.5, 6 \ \text{km}\})$ \\
				Layer-wise correlation length along y & $\mathcal{U}(\{ 1.5, 3, 4.5, 6 \ \text{km}\})$ \\
				Layer-wise correlation length along z & $\mathcal{U}(\{ 1.5, 3, 4.5, 6 \ \text{km}\})$ \\
				\hline
			\end{tabular}
		\end{center}
		\caption{Statistical distribution of each parameter describing the velocity models $a$. Mean $V_S$ values, coefficients of variation, and correlation lengths are chosen independently in each layer. Since the bottom layer has a constant thickness of \SI{1.8}{km} and value V\textsubscript{S}~=\SI{4500}{m/s}, it is not included in these parameters.}
		\label{tab:stats_materials}
	\end{table}
	
	The source position $\bm{x_s} = (x_s, y_s, z_s)$ is chosen from a Latin Hypercube sampling inside the propagation domain, not too close from the boundaries to avoid numerical issues due to absorbing boundary conditions: \begin{align*}
	x_s & \in [1.2; 8.4 \: km] \\ 
	y_s & \in [1.2; 8.4 \: km] \\
	z_s & \in [-9.0; -0.6 \: km]
	\end{align*}
	
	The moment tensor is a $3 \times 3$ symmetric matrix that represents the source of a seismic event. When the seismic event is an earthquake, the moment tensor can be derived from three angles (strike, dip, and rake) that characterize the seismic rupture. The angles were sampled from a Latin Hypercube with a strike between \ang{0} and \ang{360}, dip between \ang{0} and \ang{90}, and rake between \ang{0} and \ang{360}. This constitutes the first representation of the source orientation, $\bm{\theta_s} \in \mathbb{R}^3$. Alternatively, the equivalent moment tensor was computed for each set of angles. Then, the source orientation is described by the six components of the symmetric moment tensor $\bm{\theta_s} \in \mathbb{R}^6$. The MIFNO can be trained with any of these representations. 
	The source amplitude corresponds to a seismic moment $M_0$~=~ \SI{2.47e16}{N.m} and the source time evolution is given by $t \mapsto 1 - \left(1 + \frac{t}{\tau} \right) e^{-\frac{t}{\tau}}$ with $\tau$~=~\SI{0.1}{s}.
	
	Numerical simulations are conducted with the SEM3D code\footnote{\url{https://github.com/sem3d/SEM}} \citep{touhamiSEM3D3DHighFidelity2022} based on the Spectral Element Method \citep{faccioli2d3DElastic1997,komatitschIntroductionSpectralElement1999}. The simulation domain is discretized in elements of size \SI{300}{m} and 7 Gauss-Lobatto-Legendre quadrature points per side of mesh element allow the propagation of waves up to a \SI{5}{Hz} frequency. Velocity fields $\bm{u}$ are synthetized on a regular grid of \SI{32 x 32}{} virtual sensors located at the surface of the domain, between \SI{0.15}{km} and \SI{9.45}{km}. A time window of [0s; 6.4s] is selected as it contains the significant ground motion and the time sampling is \SI{0.02}{s}. It should be noted that the velocity field at each sensor contains three components, denoted as E-W (East-West), N-S (North-South), and Z (vertical). 
	
	It is important to emphasize that velocity fields are seen as 3D variables depending on the coordinate $(x,y)$ of the surface sensor and time $t$. Therefore, both the geological parameters given as inputs (function of $(x,y,z)$) and the output surface wavefields (function of $(x,y,t)$) are 3D variables, with a depth-to-time conversion happening in the transformation. 
	
	Predicting the space- and time-dependent solution of the 3D elastic wave equation is a complex high-dimensional problem. The complexity can be assessed with two notions of intrinsic dimension. The Principal Component Analysis (PCA) dimension describes the number of linear components required to capture \SI{95}{\%} of the variability in the ground motion wavefields database. The intrinsic dimension can also be estimated from the Maximum Likelihood Estimation (MLE) of the size of the wavefields' neighbourhood \cite{levinaMaximumLikelihoodEstimation2004}. The intrinsic dimension is at least of the order of \SI{100}{} when estimated with the non-linear MLE and it reaches nearly \SI{5000}{} with the linear PCA (Fig. \ref{fig:intrinsic_dimension}). Although it is impossible to conclude on the exact dimensionality of the ground motion wavefields, these results emphasize the complexity of the problem.

\subsection{Factorized Fourier Neural Operator (F-FNO)}
\label{sec:FFNO}
	The Fourier Neural Operator (FNO) was proposed by \cite{liFourierNeuralOperator2021} to extend neural networks to functional spaces. It is especially convenient to solve problems related to Partial Differential Equations (PDEs) since it can be interpreted as a means to learn the frequency representation of the PDE solution. The FNO is based on Fourier layers that involve computing the Fast Fourier Transform (FFT) of the model internal variables. Although only the first Fourier modes are preserved to limit the model complexity, the FFT yields a large number of model parameters when dealing with 3D variables. This increases the generalization gap of  the FNO \citep{lehmann3DElasticWave2024} and hinders its accuracy.
	
	\begin{figure}[h]
		\centering
		\includegraphics[width=0.8\textwidth]{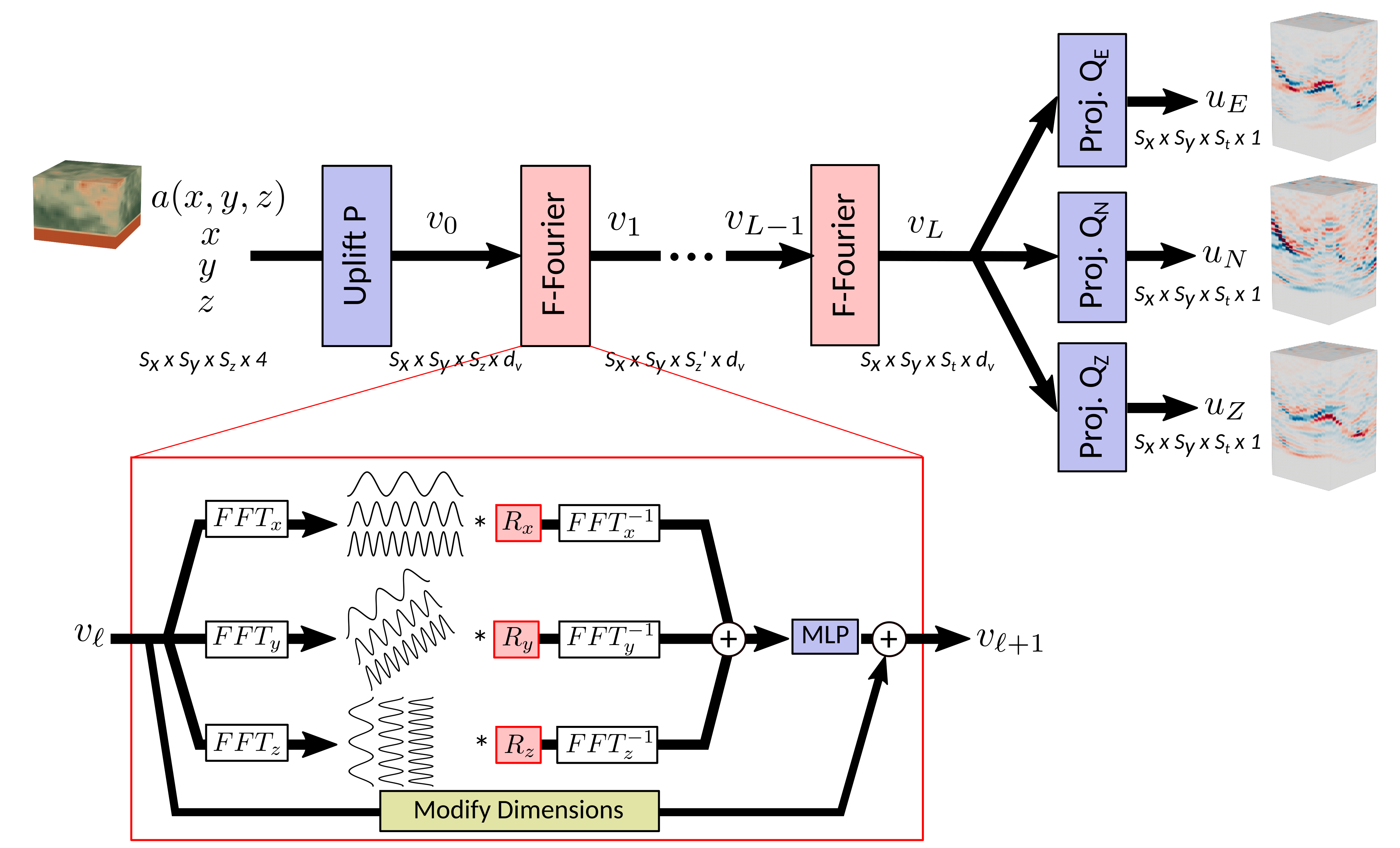}
		\caption{The F-FNO is made of one uplift block $P$, a succession of $L$ factorized Fourier layers, and three projection blocks $Q_E$, $Q_N$, $Q_Z$. The details of a factorized Fourier layer are given to show the decomposition of the FFT along each dimension.}
		\label{fig:FFNO_archi}
	\end{figure}
	
	Therefore, a more efficient approach was introduced with the Factorized Fourier Neural Operator (F-FNO, \cite{tranFactorizedFourierNeural2023}). The general architecture of the F-FNO is similar to the FNO and is illustrated in Fig. \ref{fig:FFNO_archi}. Inputs consist of the concatenation of four 3D variables: the velocity model $a$ and the grid of $x$, $y$, and $z$ coordinates. The concatenation is performed along the fourth dimension representing the channels. Then, a shallow linear block $P$ uplifts the 4-channel input variable into a variable $\bm{v}_0$ with $d_v$ channels ($d_v$=16 in this work). $P$ acts point-wise on the inputs and allows to locate the geological parameters at their exact location. Then, a succession of $L$ factorized Fourier layers is applied to transform $\bm{v}_0$ into $\bm{v}_L$. At the end of the F-FNO, three shallow linear blocks $Q_E$, $Q_N$, $Q_Z$ project $\bm{v}_L$ onto each of the velocity component $u_E$, $u_N$, $u_Z$. 
	
	In the F-FNO, the factorized Fourier layers transform the internal variable $\bm{v}_{\ell}$ into $\bm{v}_{\ell+1}$ following 
	\begin{equation}
	\begin{split}
	\bm{v}_{\ell+1} & = \bm{v}_{\ell} +  MLP \left(\mathcal{K}_{\ell}(\bm{v}_{\ell}) \right) \\ 
	MLP(\bm{k}) & = W_{\ell}^2 \: \sigma \left( W_{\ell}^1 \: \bm{k} + b_{\ell}^1 \right) + b_{\ell}^2
	\end{split}
	\label{eq:factorized_Fourier_layer}
	\end{equation}
	where $\sigma$ is the activation function, $W_{\ell}^1$ and $W_{\ell}^2$ are bias matrices acting point-wise, $b_{\ell}^1$ and $b_{\ell}^2$ are scalar biases, and $\mathcal{K}_{\ell}$ is a neural network playing the role of the integral operator. The majority of the F-FNO weights are contained in the $\mathcal{K}_{\ell}$ network. The core idea of this network \citep{liFourierNeuralOperator2021} stems from the convolution theorem that allows one to write an integral operator $\mathcal{K}$ depending on the stationary kernel $\kappa_{\psi}$ (where $\psi$ denotes the learnable weights)
	\begin{equation}
	\mathcal{K}(\bm{v}) = \int \kappa_{\psi}(\bm{x}, \bm{y}) \bm{v}(\bm{y}) d\bm{y}
	\end{equation}
	as the product 
	\begin{equation}
	\mathcal{K}(\bm{v}) = \mathcal{F}^{-1} \left( \mathbf{R}_{\psi} \cdot \mathcal{F}(\bm{v}) \right)
	\label{eq:FFT_convolution}
	\end{equation}
	In equation \ref{eq:FFT_convolution}, $\mathcal{F}$ denotes the Fourier transform and $\mathbf{R}_{\psi} = \mathcal{F}(\kappa_{\psi})$ is the tensor of learnable complex weights. In comparison with the original FNO formulation, the kernel $\mathcal{K}_{\ell}$ of the factorized Fourier layer replaces the 3D Fourier transform $\mathcal{F}$ by the sum of three 1D Fourier transforms along each dimension ($\mathcal{F}_{(1)}$, $\mathcal{F}_{(2)}$, $\mathcal{F}_{(3)}$):
	\begin{equation}
	\mathcal{K}_{\ell}(\bm{v}_{\ell}) = \mathcal{F}^{-1}_{(1)} \left( R_{(1),\ell} \mathcal{F}_{(1)} (\bm{v}_{\ell}) \right) + 
	\mathcal{F}^{-1}_{(2)} \left( R_{(2),\ell} \mathcal{F}_{(2)} (\bm{v}_{\ell}) \right) + 
	\mathcal{F}^{-1}_{(3)} \left( R_{(3),\ell} \mathcal{F}_{(3)} (\bm{v}_{\ell}) \right)
	\label{eq:factorized_kernel}
	\end{equation}
	In equation \ref{eq:factorized_kernel}, $R_{(1),\ell}$, $R_{(2),\ell}$, and $R_{(3),\ell}$ are the neural operator's weights that are optimized during training. The Fourier transform is computed efficiently with the Fast Fourier Transform (FFT) and only the first Fourier modes are selected in each dimension. In this work, 16 Fourier modes are selected in the $x$ and $y$ dimensions ($M_{(1),\ell} = M_{(2), \ell} = 16$, $\forall \ell$) and 32 Fourier modes are selected in the third dimension except for the first layer ($M_{(3), \ell} = 32$, $\forall \ell \ge 2$, $M_{(3),1} = 16$).

	It should also be noted that inputs and outputs have different dimensions in our work. Indeed, inputs are velocity models depending on the three spatial variables $(x,y,z)$, their size is $S_x \times S_y \times S_z$. While outputs are time-dependent velocity fields recorded only at the surface of the propagation domain. Therefore, they depend on the spatial location of the sensor $(x,y)$ and time $t$. The size of outputs is then $S_x \times S_y \times S_t$. Dimensions are chosen as $S_x=S_y=S_z=32$ and $S_t=320$. The third dimension increase is done gradually inside the Fourier layers by padding Fourier coefficients with zeros before applying the inverse Fourier transform \citep{rahmanUshapedNeuralOperators2023}. The same procedure is applied before the summation to match the dimensions (block \verb|ModifyDimensions| in Fig. \ref{fig:FFNO_archi}).
	
	Works like \citep{kongFeasibilityUsingFourier2023, zouDeepNeuralHelmholtz2023} add the source position as an additional input to the Neural Operator. This is done by inserting a fifth variable alongside $a$ and the grids $x$, $y$, $z$. This variable is a cube full of zeros with a single pixel of 1 denoting the source position. In this work, we propose an alternative architecture that takes advantage of the vector representation of the source to include more source parameters. It is worth noticing that the elastodynamic problem (Eq. \ref{eq:elastic_equation}) being linear with respect to the force term $f$, extended sources such as an extended seismic fault can be easily addressed using the superposition principle.

\subsection{Multiple-Input Fourier Neural Operator (MIFNO)}
\label{sec:MIFNO}
\begin{figure}[h]
		\centering
		\includegraphics[width=\textwidth]{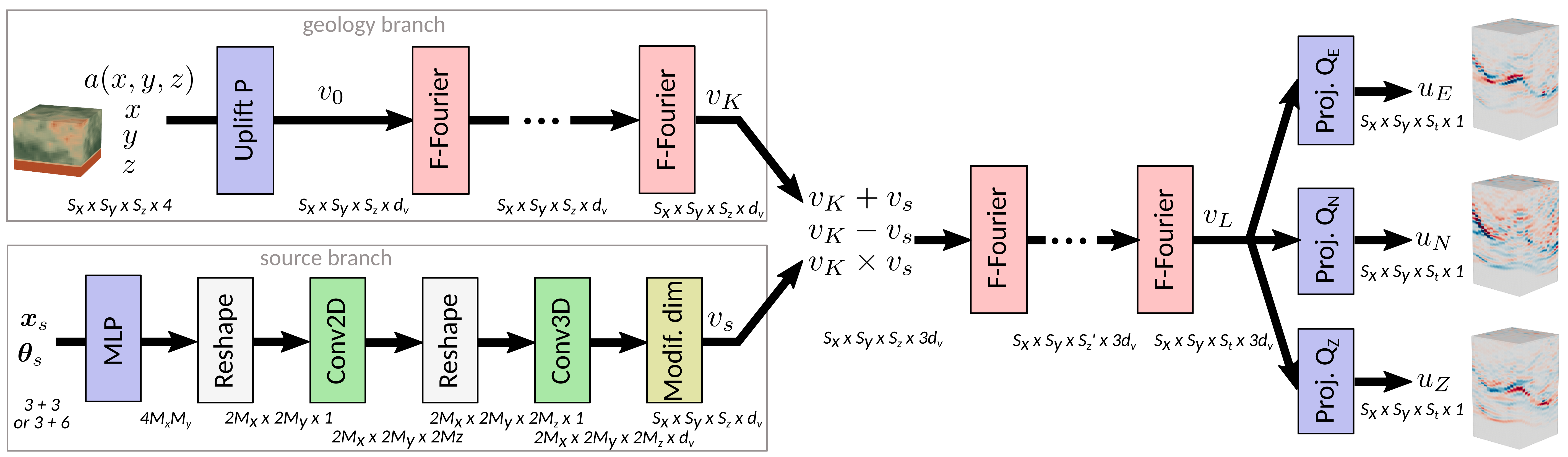}
		\caption{The MIFNO is made of a \textit{geology branch} that encodes the geology with factorized Fourier (F-Fourier) layers, and a \textit{source branch} that transforms the vector of source parameters $(\bm{x}_s, \bm{\theta}_s)$ into a 4D variable $v_S$ matching the dimensions of the \textit{geology branch} output $v_K$. Outputs of each branch are concatenated after elementary mathematical operations and the remaining factorized Fourier layers are applied. Uplift $P$ and projection $Q_E$, $Q_N$, $Q_Z$ blocks are the same as in the F-FNO.}
		\label{fig:MIFNO_archi}
	\end{figure}

	Inspired by the MIONet architecture that designs separate branches for each input \citep{jinMIONetLearningMultipleinput2022}, we propose a Neural Operator architecture that encodes the source parameters in a specific branch. Our Multiple-Input Fourier Neural Operator (MIFNO) is illustrated in Fig. \ref{fig:MIFNO_archi}. 
	
	In our MIFNO, the succession of factorized Fourier layers is split in two parts. From layers $1$ to $K$, factorized Fourier layers act on the geology as in the original F-FNO. This forms the \textit{geology branch}. In parallel, the \textit{source branch} takes as input the vector of source coordinates $\bm{x_s}$ concatenated with the source characteristics $\bm{\theta_s}$. Even though resolution invariance requires additional conditions \citep{bartolucciAreNeuralOperators2023}, FNOs can always be technically applied to inputs having a different resolution than the resolution used during training. This property must be preserved in the \textit{source branch}, which is not trivial for convolutional layers whose number of weights depend on the size of the inputs. To circumvent this issue, the layers in the \textit{source branch} are defined based on the number of modes used in the $K$-th factorized Fourier layer, denoted $M_x$, $M_y$, $M_z$, and not the actual size of the inputs.
	
	Several architectures were investigated to design the \textit{source branch} (not shown) and we found that the best one was made of a perceptron and two convolutional layers separated by reshaping operations. Our interpretation of the source branch is the following: 
	\begin{enumerate}
		\item a 2-layer perceptron encodes the source characteristics in the $(x,y)$ plane. Its first layer contains 128 neurons and its second layer $4M_xM_y$ neurons. The reshaping operation leads to a 3D variable of size $2M_x \times 2M_y \times 1$.
		\item a 2D CNN (with kernel size of 3) creates the third dimension. A first convolutional layer creates 8 channels and a second layer creates $2M_z$ channels. After reshaping, one gets a 4D variable of size $2M_x \times 2M_y \times 2M_z \times 1$.
		\item a 3D CNN (kernel size of 3) adds the channel dimension with two convolutional layers.
	\end{enumerate}
	At the end of the source branch, the \verb|ModifyDimensions| layer transforms variables into the Fourier space where Fourier coefficients are padded with 0 if necessary, and then inverse transformed to obtain the same dimension as the output of the \textit{geology branch} $v_K$, i.e. $S_x \times S_y \times S_z \times d_v$.
	
	To combine variables from the \textit{geology} and \textit{source branches}, we take inspiration from \cite{haghighatEnDeepONetEnrichmentApproach2024} who propose a 2D eXtended-DeepONet for earthquake localization. Following this idea, $v_K$ and $v_s$ are summed, substracted, and multiplied before being concatenated along the channel dimension. This leads to a variable of size $S_x~\times~S_y~\times~S_z~\times~3d_v$ that is then sent to the remaining factorized Fourier layers, $K+1$ to $L$. The final projection layers $Q_E$, $Q_N$, $Q_Z$ are identical to the F-FNO.
	
	It can also be noted that the MIFNO is exemplified with factorized Fourier layers but the architecture can be extended to other types of layers. The MIFNO used in this work contains 16 factorized Fourier layers in total ($L=16$), including 4 layers in the \textit{geology branch} ($K=4$). The number of channels is fixed to $d_v=16$. Its total number of parameters amount to \SI{3.40}{} million. The initial learning rate is $0.0004$ and is halved on plateau, and the loss function is the relative Mean Absolute Error. The choice of hyperparameters and training strategies comes from extensive investigations conducted for the F-FNO in \cite{lehmann3DElasticWave2024}. \SI{27 000}{} samples were used for training, \SI{3 000}{} for validation, and \SI{1000} for testing. Training was performed on 4 Nvidia A100 GPUs for 200 epochs, which took \SI{29.6}{h}.
	
	Some care should be taken to normalize the inputs and outputs since their variability ranges are significantly different. Geological models are normalized to a Gaussian distribution, by centering them with respect to the mean geology and normalizing by four times the geological standard deviation to yield values approximately in $[-0.5, 0.5]$. The source coordinates and the source angles are mapped to the $[0, 1]^3$ cube. Velocity wavefields predicted by the MIFNO are normalized with a scalar (i.e. independent of the sensor coordinates) that depends only on parameters known beforehand: the S-wave velocity at the source location $V_S(\bm{x}_s)$ and the source depth $z_s$. The normalization value is $c = V_s(\bm{x}_s) \sqrt{z_s^2 + (\Lambda/4)^2}$ where $\Lambda$ is the length of the domain. The first term relates to the seismic energy released by an earthquake, which is inversely proportional to the shear modulus (hence, to the S-wave velocity) at the rupture location. The second term accounts for the amplitude decrease of seismic waves generated by deep sources since they undergo more geometrical dispersion and diffraction.

\section{Results}
\label{sec:results}	
	When comparing the surrogate model predictions with the reference numerical simulations, one needs to use appropriate metrics that explain the physical meaning of the misfit. In particular, the usual Mean Absolute Error (MAE) and Root Mean Square Error (RMSE) have limited interpretability to analyze time series predictions. In seismology, time series are often compared in terms of time-frequency envelope and phase misfits \cite{kristekovaTimefrequencyMisfitGoodnessoffit2009}. Misfits can be summarized by two scalar values: the envelope Goodness-Of-Fit (GOF) and the phase GOF. The envelope GOF expresses the error of the predicted amplitude compared to the reference time series while the phase GOF describes the error of the phase arrivals. GOFs are given on a scale between 0 and 10 where 10 means a perfect agreement and the following score assessment is well accepted: 0-4 is a poor score, 4-6 is fair, 6-8 is good, 8-10 is excellent \citep{kristekovaTimefrequencyMisfitGoodnessoffit2009}.

\subsection{Predictions analyses}
\begin{figure}[h]
		\centering
		\begin{subfigure}[t]{0.02\textwidth}
			(a)
		\end{subfigure}
		\hfill
		\begin{subfigure}[t]{0.4\textwidth}
			\includegraphics[width=\textwidth]{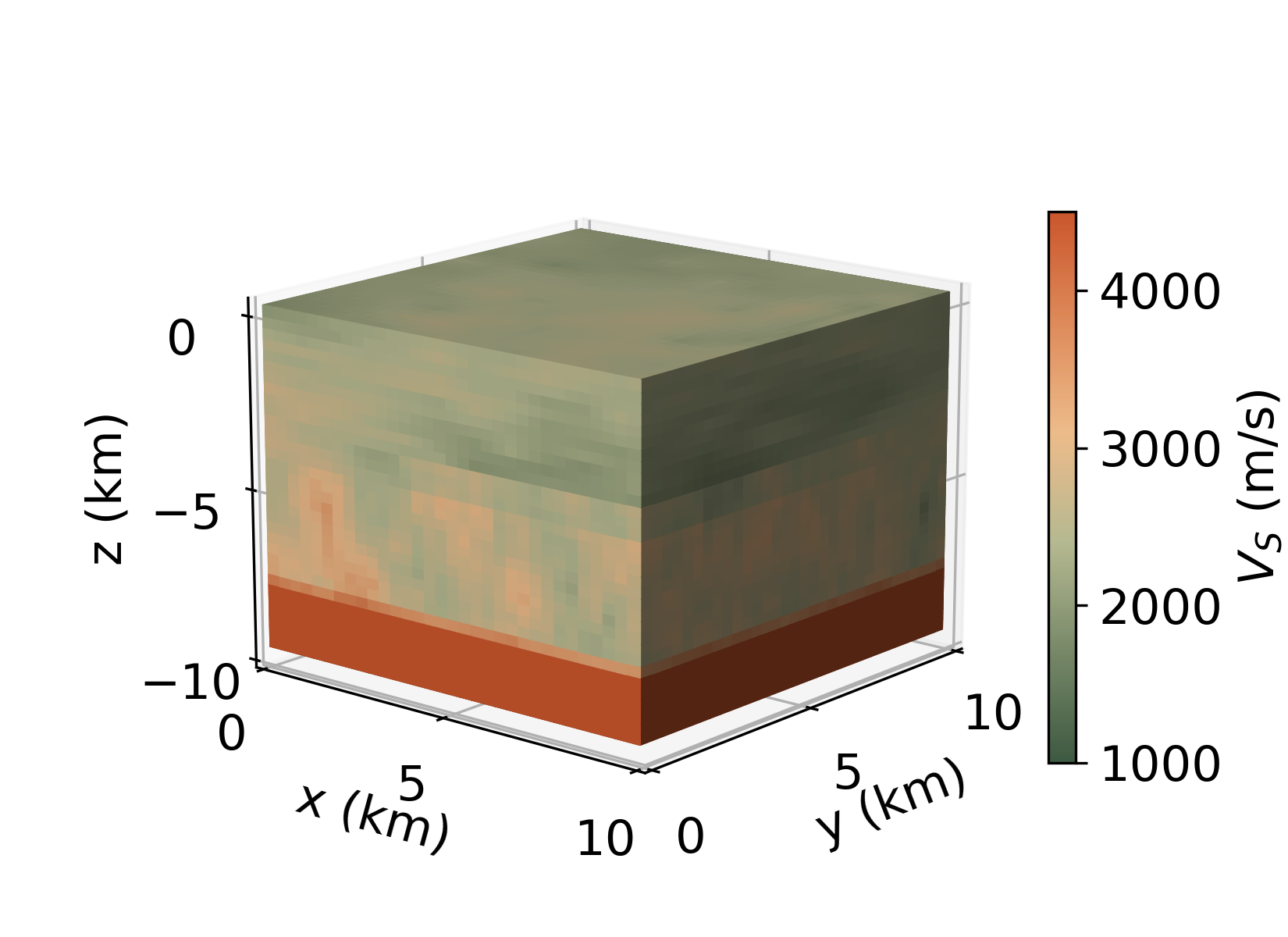}
		\end{subfigure}
		\hfill
		\begin{subfigure}[t]{0.02\textwidth}
			(b)
		\end{subfigure}
		\hfill
		\begin{subfigure}[t]{0.5\textwidth}
			\includegraphics[width=\textwidth]{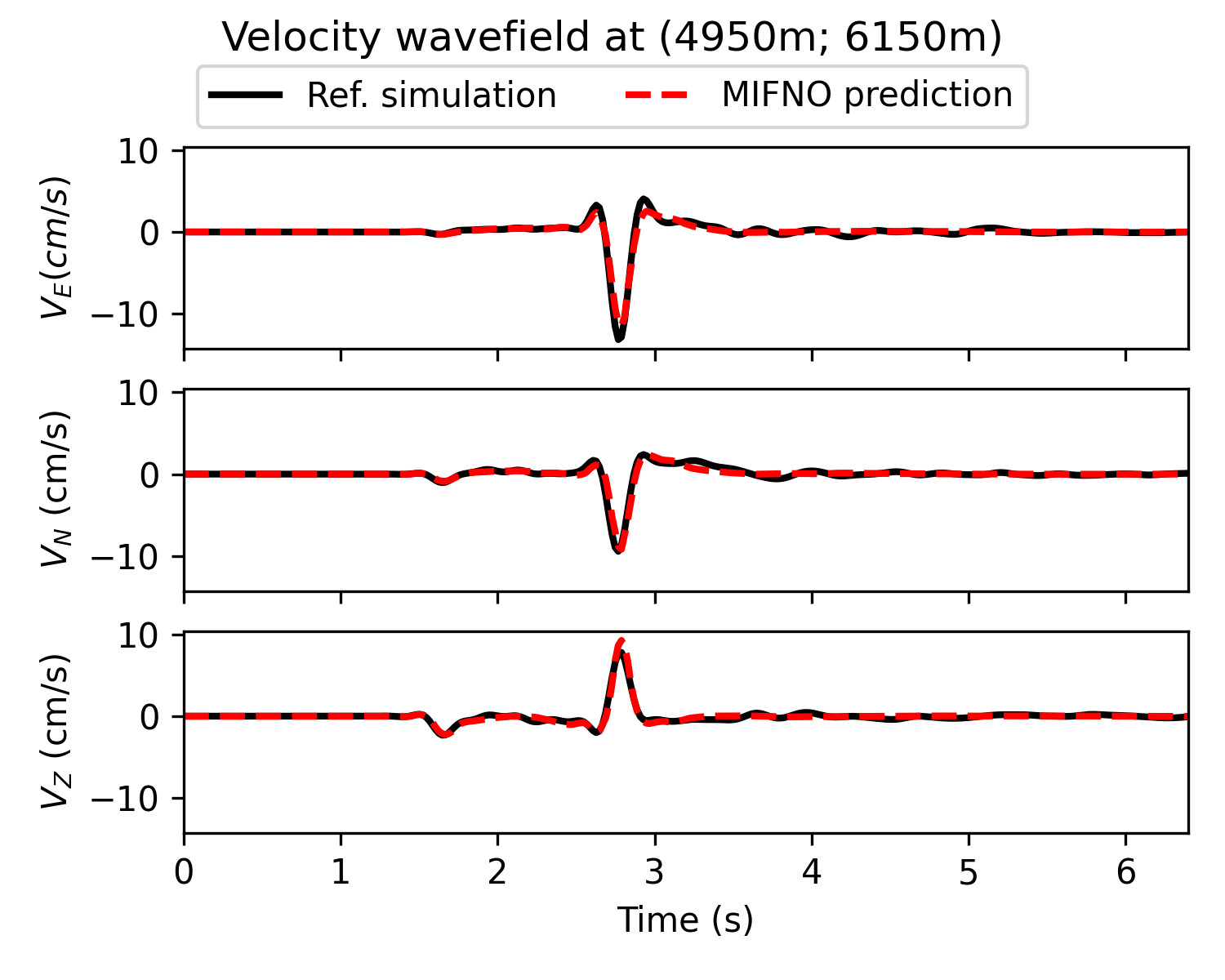}
		\end{subfigure}
		\hfill
		\begin{subfigure}[t]{0.08\textwidth}
			(c)
		\end{subfigure}
		\hfill
		\begin{subfigure}[t]{0.9\textwidth}
			\includegraphics[width=\textwidth]{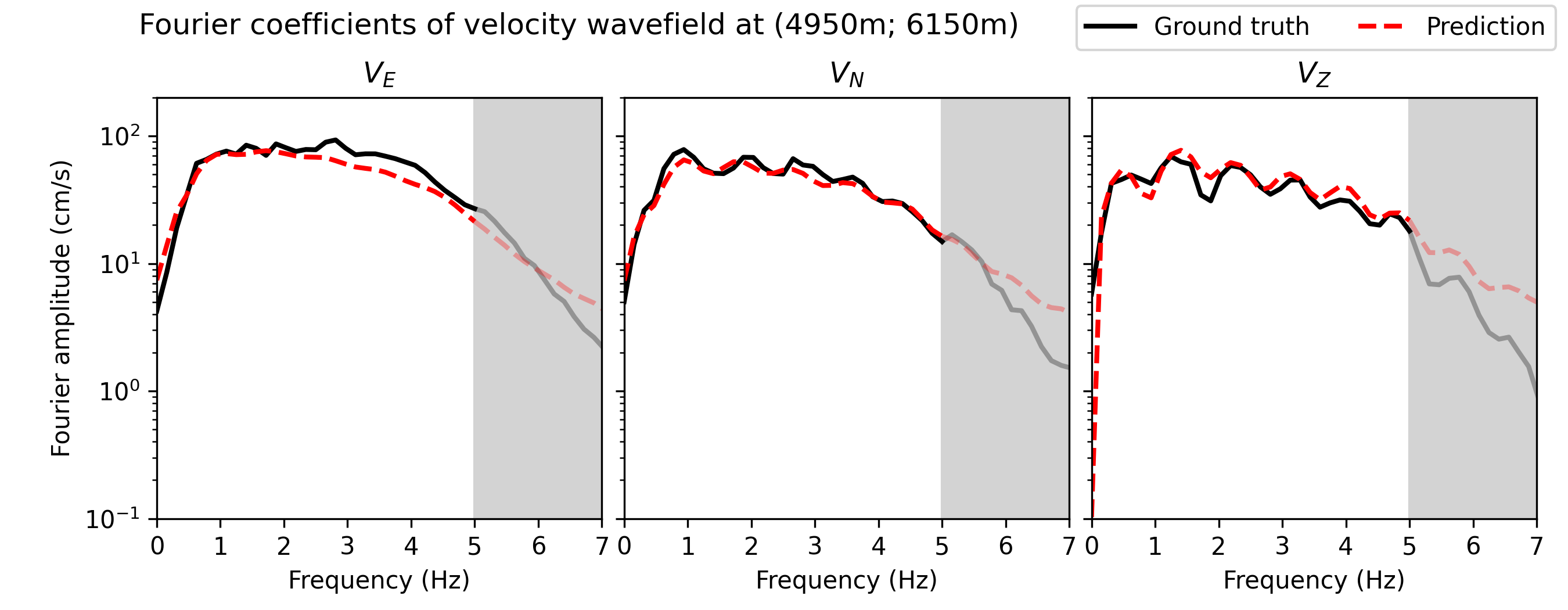}
		\end{subfigure}
		\caption{(a) Geology in the test dataset corresponding to the predictions in panels (b) and (c). The source is located at (\SI{3.9}{}, \SI{2.6}{}, \SI{-6.2}{km}) with a strike of \ang{298.7}, dip of \ang{85.3}, and rake of \ang{15.4}. (b) Velocity time series simulated (black) and predicted (dashed red line) in the three components: East-West (E), North-South (N), Vertical (Z). (c) For the same sensor as panel (b), amplitude of the Fourier coefficients of the velocity time series.}
		\label{fig:timeseries_testgeol}
	\end{figure}

	Figure \ref{fig:timeseries_testgeol} illustrates the MIFNO predictions for a geology in the test dataset (Fig. \ref{fig:timeseries_testgeol}a) and a source located at (\SI{3.9}{}, \SI{2.6}{}, \SI{-6.2}{km}) with a strike of \ang{298.7}, dip of \ang{85.3}, and rake of \ang{15.4}. The time series show an excellent accuracy on the three components since the wave arrival times are exactly reproduced: P-waves around \SI{1.6}{s} and S-waves around \SI{2.7}{s}. The peaks amplitude is also very similar between the prediction and the simulated ground truth. The frequency representation confirms the excellent agreement for all frequency ranges (Fig. \ref{fig:timeseries_testgeol}c) and the three components.
	
	At the depicted sensor, the envelope GOF is \SI{8.7}{} and the phase GOF is \SI{9.5}{}. Since both scores are excellent, they reflect the visual agreement that can be observed in Fig. \ref{fig:timeseries_testgeol}. The envelope GOF is slightly lower than the phase GOF due to the presence of small-scale and small-amplitude fluctuations that are challenging to predict with the MIFNO (visible around \SI{4}{s} in Fig. \ref{fig:timeseries_testgeol}b). 	
	
	\begin{figure}[h]
		\centering
		\includegraphics[width=\textwidth]{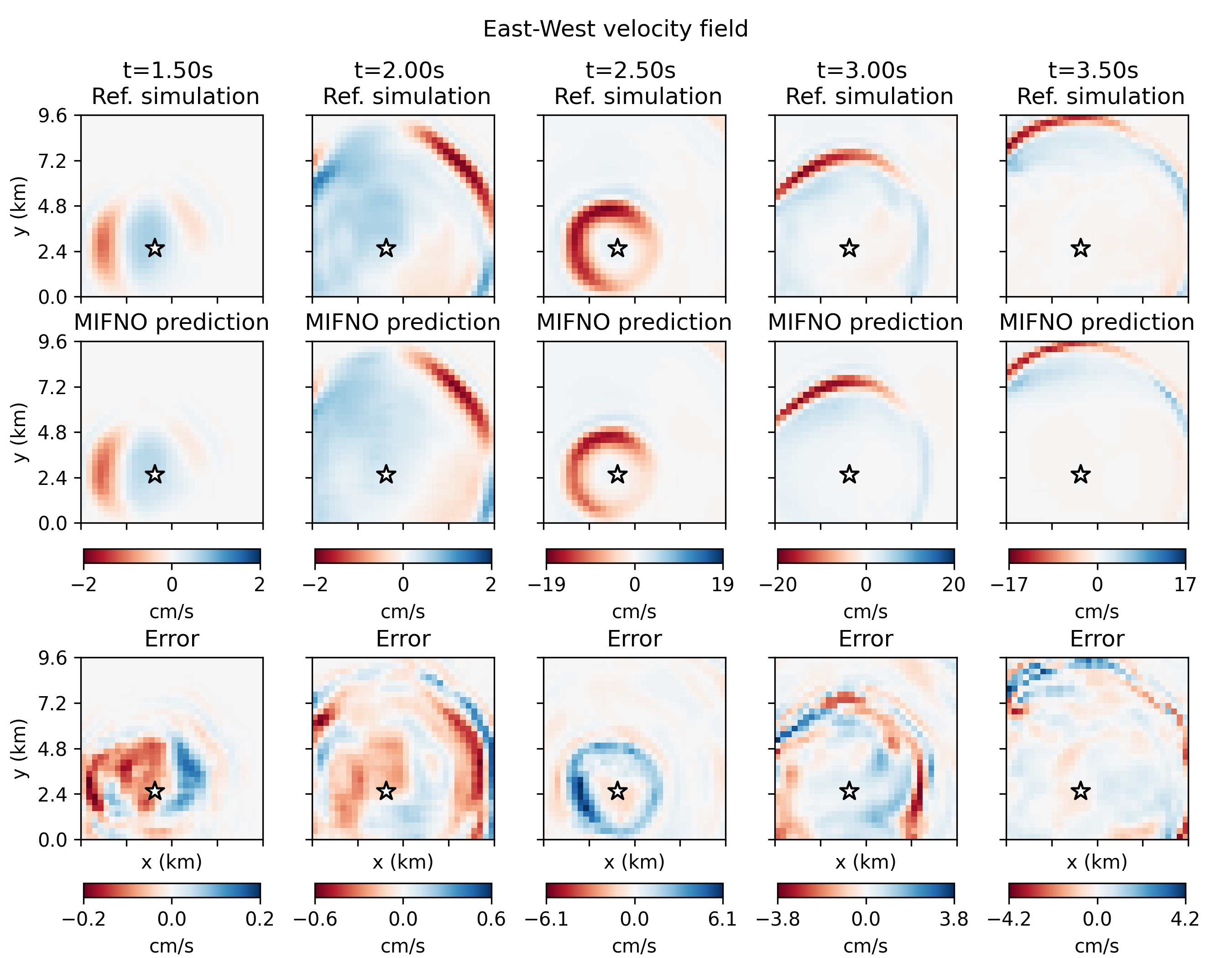}
		\caption{East-West component of the simulated (upper row) and predicted (center row) velocity fields for the geology illustrated in Fig. \ref{fig:timeseries_testgeol} at five time instants. The error between simulation and prediction is given in the lower row. The white star indicates the epicenter.}
		\label{fig:snapshots_time}
	\end{figure}
	
	When looking at the spatial evolution of the predictions, Fig. \ref{fig:snapshots_time} shows that it is very accurate for the different time steps. The wavefront propagates from the source (epicenter denoted with the white star) with the correct speed since the arrival times are correct for all spatial locations. The source orientation is clearly visible in the first snapshot and it is correctly reproduced by the MIFNO. It is also remarkable that both P waves and S waves are well predicted, although their amplitude is very different (maximum East-West amplitude around \SI{2}{cm/s} for the P waves and around \SI{20}{cm/s} for the S waves). One can however notice that the predictions are smoother after the main wavefront, which reflects that small-scale fluctuations are under-estimated. This is visible for instance at $t$=\SI{2.00}{s} in Fig. \ref{fig:snapshots_time} where the predicted wavefields show less fluctuations than the simulated ground truth.

\subsection{Metrics analyses}
\begin{figure}[h]
		\centering
		\includegraphics[width=0.8\textwidth]{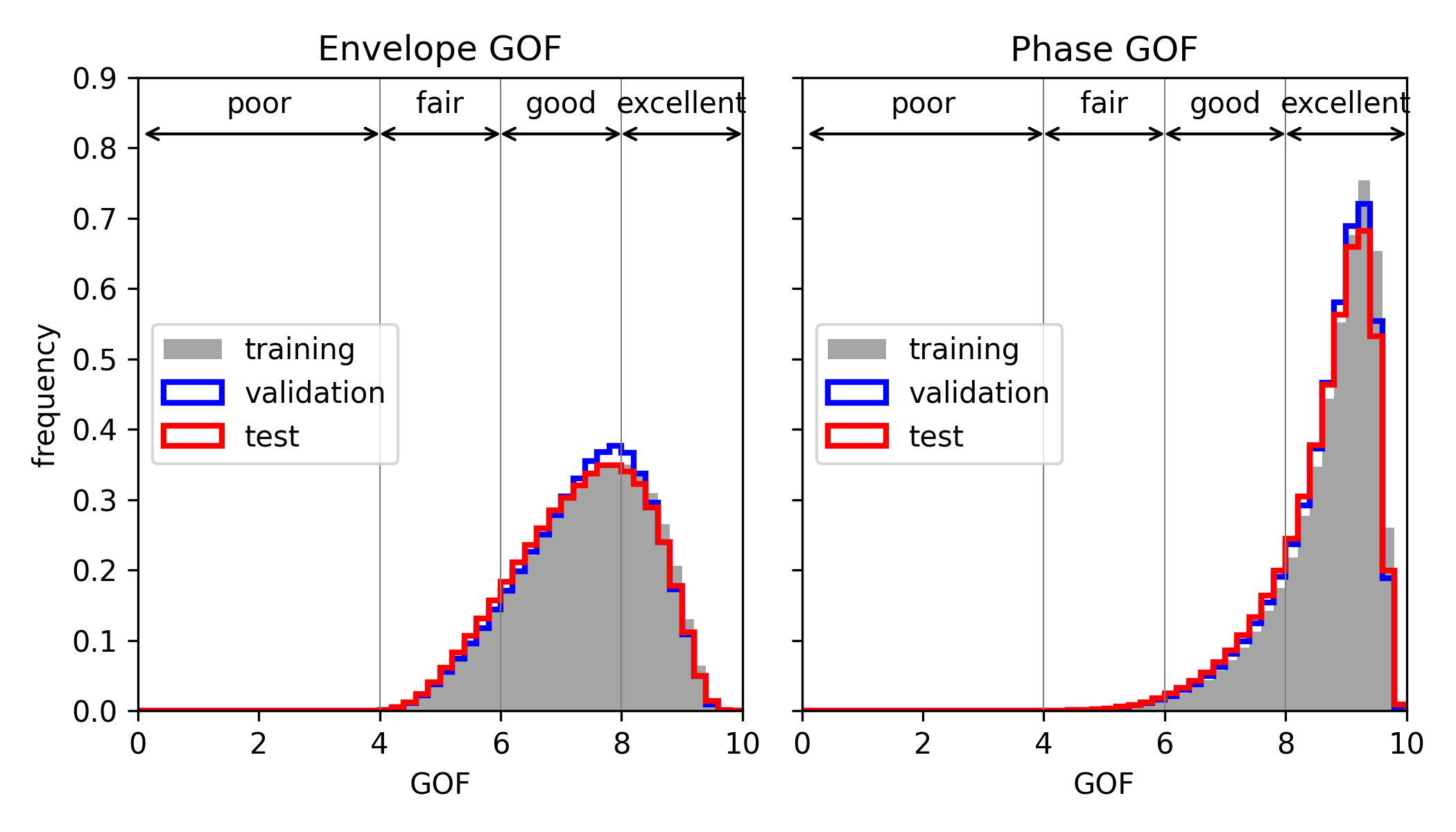}
		\caption{Distribution of the envelope GOF (left) and phase GOF (right) for each sensor and 1000 samples in the training (grey area), validation (blue line), test (red line) datasets.}
		\label{fig:GOF_distrib}
	\end{figure}

	To quantify the predictions accuracy more systematically, the envelope and phase GOFs were computed for 1000 samples in the test dataset. Figure \ref{fig:GOF_distrib} shows that the GOF distributions are very similar on the training, validation, and test datasets. This indicates that the MIFNO is not subject to overfitting. Although the loss evolution during training shows a stable convergence with a small generalization gap (Fig. \ref{fig:loss_history}), this does not seem to impact the prediction accuracy.
	
	The GOF distributions confirm that the phase accuracy is better than the envelope accuracy (Fig. \ref{fig:GOF_distrib} and Tab. \ref{tab:metrics_MIFNO}). Indeed, \SI{80.2}{\%} of predictions have an excellent phase GOF (phase GOF larger than 8). Envelope GOFs are good for \SI{87.3}{\%} of predictions (envelope GOF larger than 6), and \SI{27.8}{\%} are excellent (envelope GOF larger than 8). 
	
	Table \ref{tab:metrics_MIFNO} indicates that the relative Root Mean Square Error (rRMSE) is \SI{20}{\%} on the test dataset. The rRMSE and relative Mean Absolute Error (rMAE) are computed between (normalized) predicted and simulated velocity time series at each sensor, following
	\begin{equation}
	rRMSE(x,y) = \sqrt{\dfrac{1}{N_t} \sum_{k=1}^{Nt} \dfrac{(\hat{u}(x, y, t_k) - u(x, y, t_k))^2}{u(x, y, t_k)^2 + \epsilon^2}}
	\label{eq:rRMSE}
	\end{equation}
	\begin{equation}
	rMAE(x,y) = \dfrac{1}{N_t} \sum_{k=1}^{Nt} \dfrac{|\hat{u}(x, y, t_k) - u(x, y, t_k)|}{|u(x, y, t_k) + \epsilon|}
	\label{eq:rMAE}
	\end{equation}
	where $\epsilon = 0.01$, $x$ and $y$ indicate the sensor coordinates, and $t$ iterates from \SI{0}{s} to \SI{6.4}{s}.
	Errors are then averaged over all sensors and samples. Relative errors on the order of \SI{20}{\%} are common in 3D PDE predictions, especially when complexity arises from the presence of heterogeneous parameters. 
	
	Due to the limited interpretability of the RMSE, it is complemented by the frequency biases, computed as the errors between the amplitudes of Fourier coefficients in simulated $u(x,y)$ and predicted $\hat{u}(x,y)$ velocity fields. Frequency biases are given for low frequency (0-1Hz), medium frequency (1-2Hz), and high frequency (2-5Hz) components following 
	\begin{equation}
	\text{rFFT\textsubscript{low}} = \dfrac{\overline{\mathcal{F}\left( \hat{u}(x,y) \right)}_{low} - \overline{\mathcal{F}\left( u(x,y) \right)}_{low}} {\overline{\mathcal{F}\left( u(x,y) \right)}_{low}}
	\end{equation}
	with 
	\begin{equation}
	\overline{\mathcal{F}\left( u(x,y) \right)}_{low} = \dfrac{1}{N_f} \sum_{0 \le f < 1} |\mathcal{F}(u(x,y))(f)|
	\label{eq:mean_fourier_amplitude}
	\end{equation}
	In equation (\ref{eq:mean_fourier_amplitude}), $N_f$ denotes the number of frequencies $f$ between 0 and 1Hz and $\mathcal{F}$ is the Fourier transform. Definitions naturally extend to medium and high frequencies. It should be noted that positive frequency biases indicate an overestimation of the frequency content compared to the reference ($\overline{\mathcal{F}\left( \hat{u}(x,y) \right)}~\ge~\overline{\mathcal{F}\left( u(x,y) \right)}$) while negative values indicate underestimation. Noting that the mean of Fourier coefficients amplitudes $\overline{\mathcal{F}\left( u(x,y) \right)}$ is always positive, this implies that the frequency biases cannot be lower than $-1$. This explains the steep decrease of the high-frequency distribution around $-1$ in Fig. \ref{fig:error_distrib}.

	\begin{table}[h]
		\hspace{-1cm}
		\footnotesize
		\begin{center}	
			\begin{tabular}{|c|c|c|c|c|c|c|c|}
				\hline
				Model & rMAE & rRMSE & rFFT\textsubscript{low} & rFFT\textsubscript{mid} & rFFT\textsubscript{high} & EG & PG \\
				\hline
				MIFNO 8 layers & 0.14 $\pm$ 0.05 & 0.21 $\pm$ 0.06 & -0.30 $\pm$ 0.18 & -0.41 $\pm$ 0.19 & -0.50 $\pm$ 0.21 & 6.98 $\pm$ 0.77 & 8.46 $\pm$ 0.58 \\
				MIFNO 16 layers & 0.13 $\pm$ 0.05 & 0.20 $\pm$ 0.07 & -0.21 $\pm$ 0.17 & -0.29 $\pm$ 0.19 & -0.37 $\pm$ 0.21 & 7.34 $\pm$ 0.75 & 8.63 $\pm$ 0.54 \\
				\hline
			\end{tabular}
		\end{center}
		\caption{Mean and standard deviation of the metrics computed on 1000 test samples. rMAE: relative MAE (0 is best), rRMSE: relative RMSE (0 is best), rFFT\textsubscript{low}: relative frequency bias 0-1Hz (0 is best), rFFT\textsubscript{mid}: relative frequency bias 1-2Hz (0 is best), rFFT\textsubscript{high}: relative frequency bias 2-5Hz (0 is best), EG: Envelope GOF (10 is best), PG: Phase GOF (10 is best). For frequency biases, negative values indicate underestimation while positive values indicate overestimation.}
		\label{tab:metrics_MIFNO}
	\end{table}
	
	Table \ref{tab:metrics_MIFNO} indicates that the MIFNO mainly underestimates the frequency content of signals since frequency biases are mostly negative. This can be observed in Fig. \ref{fig:timeseries_testgeol} where the small fluctuations at the end of the signal tend to be ignored in the predictions. Previous findings on the F-FNO for a dataset with a fixed source corroborate these results \citep{lehmann3DElasticWave2024}. Also, the MIFNO leads to larger underestimations for high-frequency components (the inter-quartile range is \SI{-62}{\%} ;  \SI{-11}{\%} for high-frequency components while it reduces to \SI{-49}{\%}; \SI{-5}{\%} for low frequencies, Tab. \ref{tab:metrics_MIFNO}). This reflects the well-known spectral bias, which states that small-scale (i.e. high-frequency) features are more difficult to predict than large-scale features \citep{basriFrequencyBiasNeural2020, rahamanSpectralBiasNeural2019}. In addition, the small-scale fluctuations also reflect complex physical phenomena due to the refraction of seismic waves on scatterers. Indeed, fluctuations are mostly observed in the coda, after the S-wave arrival. Hence, they originate from the late arrivals of diffracted waves. Late velocity fields are subject to a large inter-sample variability, both between different sensors and between similar geologies with different locations of scatterers. These factors make high-frequency predictions very challenging.

\subsection{On the influence of hyperparameters}
\label{sec:hyperparams}
	The study of hyperparameters was limited to the most influential parameters highlighted by a previous work on the F-FNO \citep{lehmann3DElasticWave2024}: the number of layers, the number of channels ($d_v$), and the dataset size.
	
\begin{figure}[h]
		\centering
		\includegraphics[width=\textwidth]{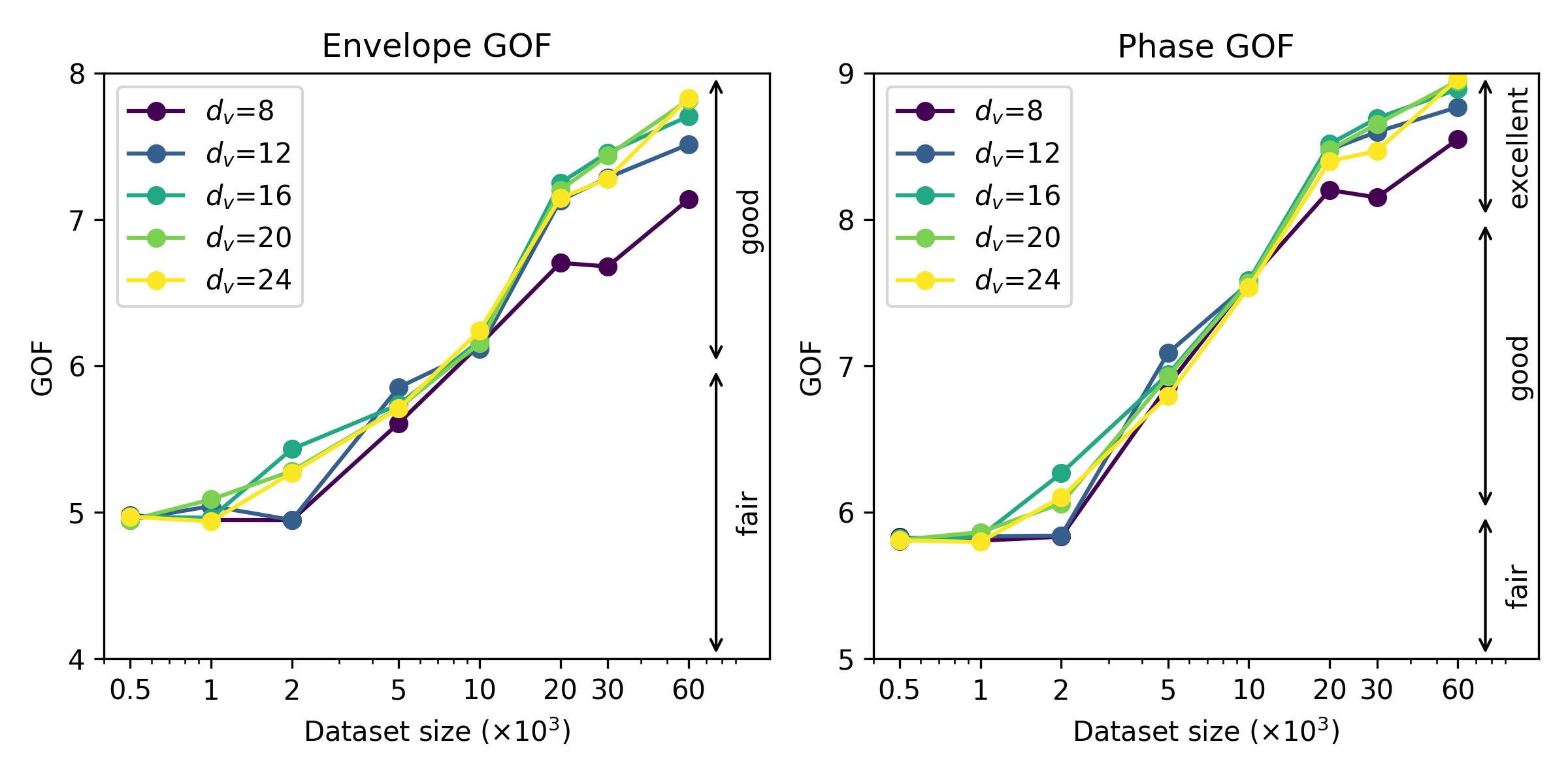}
		\caption{Envelope GOF (left) and phase GOF (right) as a function of the number of training samples, for different number of channels ($d_v$). The MIFNO contains 16 layers and all other hyperparameters are fixed to the values described in Section \ref{sec:MIFNO}}
		\label{fig:error_dv}
	\end{figure}

	One can notice in Tab. \ref{tab:metrics_MIFNO} that the deeper MIFNO with 16 layers gives better predictions than the MIFNO with 8 layers. Figure \ref{fig:snapshots_8_16layers} illustrates that wavefronts are more accurately defined with the 16-layer MIFNO and the largest amplitudes are closer to the reference.
	The number of training samples also has a significant impact on the prediction accuracy. Figure \ref{fig:error_dv} shows a steep increase in the GOF values when the dataset size increases. One can especially notice that at least \SI{10 000}{} samples are required to obtain correct results (mean envelope GOF considered as good). In addition, metrics tend to improve when the number of channels is higher, although the benefits are limited beyond $d_v=16$. One has to keep in mind that the number of channels is multiplied by three after the combination of the geology and source branches. Therefore, an initial number of channels $d_v=16$ becomes $d_v=48$ in the 12 last Fourier layers, which is in line with large values found in common F-FNO implementations.
	Therefore, increasing the complexity of the model is beneficial to improve its accuracy and does not lead to overfitting. This improvement is not reflected by the rRMSE and rMAE while it is obvious on the other metrics (Tab. \ref{tab:metrics_MIFNO}), thereby emphasizing the need to consider physically-rooted metrics to evaluate the predictions accuracy.
	
	To train the MIFNO with \SI{60 000}{} samples while the HEMEW\textsuperscript{S}-3D database contains only \SI{30 000}{} samples, we performed data augmentation from the rotation invariance property of the data. Indeed, by rotating geologies, source, and wavefields with a 90\textdegree \ angle around the vertical axis, one creates new samples that obey the same physical laws. The original database of \SI{30 000}{} samples can then be extended to $4 \times 30,000 = 120,000$ samples. However, due to the long training times with \SI{120 000}{} samples, investigations were performed only up to \SI{60 000}{} samples.
	
	Table \ref{tab:metrics_MIFNO_rotations} shows that training the MIFNO (parameterized with the moment tensor) on the augmented database (\SI{30 000}{} original samples, third row) yields slightly better metrics than training on the original database (\SI{7500}{} original samples augmented by three rotations to give \SI{30 000}{} training samples, second row). The test data are the same in all cases and contain no data augmentation. Table \ref{tab:metrics_MIFNO_rotations} also emphasizes that the dataset size is a determining factor, since results are considerably worse when using only \SI{7500}{} samples without data augmentation. Therefore, including rotations in the training dataset is an efficient way to increase the dataset size without adding data redundancy and hence, improve the prediction accuracy.

	\begin{table}[h]
		\footnotesize
		\begin{center}
			\begin{tabular}{|c|c|c|c|c|c|c|c|}
				\hline
				Data & rMAE & rRMSE & rFFT\textsubscript{low} & rFFT\textsubscript{mid} & rFFT\textsubscript{high} & EG & PG \\
				\hline
				7,500 original & 0.15 $\pm$ 0.05 & 0.24 $\pm$ 0.06 & -0.44 $\pm$ 0.17 & -0.61 $\pm$ 0.16 & -0.73 $\pm$ 0.14 & 6.24 $\pm$ 0.60 & 7.56 $\pm$ 0.64 \\
				30,000 original & 0.13 $\pm$ 0.05 & 0.20 $\pm$ 0.07 & -0.21 $\pm$ 0.17 & -0.29 $\pm$ 0.19 & -0.37 $\pm$ 0.21 & 7.34 $\pm$ 0.75 & 8.63 $\pm$ 0.54 \\
				7,500 $\times$ 4 rotations & 0.13 $\pm$ 0.05 & 0.20 $\pm$ 0.07 & -0.18 $\pm$ 0.16 & -0.26 $\pm$ 0.18 & -0.33 $\pm$ 0.21 & 7.46 $\pm$ 0.73 & 8.69 $\pm$ 0.52 \\
				\hline
			\end{tabular}
		\end{center}
		\caption{Mean and standard deviation of the metrics computed on 1000 test samples. 1st line: MIFNO trained with \SI{7 500}{} samples from the HEMEW\textsuperscript{S}-3D database. 2nd line: MIFNO trained with \SI{30 000}{} samples from the HEMEW\textsuperscript{S}-3D database. 3rd line: MIFNO trained with \SI{7 500}{} samples from the HEMEW\textsuperscript{S}-3D database augmented by rotations of 90\textdegree, 180\textdegree, and 270\textdegree \ around the vertical axis. rRMSE: relative RMSE (0 is best), rFFT\textsubscript{low}: relative frequency bias 0-1Hz (0 is best), rFFT\textsubscript{mid}: relative frequency bias 1-2Hz (0 is best), rFFT\textsubscript{high}: relative frequency bias 2-5Hz (0 is best), EG: envelope Goodness-of-Fit (10 is best), PG: phase Goodness-of-Fit (10 is best). For frequency biases, negative values indicate underestimation.}
		\label{tab:metrics_MIFNO_rotations}
	\end{table}

	It can also be noted that there is no significant difference between the MIFNO taking as inputs the three angles describing the source orientation or the corresponding moment tensor (Tab. \ref{tab:metrics_MIFNO_angle_moment}). Therefore, the MIFNO with angles is used in the following for more direct interpretations of the results, although the MIFNO with moment should be used with data augmentation due to the existence of two sets of angles for each moment tensor.

\subsection{Comparison with baseline models}
\label{sec:baselines}
	Due to the difficulty of finding existing implementations of models for 3D and vector inputs, the MIFNO is compared to F-FNO models with less flexibility. This is meant to ensure that the additional complexity induced by the \textit{source branch} does not deteriorate the predictions. To do so, we use three different databases that contain the same geological models but different surface velocity wavefields depending on the source parametrization:
	\begin{enumerate}
		\item in the HEMEW-3D database, both the source position and orientation are fixed. The source is located in the middle of the bottom layer: $x_s$=\SI{4.8}{km}, $y_s$=\SI{4.8}{km}, $z_s$=\SI{-8.4}{km} and its orientation is fixed to strike=\ang{48}, dip=\ang{45}, rake=\ang{88} \citep{lehmann3DElasticWave2024}. This database contains N\textsubscript{train}=\SI{27000}{} training, N\textsubscript{val}=\SI{3000}{} validation and N\textsubscript{test}=\SI{1000}{} test samples.
		\item in an intermediate database, the source is located randomly inside the bottom layer, i.e. $x_s \sim \mathcal{U}([1200; 8100 \: m])$, $y_s \sim \mathcal{U}([1200; 8100 \: m])$, $z_s \sim \mathcal{U}([-9000; -8100 \: m])$ and its orientation is fixed to the same value as the HEMEW-3D database. This database contains N\textsubscript{train}=\SI{20000}{} training, N\textsubscript{val}=\SI{2000}{} validation and N\textsubscript{test}=\SI{1000}{} test samples.
		\item the HEMEW\textsuperscript{S}-3D database described in Section \ref{sec:data_HEMEWS} contains sources located randomly anywhere in the domain and with any source orientation. It contains N\textsubscript{train}=\SI{27000}{} training, N\textsubscript{val}=\SI{3000}{} validation and N\textsubscript{test}=\SI{1000}{} test samples.
	\end{enumerate}
	
	First, let us consider the prediction accuracy with respect to the geological properties. The baseline model is the 16-layer F-FNO described in Section \ref{sec:FFNO}. To match the number of parameters of the MIFNO, the F-FNO was designed with $d_v$=\SI{16}{} channels in layers 1 to 4, and $d_v$=\SI{48}{} channels in layers 5 to 16. The F-FNO was trained on the first database HEMEW-3D with fixed source location and orientation. The MIFNO was trained as described above on the HEMEW\textsuperscript{S}-3D database with random source location and orientation. Therefore, it should be noted that the training task of the MIFNO is much more complex than the F-FNO's task.
	
	\begin{figure}[h]
		\centering
		\includegraphics[width=\textwidth]{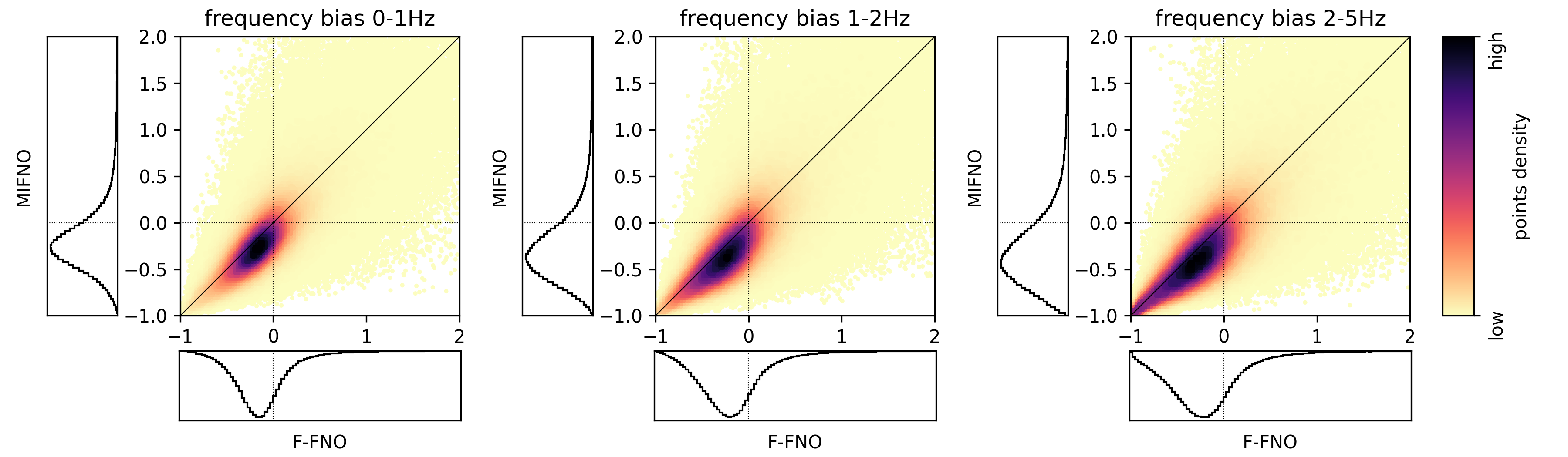}
		\caption{For 1000 samples in the HEMEW-3D test dataset (fixed position and fixed orientation) and each sensor, the frequency bias of the MIFNO (vertical axis) is shown against the F-FNO frequency bias (horizontal axis). Each subplot corresponds to a different frequency bias, 0-1Hz (left), 1-2Hz (middle), 2-5Hz (right). Optimal predictions correspond to a frequency bias of \SI{0}{}.}
		\label{fig:MIFNO_vs_FFNO}
	\end{figure}
	
	At inference stage, both the F-FNO and the MIFNO were used to predict surface velocity wavefields for 1000 test geologies with a fixed source corresponding to the reference source in the HEMEW-3D database. Figure \ref{fig:MIFNO_vs_FFNO} compares the F-FNO predictions with the MIFNO predictions for each sensor and each sample. One can firstly notice that points are widespread on both sides of the diagonal, meaning that the MIFNO or the F-FNO can give better predictions depending on the situation. However, points are more densely distributed in the lower right triangle of the graph, close to the diagonal (Fig. \ref{fig:MIFNO_vs_FFNO}). This indicates that the F-FNO is in average more accurate than the MIFNO when predicting only solutions with a fixed source. 
	
	However, the difference between the F-FNO and the MIFNO is around \SI{0.8}{} GOF units, which remains acceptable knowing the extended complexity of the MIFNO's task (Tab. \ref{tab:metrics_FNO_fixedsource}). Figure \ref{fig:line_timeseries_MIFNO_vs_FNO} illustrates velocity time series predicted by the F-FNO and the MIFNO. It shows that the MIFNO predictions are already reasonable while the F-FNO locally improves the accuracy.	
	
	For the next comparison, we focus on the influence of the source location. To do so, we designed a F-FNO taking as input the geological parameters and a binary encoding of the source position, i.e. a cube full of zeros with a single 1 indicating the position of the source. The MIFNO was trained with only the source coordinates given as inputs to the source branch. In this experiment, both the F-FNO and the MIFNO were trained on the second database (random source location and fixed orientation).

	\begin{table}[h]
		\begin{center}
			Dataset with a random source position and fixed source orientation
			\begin{tabular}{|c|c|c|c|c|c|c|}
				\hline
				Model & rRMSE & rFFT\textsubscript{low} & rFFT\textsubscript{mid} & rFFT\textsubscript{high} & EG & PG \\
				\hline
				F-FNO & 0.16 $\pm$ 0.07 & -0.03 $\pm$ 0.96 & -0.06 $\pm$ 0.72 & -0.08 $\pm$ 1.37 & 7.99 $\pm$ 0.85 & 8.99 $\pm$ 0.60 \\
				MIFNO & 0.17 $\pm$ 0.07 & -0.04 $\pm$ 0.80 & -0.07 $\pm$ 0.76 & -0.09 $\pm$ 1.04 & 7.88 $\pm$ 0.87 & 8.92 $\pm$ 0.63 \\
				\hline
			\end{tabular}
		\end{center}
		\caption{1st and 3rd quartiles of the metrics computed on 1000 validation samples. rRMSE: relative RMSE (0 is best), rFFT\textsubscript{low}: relative frequency bias 0-1Hz (0 is best), rFFT\textsubscript{mid}: relative frequency bias 1-2Hz (0 is best), rFFT\textsubscript{high}: relative frequency bias 2-5Hz (0 is best), EG: envelope GOF (10 is best), PG: phase GOF (10 is best). For frequency biases, negative values indicate underestimation. Both models contain 16 layers and were trained with \SI{20000}{} samples for 300 epochs.}
		\label{tab:metrics_FNO_movingsource}
	\end{table}
	
	When both models have 8 layers, the MIFNO predictions are slightly better than the F-FNO (Tab. \ref{tab:metrics_FNO_movingsource_8layers}). With 16 layers, both models give a similar accuracy (second digit fluctuations in Tab. \ref{tab:metrics_FNO_movingsource} are within the variability of the model due to a different random initialization). Phase GOFs are excellent for both models and the envelope GOFs are close to excellent (Tab. \ref{tab:metrics_FNO_movingsource}).  Overall, these results show that the \textit{source branch} architecture in the MIFNO preserves the expressivity of the F-FNO while providing the flexibility to add more source parameters. 
	
	Lastly, we compare the MIFNO with the F-FNO on the HEMEW\textsuperscript{S}-3D database. In this setting, the MIFNO architecture is the reference one described in Section \ref{sec:MIFNO} and the F-FNO considers all the source parameters as constant cubes. More precisely, the geological model is concatenated with the six cubes $x_s * \mathbb{J}$, $y_s * \mathbb{J}$, $z_s * \mathbb{J}$, $\bm{\theta_s}[1] * \mathbb{J}$, $\bm{\theta_s}[2] * \mathbb{J}$, $\bm{\theta_s}[3] * \mathbb{J}$ (where $\mathbb{J}$ denotes a cube full of ones), in addition to the grids of coordinates. With this implementation, the F-FNO uplift layer has slightly more parameters than the MIFNO uplift layer. But since the F-FNO does not contain the additional source branch, it globally contains less parameters than the MIFNO (3.11 million parameters).

	\begin{table}[h]
		\footnotesize
		\begin{center}
			Dataset with a random source position and random source orientation
			\begin{tabular}{|c|c|c|c|c|c|c|c|}
				\hline
				Model & rMAE & rRMSE & rFFT\textsubscript{low} & rFFT\textsubscript{mid} & rFFT\textsubscript{high} & EG & PG \\
				\hline
				F-FNO & 0.13 $\pm$ 0.05 & 0.21 $\pm$ 0.07 & -0.22 $\pm$ 0.17 & -0.31 $\pm$ 0.19 & -0.39 $\pm$ 0.21 & 7.25 $\pm$ 0.75 & 8.52 $\pm$ 0.57 \\
				MIFNO & 0.13 $\pm$ 0.05 & 0.2 $\pm$ 0.07 & -0.21 $\pm$ 0.17 & -0.29 $\pm$ 0.19 & -0.37 $\pm$ 0.21 & 7.34 $\pm$ 0.75 & 8.63 $\pm$ 0.54 \\
				\hline
			\end{tabular}
		\end{center}
		\caption{Mean and standard deviation of the metrics computed on 1000 test samples. rMAE: relative MAE (0 is best), rRMSE: relative RMSE (0 is best), rFFT\textsubscript{low}: relative frequency bias 0-1Hz (0 is best), rFFT\textsubscript{mid}: relative frequency bias 1-2Hz (0 is best), rFFT\textsubscript{high}: relative frequency bias 2-5Hz (0 is best), EG: envelope GOF (10 is best), PG: phase GOF (10 is best). For frequency biases, negative values indicate underestimation. Both models contain 16 layers and were trained with \SI{27000}{} samples for 250 epochs.}
		\label{tab:metrics_FNO_randomsource}
	\end{table}

	Table \ref{tab:metrics_FNO_randomsource} shows that the MIFNO is slightly more accurate than the F-FNO on the test dataset. All physics-based metrics improve when encoding the source parameters in the dedicated source branch of the MIFNO compared to the F-FNO. As we shall see in Section \ref{sec:sources_OOD}, the MIFNO advantage is even more important when considering out-of-distribution sources.

\subsection{Influence of the source parameters}
\label{sec:results_sourceparams}
\begin{figure}[h]
		\begin{subfigure}[h]{0.48\textwidth}
			\caption{}
			\includegraphics[width=\textwidth]{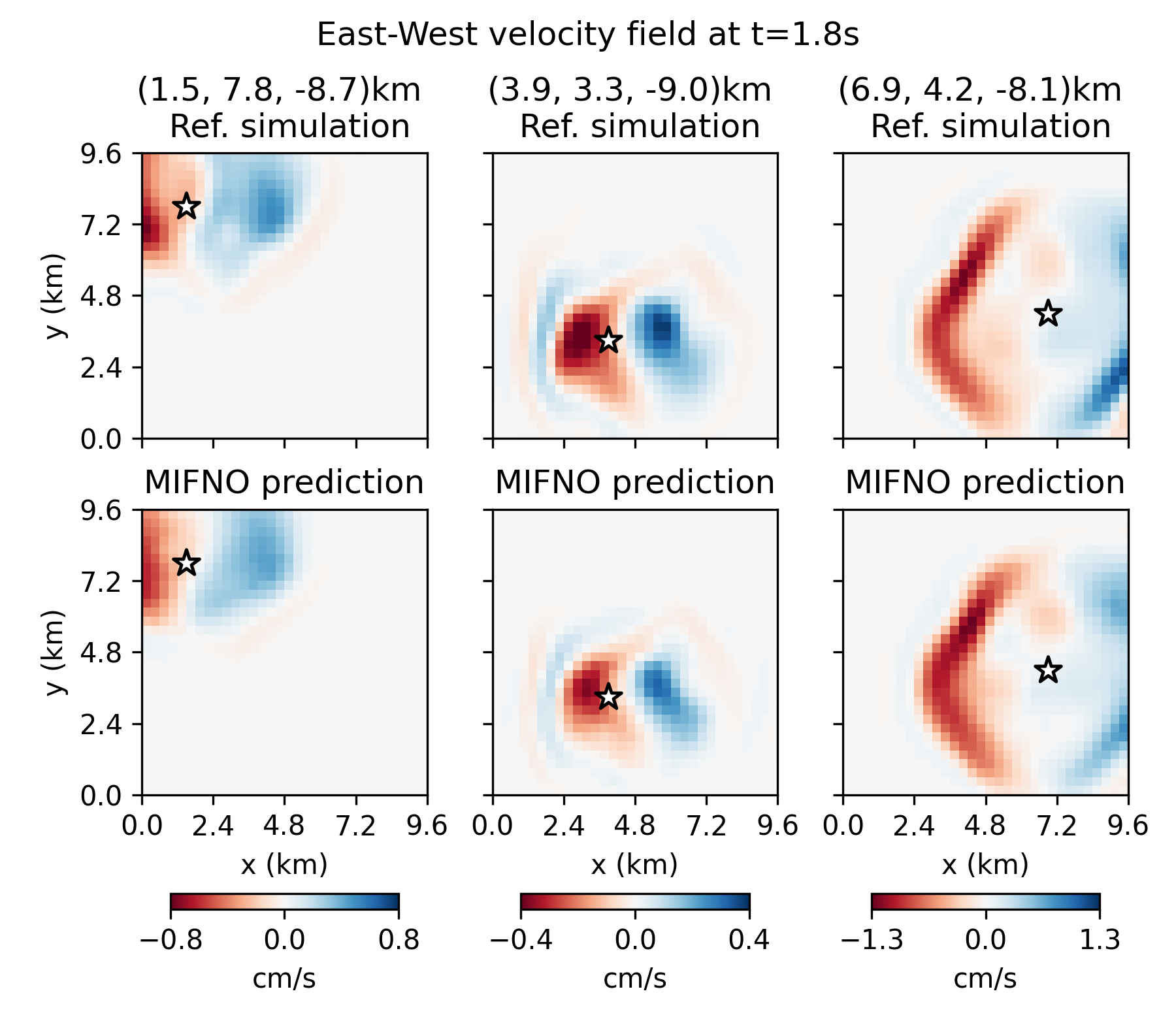}
		\end{subfigure}
		\hfill
		\begin{subfigure}[h]{0.48\textwidth}
			\caption{}
			\includegraphics[width=\textwidth]{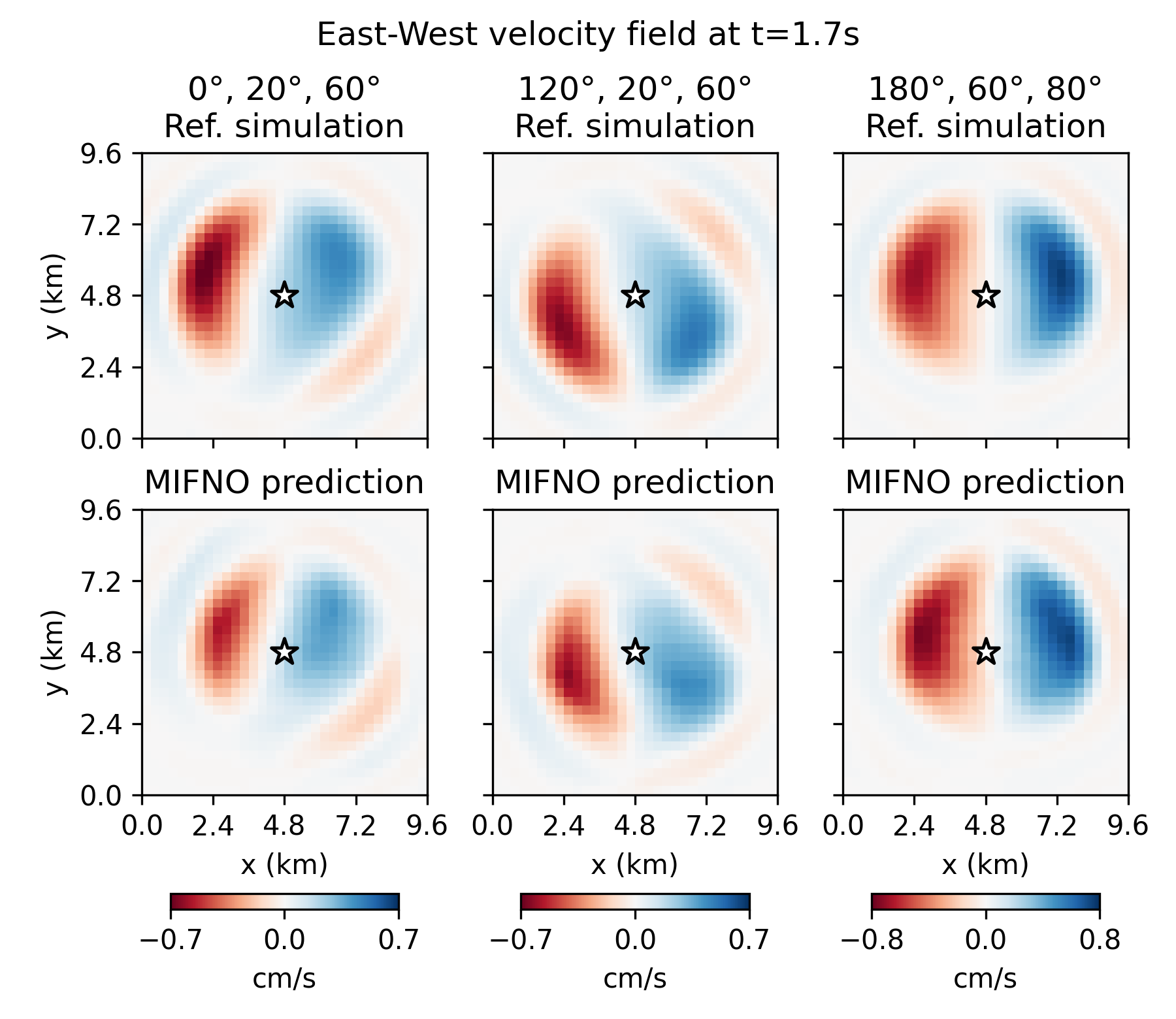}
		\end{subfigure}
		\caption{Comparison of the MIFNO predictions (lower rows) with the simulated ground truth (upper rows) for a given geology and time. (a) the source position changes (indicated by the white star) but the orientation remains constant. (b) the source position is fixed but the orientation changes and is indicated with the (strike, dip, rake) angles.}
		\label{fig:snapshots_moving_rotating_source}
	\end{figure}

	In this section, we highlight the influence of the source parameters on the predictions and on the error. For a given geology, Figure \ref{fig:snapshots_moving_rotating_source}a compares the predictions and the ground truth for three different positions of the source (with the same orientation). One can firstly note that the predicted wavefields are very similar to the ground truth for all source positions. Especially, their spatial location closely follows the source position. This confirms that the source position encoding in the \textit{source branch} is accurate. 
	
	Figure \ref{fig:snapshots_moving_rotating_source}a also illustrates the variability of surface wavefields that can be obtained by moving the source in a heterogeneous geology. Indeed, the same geology was used for all three snapshots and the source orientation was fixed. Since seismic waves originating from different source locations encounter different scatterers, they give rise to different surface wavefields. This large variability on the outputs - only due to the source position - reiterates the complexity of the task we aim at solving. 
	
	Next, we focus on the source orientation while fixing the source position. A geology with low heterogeneities is chosen to emphasize the source orientation. Three sets of (strike, dip, rake) angles are examined in Fig. \ref{fig:snapshots_moving_rotating_source}b. Surface wavefields are depicted shortly after the first P waves arrived at the surface of the domain. The wavefields phases are accurately predicted for all orientations, meaning that the MIFNO can predict the influence of the source orientation on ground motion. As already mentioned, a slight amplitude underestimation is visible in some regions but this is not due to the source orientation. 
	
	\begin{figure}[h]
		\centering
		\includegraphics[width=\textwidth]{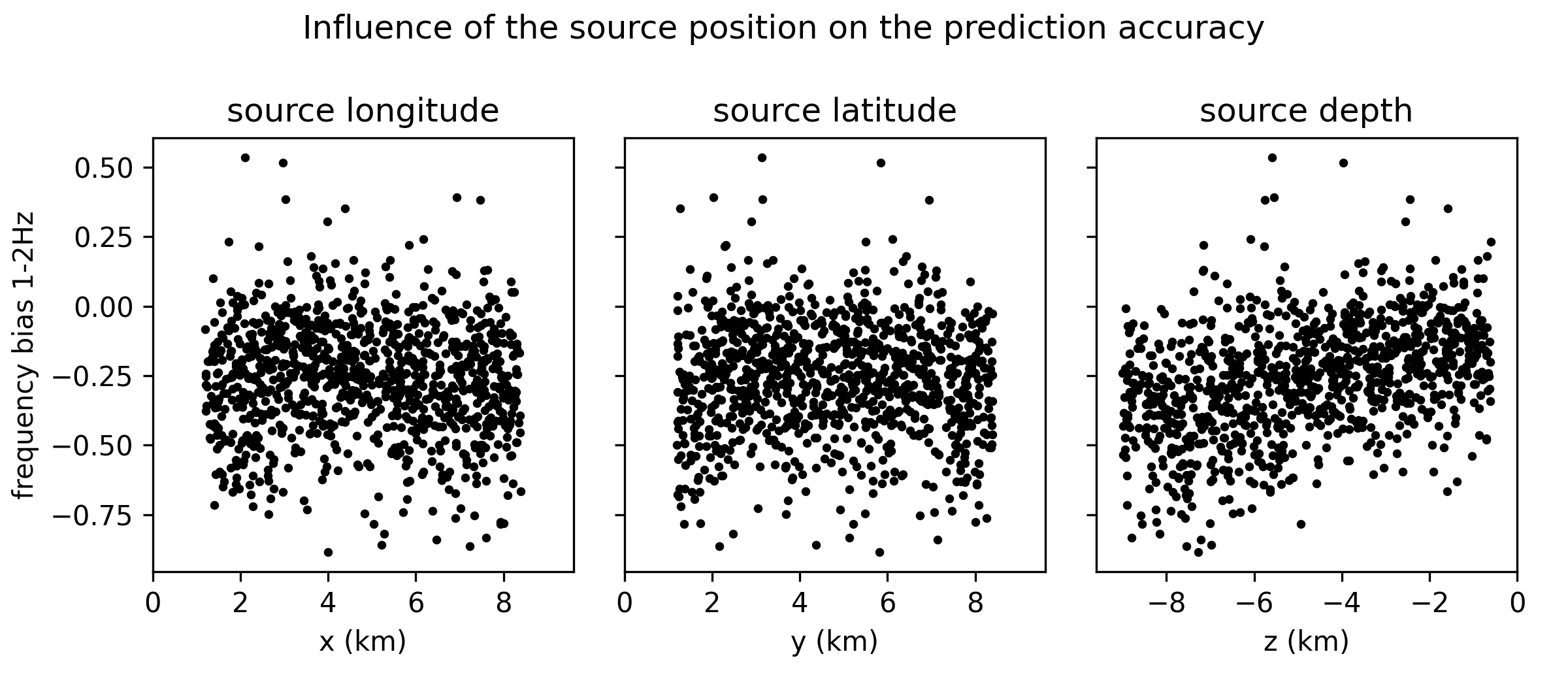}
		\caption{For 1000 samples, the 1-2 Hz frequency bias is shown against the source position ($x$, $y$, $z$). Negative (resp. positive) values indicate underestimation (resp. overestimation) of the frequency content.}
		\label{fig:error_vs_source_position}
	\end{figure}
	
	For more systematic analyses, the 1-2 Hz frequency bias is evaluated on 1000 samples with various geologies, source locations, and source orientations. Figure \ref{fig:error_vs_source_position} shows the frequency bias against the source coordinates. Similar errors are obtained for all source longitudes and latitudes, meaning that the prediction accuracy is independent from the epicenter location. There is a slightly positive correlation between the source depth and the frequency bias but the inter-sample variability is of the same order. Figure \ref{fig:error_vs_source_position} suggests that the underestimation of frequency content is greater when the source is deeper while errors are smaller and overestimation is more frequent for shallow sources.
	
	Since both the phase and envelope GOFs indicate better results for shallow sources (Fig. \ref{fig:GOF_vs_source_depth}), this effect is likely due to the increased complexity of the wavefields generated by long propagation paths. Indeed, seismic waves originating from deep sources have encountered more heterogeneities on their path, which leads to diffraction and dispersion and hence, perturbs the surface wavefields. Therefore, it is generally more complex to predict accurate wavefields generated by deep sources. 
	
	Additionally, it can be observed that the source orientation parameters (strike, dip, and rake angles) have no effect on the prediction accuracy (Fig. \ref{fig:error_vs_source_orientation}). This confirms that the MIFNO produces equally accurate predictions for all types of sources.

\subsection{Relationship between energy content and accuracy}
	It has already been mentionned that deeper sources tend to lead to larger errors. To provide a complimentary understanding, let us define the Energy Integral $EI(\bm{x})$ of the ground motion time series at sensor $\bm{x}$ as
	\begin{equation}
	EI(\bm{x}) = \sum_{k=1}^{N_t} \dfrac{u_E(\bm{x}, t_k)^2 + u_N(\bm{x}, t_k)^2 + u_Z(\bm{x}, t_k)^2}{3}
	\end{equation}
	and normalize it with respect to the maximum over all sensors 
	\begin{equation}
	\widetilde{EI}(\bm{x}) = \dfrac{EI(\bm{x})}{\max_{\bm{x' \in \partial \Omega_{top}}} EI(\bm{x'})}
	\end{equation}

	\begin{figure}[h]
		\centering
		\includegraphics[width=0.8\textwidth]{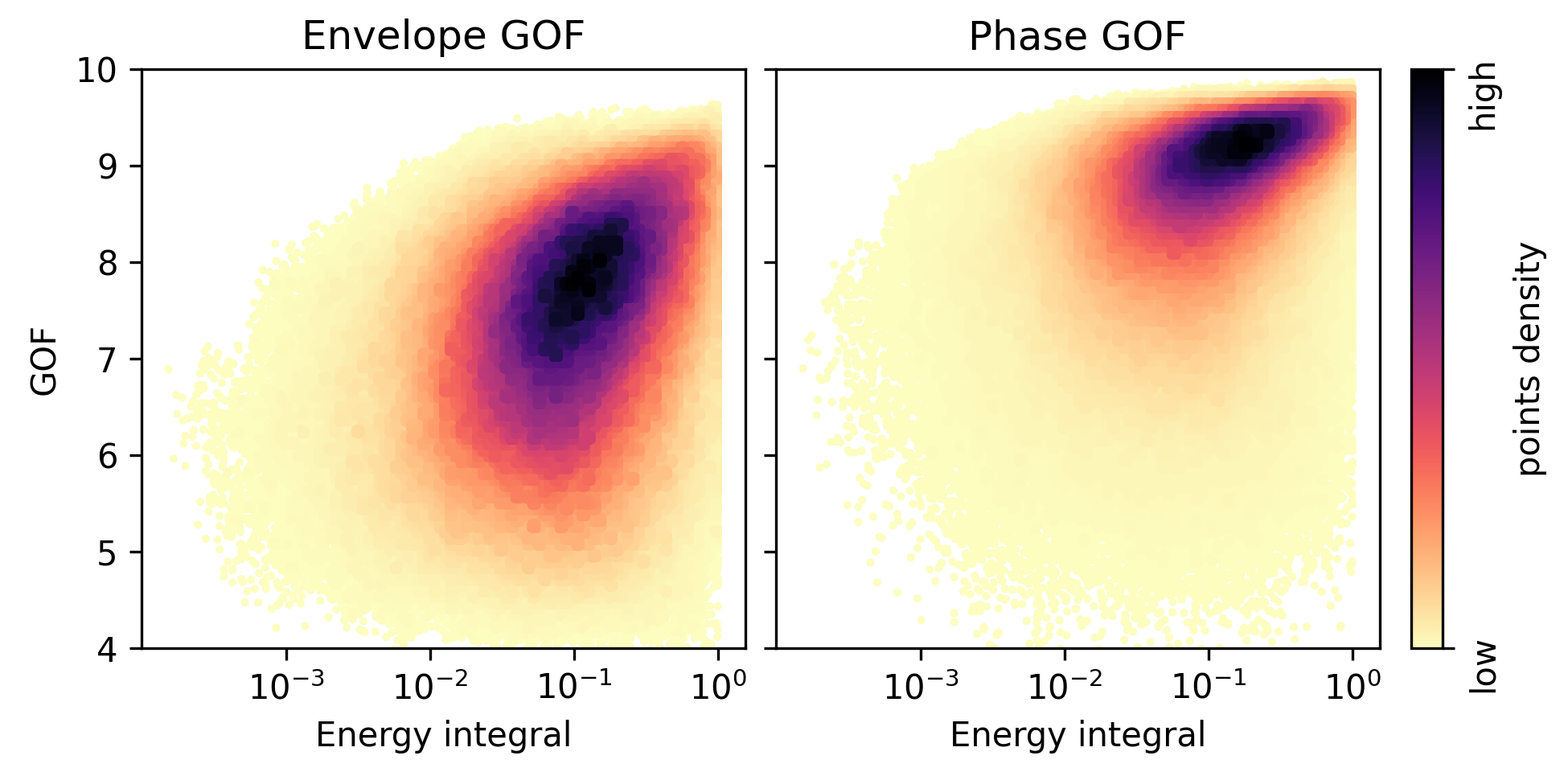}
		\caption{For 1000 samples and all sensors, the envelope GOF and phase GOF of each signal is shown against the energy integral. Color indicates the point density.}
		\label{fig:GOF_energy}
	\end{figure}

	Figure \ref{fig:GOF_energy} shows a positive correlation between the energy content $\widetilde{EI}$ of a signal and the GOF scores. In particular, this means that ground motion time series with the largest errors (i.e. the lowest GOF) generally do not contain much energy. For several practical applications such as seismic hazard analyses, errors on those low-energy signals cause less issues.

\section{Generalization ability}
\label{sec:generalization}
	Neural networks and neural operators are not meant to extrapolate data beyond the training distribution. However, when evaluating predictions on real data, one cannot guarantee that test data are perfectly in-distribution. Although predictions are not expected to be excellent in this situation, it is crucial that they remain reasonable. For the elastic wave equation, this especially means that predicted amplitudes should be of the right order of magnitude and wave arrival times should be close to the reference.

\subsection{Generalization to out-of-distribution sources}
\label{sec:sources_OOD}
	In the HEMEW\textsuperscript{S}-3D database, the propagation domain extends from \SI{0}{km} to \SI{9.6}{km} and the sources are located between \SI{1.2}{km} and \SI{8.4}{km}. However, since the source location is provided as a vector of coordinates, the MIFNO is not constrained to sources located inside the domain. Thus, in this section, we explore its ability to predict surface ground motion originating from a source outside the training domain. 
	
	To simulate the reference ground motion, we create 1000 new geologies that are twice larger than the original propagation domains and have the same depth. They are all heterogeneous geologies with parameters following the same distributions as the HEMEW\textsuperscript{S}-3D database (Tab. \ref{tab:stats_materials}). In these large geologies, sources are located randomly within $[-3.6\text{km}; 13.2\text{km}]$ and their orientation is also randomly chosen. The same normalization of the source position is applied on the training and out-of-distribution datasets, i.e. dividing the source coordinates by the domain length, $\Lambda$=\SI{9.6}{km}. While training sources were normalized to $[0, 1]^2$ by this procedure, the out-of-distribution sources were mapped to $[-0.5, 1.5]^2$. The reference ground motion is acquired only at the surface of the original propagation domain, i.e. $[0\text{km}; 9.6\text{km}]$.
	
	\begin{figure}[h]
		\centering
		\includegraphics[width=\textwidth]{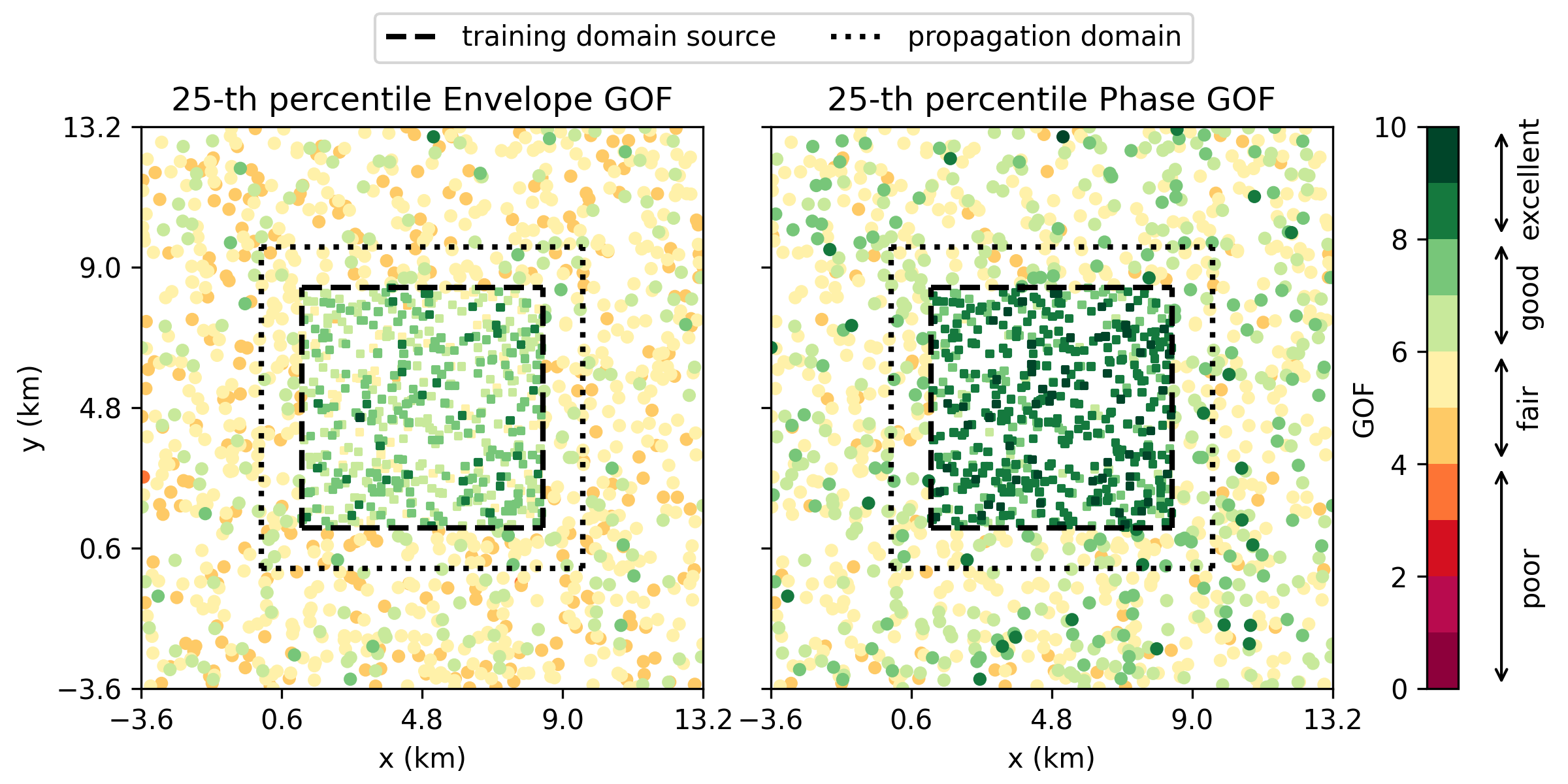}
		\caption{25-th percentile of envelope GOF (left) and phase GOF (right) for 1000 samples where the source is located outside the training domain (dashed square) or even outside the physical domain (dotted square). For reference, 1000 samples with a source inside the training domain are shown.}
		\label{fig:GOF_source_outside}
	\end{figure}
	
	Figure \ref{fig:GOF_source_outside} shows that the MIFNO maintains very good predictions when the source is located outside the training domain, both for the envelope GOF and the phase GOF. When the source remains inside the propagation domain (dotted square in Fig. \ref{fig:GOF_source_outside}), the inter-quartile range is \SI{6.24}{}-\SI{7.70}{} for the envelope GOF and \SI{7.41}{}-\SI{8.77}{} for the phase GOF. These good to excellent results are satisfying for out-of-distribution predictions since they are not too far from the reference results on in-distribution data (Tab. \ref{tab:metrics_MIFNO}). When the source locations are farther from the training domain, predictions worsen. This can be seen in Fig. \ref{fig:GOF_source_outside} for sources out of the training domain and it is perfectly expected since neural operators are not designed for severe extrapolation tasks. 
	
	It is worth noticing that the F-FNO produces results significantly worse than the MIFNO on this out-of-distribution task. Figure \ref{fig:timeseries_source_outside} shows that the wave arrival times are especially wrong with the F-FNO. Figure \ref{fig:GOF_source_outside_FFNO} confirms these findings since the phase GOFs are only \textit{fair} with the F-FNO. This confirms the benefits of the MIFNO architecture to fully apprehend the source characteristics.

\subsection{Generalization to out-of-distribution geologies}
\label{sec:out_of_distrib}
\begin{figure}[p]
		\centering
		\begin{subfigure}[b]{0.02\textwidth}
			(a)
		\end{subfigure}
		\hfill
		\begin{subfigure}[b]{0.43\textwidth}
			\centering
			\includegraphics[width=\textwidth]{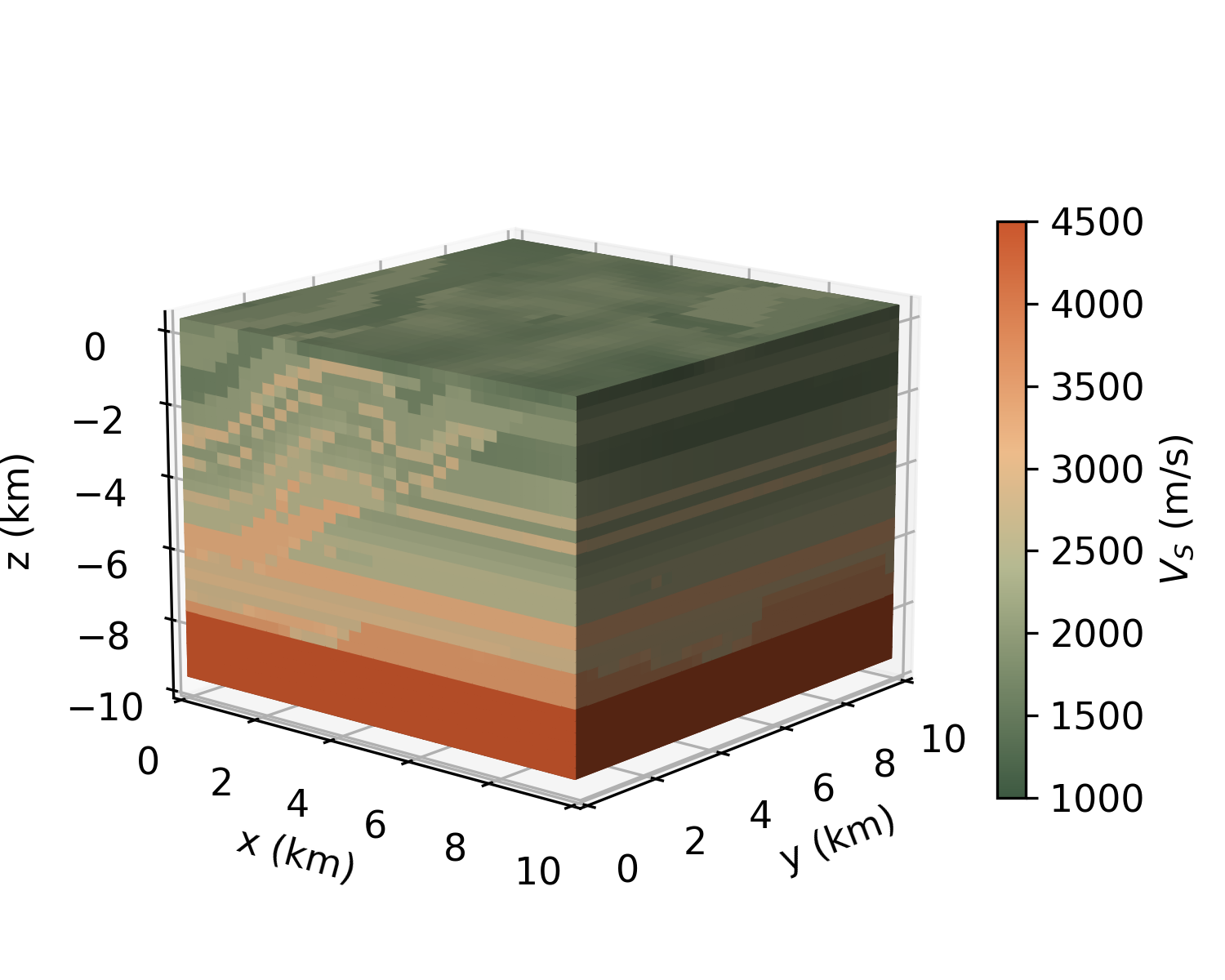}
		\end{subfigure}
		\hfill
		\begin{subfigure}[b]{0.02\textwidth}
			(b)
		\end{subfigure}
		\hfill
		\begin{subfigure}[b]{0.50\textwidth}
			\centering
			\includegraphics[width=0.9\textwidth]{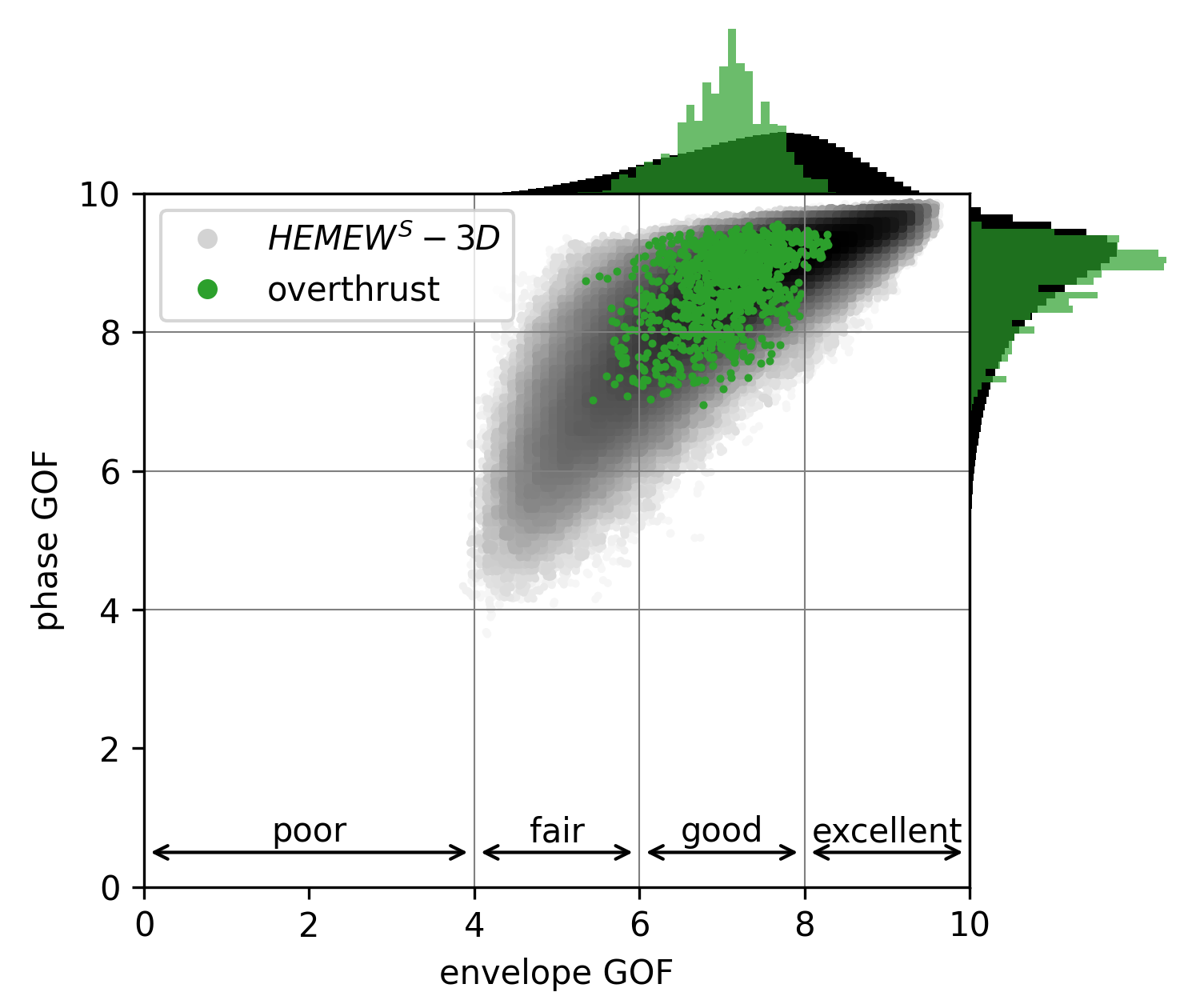}
		\end{subfigure}
		\hfill
		\begin{subfigure}[b]{0.02\textwidth}
			(c)
		\end{subfigure}
		\hfill
		\begin{subfigure}[b]{0.96\textwidth}
			\includegraphics[width=\textwidth]{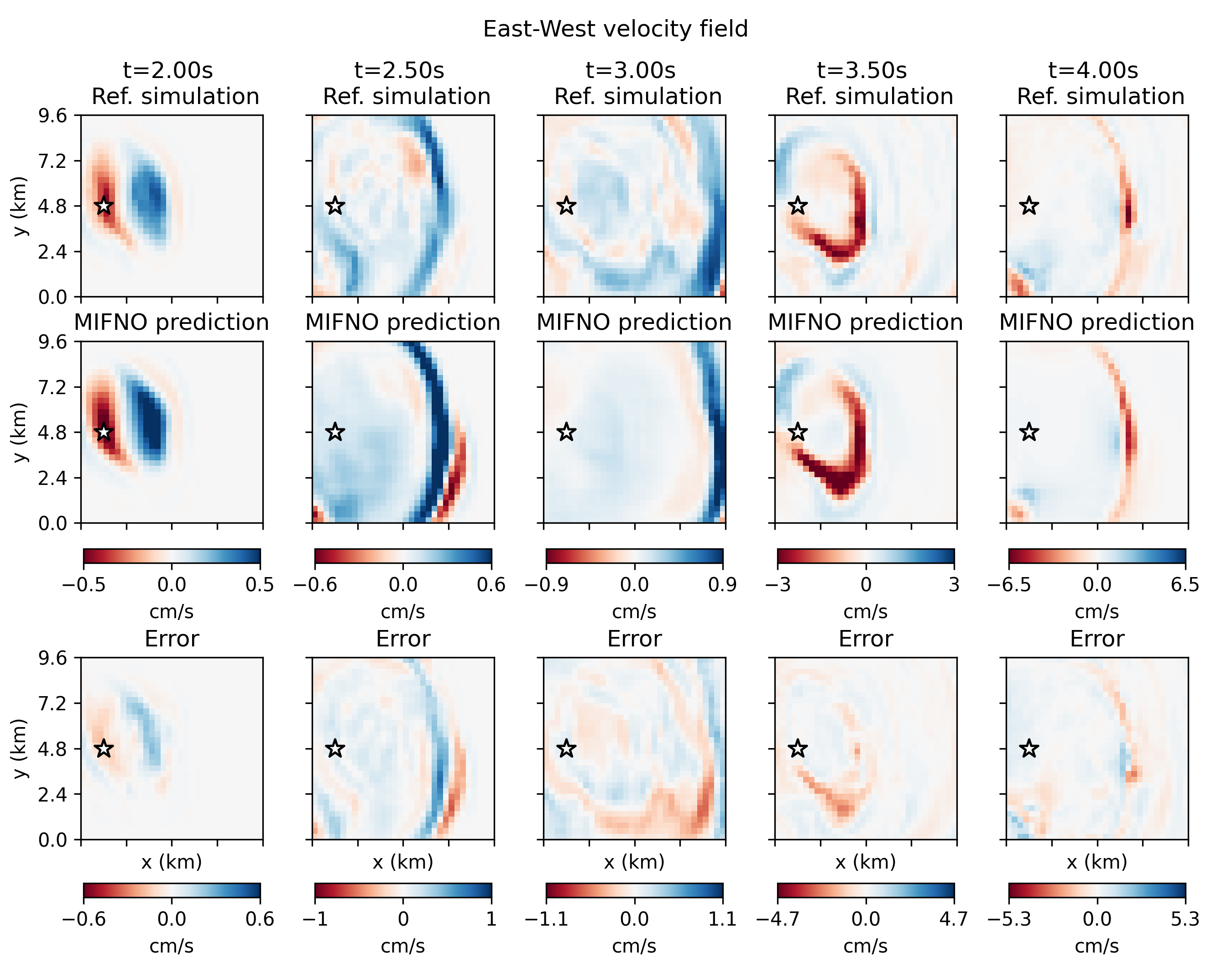}
		\end{subfigure}
		\hfill
		\caption{(a) One sample of an overthrust geology downscaled to \SI{300}{m}. (b) Distribution of the envelope and phase GOF for the HEMEW\textsuperscript{S}-3D database (grey, darker colors indicate higher points density) and for each surface sensor predicted with the overthrust geology (green). (c) East-West component of the simulated (upper row) and predicted (middle row) velocity fields at five time instants. The white star denotes the position of the source (depth=\SI{-9}{km}). The error between simulation and prediction is given in the lower row.}
		\label{fig:overthrust}
	\end{figure}

	To test the generalization ability of the MIFNO to out-of-distribution geologies, a 3D overthrust model was adapted to the size of our geological domain by dilating its depth to \SI{7.8}{km} \citep{aminzadeh3DSaltOverthrust1997}. This model is a 3D extension of the well-used Marmousi model. It shows a succession of thin layers with contrasting velocities and several folds. It is downscaled to a spatial resolution of \SI{300}{m} to match the resolution of the HEMEW\textsuperscript{S}-3D database (Fig. \ref{fig:overthrust}a). 1000 geological models were sampled from the original \SI{20.03 x 20.03 x 7.8}{km} model and sources were randomly chosen inside each domain, following the same distribution as the HEMEW\textsuperscript{S}-3D database.
	
	Figure \ref{fig:overthrust} shows the MIFNO predictions for one sample. The predicted wavefronts are close to the simulations, in all parts of the domain. In particular, the wave arrival times are accurately predicted and the influence of the source orientation is well preserved since the phases correspond to the reference. The amplitude of the main fluctuations is also accurate. The P wave peaks tend to be overestimated (see $t$=\SI{2.50}{s} in Fig. \ref{fig:overthrust}c) but they have a small amplitude on the East-West component. The focus is instead on the S waves, for which the amplitudes agree well with the reference simulation (Fig. \ref{fig:line_timeseries_overthrust}). 
	
	The envelope GOFs are good for a large majority of sensors and phase GOFs are very good to excellent (Fig. \ref{fig:overthrust}b). The GOF distributions are concentrated around the mean of the GOFs obtained on the HEMEW\textsuperscript{S}-3D database, thereby showing the generalization ability of the MIFNO to real complex geologies.
	Although Fig. \ref{fig:overthrust}b shows a significant GOF variability between sensors for this out-of-distribution geology (around 2 GOF units), Fig. \ref{fig:gof_overthrust} illustrates that GOFs below 6 are concentrated on small areas of the domain and GOFs remain good or excellent on most of the surface. A similar variability was observed on the HEMEW-3D database when geological fluctuations create complex wavefields patterns in localized regions.
	
	Finally, one can notice the absence of late fluctuations in the predictions (mainly visible for low amplitudes around $t$=\SI{3}{s}). This was expected from our previous analyses since these ground motion fluctuations are created by the multiple wave reflections and refractions on the thin geological layers and they are difficult to predict. 
	
	\begin{table}[h]
		\footnotesize
		\begin{center}
			Overthrust samples
			\begin{tabular}{|c|c|c|c|c|c|c|c|}
				\hline
				Model & rMAE & rRMSE & rFFT\textsubscript{low} & rFFT\textsubscript{mid} & rFFT\textsubscript{high} & EG & PG \\
				\hline
				MIFNO & 0.16 $\pm$ 0.03 & 0.24 $\pm$ 0.03 & -0.19 $\pm$ 0.18 & -0.29 $\pm$ 0.17 & -0.37 $\pm$ 0.16 & 7.39 $\pm$ 0.45 & 8.74 $\pm$ 0.26 \\
				\hline
			\end{tabular}
		\end{center}
		\caption{Mean and standard deviation of the metrics computed on 1000 overthrust samples. rMAE: relative MAE (0 is best), rRMSE: relative RMSE (0 is best), rFFT\textsubscript{low}: relative frequency bias 0-1Hz (0 is best), rFFT\textsubscript{mid}: relative frequency bias 1-2Hz (0 is best), rFFT\textsubscript{high}: relative frequency bias 2-5Hz (0 is best), EG: envelope GOF (10 is best), PG: phase GOF (10 is best). For frequency biases, negative values indicate underestimation.}
		\label{tab:metrics_MIFNO_overthrust}
	\end{table}

	Table \ref{tab:metrics_MIFNO_overthrust} quantifies the prediction accuracy on the 1000 samples with an overthrust geology and confirms that the MIFNO produces meaningful results for a large majority of overthrust samples. Metrics are close to the results previously obtained on the test dataset. Envelope GOFs remain good and phase GOFs excellent, with only \SI{0.1}{} difference compared to the test dataset.

\subsection{Generalization to higher resolution}
	Although zero-shot super resolution can always be technically performed with FNOs, FNOs are not invariant with respect to the resolution in the general framework \citep{bartolucciAreNeuralOperators2023}. This means that high-resolution predictions may be less accurate than predictions at the training resolution. Since there is no theoretical result for FNOs including source and PDE parameter, we investigate zero-shot super resolution experimentally with the MIFNO. While the MIFNO was trained with inputs of spatial resolution of \SI{32 x 32 x 32}{}, geologies are interpolated to \SI{48 x 48 x 32}{} and \SI{64 x 64 x 32}{} to obtain high-resolution predictions. The vertical dimension is preserved to maintain the same depth-to-time conversion. Since the source is given with absolute coordinates, there is no need to modify the \textit{source branch} when modifying the resolution. It is important to notice that the high-resolution wavefields are obtained by running simulations with a grid of \SI{48 x 48}{} (resp. \SI{64 x 64}{}) sensors. They are not an interpolation of the original wavefields. Therefore, they exhibit small-scale features that are absent from the training wavefields.
	
	\begin{figure}[p]
		\centering
		\includegraphics[width=\textwidth]{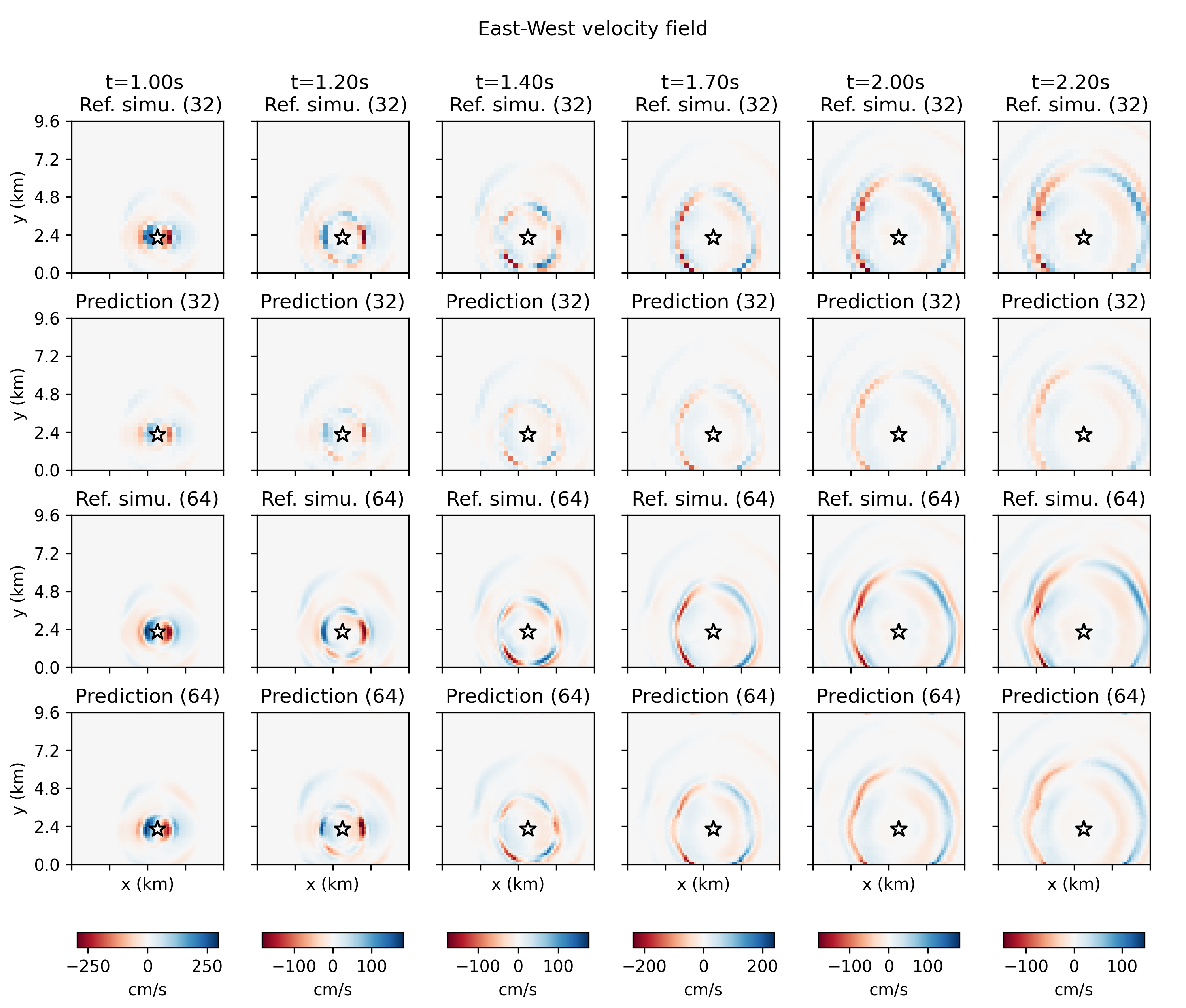}
		\caption{East-West component of the velocity wavefield at five time instants. first row: simulation with \SI{32 x 32}{} sensors. second row: prediction with \SI{32 x 32}{} sensors. third row: simulation with \SI{64 x 64}{} sensors. fourth row: prediction with \SI{64 x 64}{} sensors. The white star denotes the position of the source (depth=\SI{-1.3}{km}).}
		\label{fig:snapshots_res64}
	\end{figure}

	Figure \ref{fig:snapshots_res64} shows that it is possible to obtain accurate velocity wavefields with a resolution of 64 that has not been seen during training. The wavefronts are better defined with the increased resolution, which allows to capture peaks more accurately in this example (see snapshots at t=\SI{1.40}{s} in Fig. \ref{fig:snapshots_res64} for instance). However, some inaccuracies are visible at higher resolution, especially on the boundaries. These artefacts can be explained by the fact that high-resolution predictions are in advance at the edge of the domain, which induces a phase difference between the prediction and the reference (Fig. \ref{fig:line_timeseries_res64}).

	\begin{table}[h]
		\footnotesize
		\begin{center}
			\begin{tabular}{|c|c|c|c|c|c|c|c|}
				\hline
				Resolution & rMAE & rRMSE & rFFT\textsubscript{low} & rFFT\textsubscript{mid} & rFFT\textsubscript{high} & EG & PG \\
				\hline
				32 & 0.13 $\pm$ 0.05 & 0.2 $\pm$ 0.07 & -0.21 $\pm$ 0.17 & -0.29 $\pm$ 0.19 & -0.37 $\pm$ 0.21 & 7.34 $\pm$ 0.75 & 8.63 $\pm$ 0.54 \\
				48 & 0.17 $\pm$ 0.08 & 0.28 $\pm$ 0.09 & -0.39 $\pm$ 0.32 & -0.48 $\pm$ 0.31 & -0.55 $\pm$ 0.3 & 6.46 $\pm$ 0.74 & 7.79 $\pm$ 0.48 \\
				64 & 0.19 $\pm$ 0.09 & 0.28 $\pm$ 0.1 & -0.52 $\pm$ 0.32 & -0.6 $\pm$ 0.3 & -0.65 $\pm$ 0.29 & 6.15 $\pm$ 0.74 & 7.95 $\pm$ 0.47 \\
				\hline
			\end{tabular}
		\end{center}
		\caption{Mean and standard deviation of the metrics computed on 1000 test samples. rRMSE: relative RMSE (0 is best), rFFT\textsubscript{low}: relative frequency bias 0-1Hz (0 is best), rFFT\textsubscript{mid}: relative frequency bias 1-2Hz (0 is best), rFFT\textsubscript{high}: relative frequency bias 2-5Hz (0 is best), EG: envelope Goodness-of-Fit (10 is best), PG: phase Goodness-of-Fit (10 is best). For frequency biases, negative values indicate underestimation.}
		\label{tab:metrics_MIFNO_resolution}
	\end{table}
	
	Quantitatively, Tab. \ref{tab:metrics_MIFNO_resolution} shows that results remain reasonable when the resolution increases, even though metrics degrade. Envelope GOFs are still good and half of the predictions have excellent phase GOFs on a set of 1000 high-resolution samples. This shows that the physics of wave propagation has been correctly learnt, especially the dependency between the source position and the wave arrival times. However, the training resolution of 32 is probably too coarse to transfer all fine details to higher resolutions.

\section{Transfer learning on sources and geologies}
\label{sec:transfer_learning}
	When focusing on a region of interest, it is common to have prior knowledge of the local geology and/or the type of sources that can generate earthquakes. Then, transfer learning is a well-used approach to improve the MIFNO accuracy for this specific setting. With transfer learning, one designs a small database tailored to the region under study and the MIFNO is fine-tuned on these data. Since one benefits from the knowledge acquired on the general database, transfer learning involves limited simulation and training costs. Our case study is chosen as the Le Teil earthquake (Mw4.9, France, 2019), the most severe earthquake in metropolitan France in the last decade.

\subsection{A database for the Le Teil earthquake}
	Based on a local 1D geological model, we designed a database of \SI{4000}{} geologies tailored to the Le Teil region. The 1D model contains six homogeneous horizontal layers (\cite{smerziniRegionalPhysicsbasedSimulation2023}, Tab. \ref{tab:velocity_LeTeil}). To create realistic variations of this geology, heterogeneities were added as random fields inside each layer. Random fields follow the same parameter distributions as the HEMEW\textsuperscript{S}-3D database (Tab. \ref{tab:stats_materials}). 

\begin{figure}[h]
		\centering
		\includegraphics[width=0.6\textwidth, trim=0 0 0 40, clip=true]{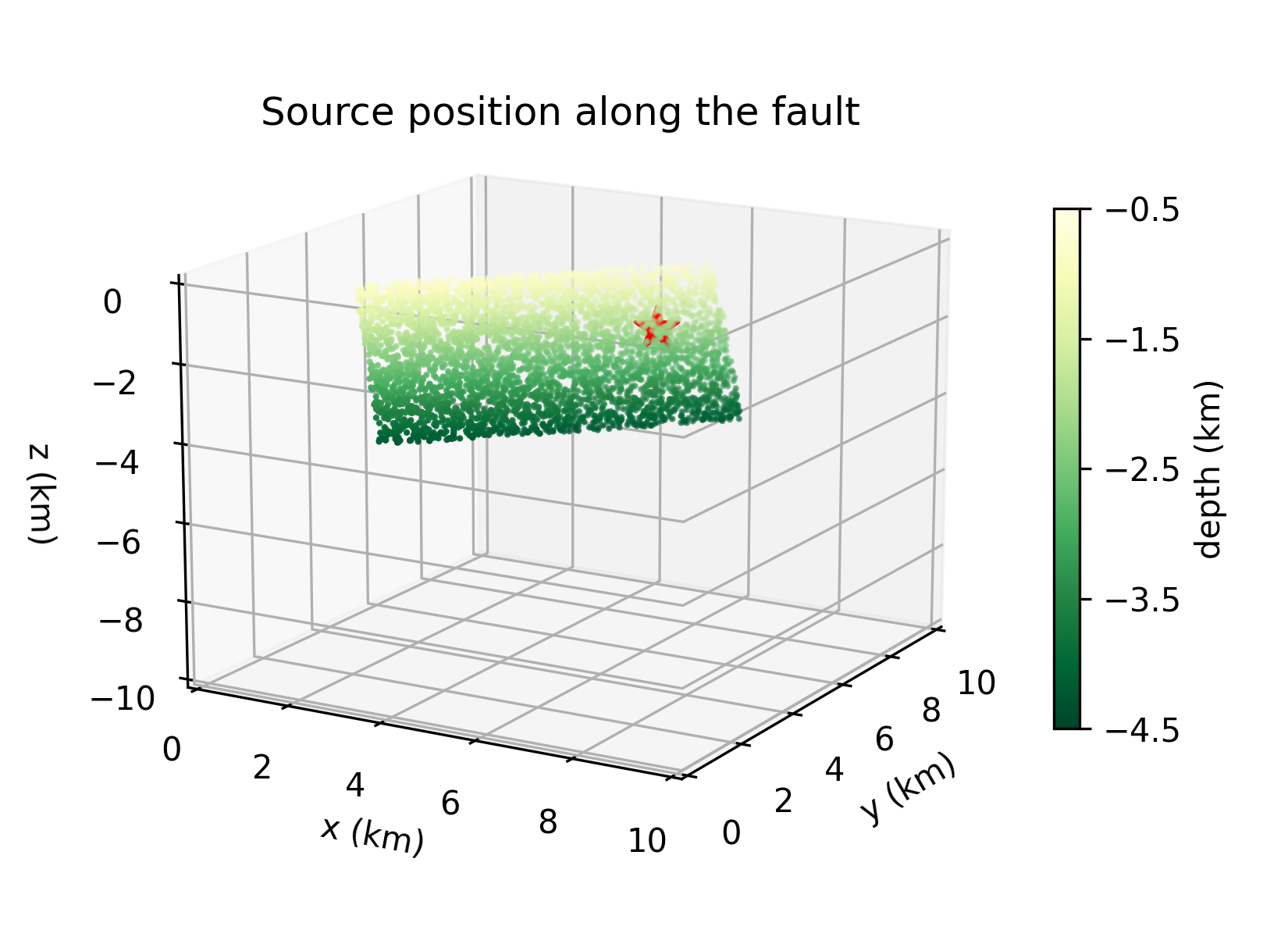}
		\caption{Position of the 4000 sources in the Le Teil database, colored by their depth. The red star denotes the source that generated the ground motion in Fig. \ref{fig:LeTeil_timeseries}}
		\label{fig:LeTeil_sources}
	\end{figure}

	Several studies estimated the source parameters of the Le Teil earthquake, both in the form of a fault plane and an equivalent point-source. In our Le Teil database, we choose the source locations close to the fault plane obtained by \cite{vallageMultitechnologyCharacterizationUnusual2021}. The fault is \SI{7}{km}-wide, \SI{4}{km}-high and extends between \SI{-780}{m} and \SI{-4210}{m}. Following a latin hypercube sampling, \SI{4000}{} point sources were randomly placed along the fault plane, with a normal distance to the fault plane up to \SI{100}{m} to account for uncertainties in the fault localization (Fig. \ref{fig:LeTeil_sources}).

	Each point source is assigned a random moment tensor, whose angles are chosen from the moment tensor inversion conducted by \cite{delouisConstrainingPointSource2021}: strike between \ang{30} and \ang{70}, dip between \ang{20} and \ang{70}, rake between \ang{70} and \ang{120}. Similarly to the HEMEW\textsuperscript{S}-3D database, \SI{6.4}{s} of ground motion were generated at the \SI{32 x 32}{} surface sensors. 
	
	Within this framework, we obtained \SI{4000}{} samples (geology, source, ground motion wavefields) tailored to the Le Teil earthquake. Data are considered in-distribution since parameters are within the range of the HEMEW\textsuperscript{S}-3D database but vary in smaller intervals. 	
	Among the \SI{4 000}{} samples in the Le Teil database, up to \SI{3 000}{} were used for training, \SI{300}{} for validation, and \SI{700}{} for testing.

\subsection{Transfer learning predictions}
\begin{figure}[h]
		\centering
		\includegraphics[width=0.8\textwidth]{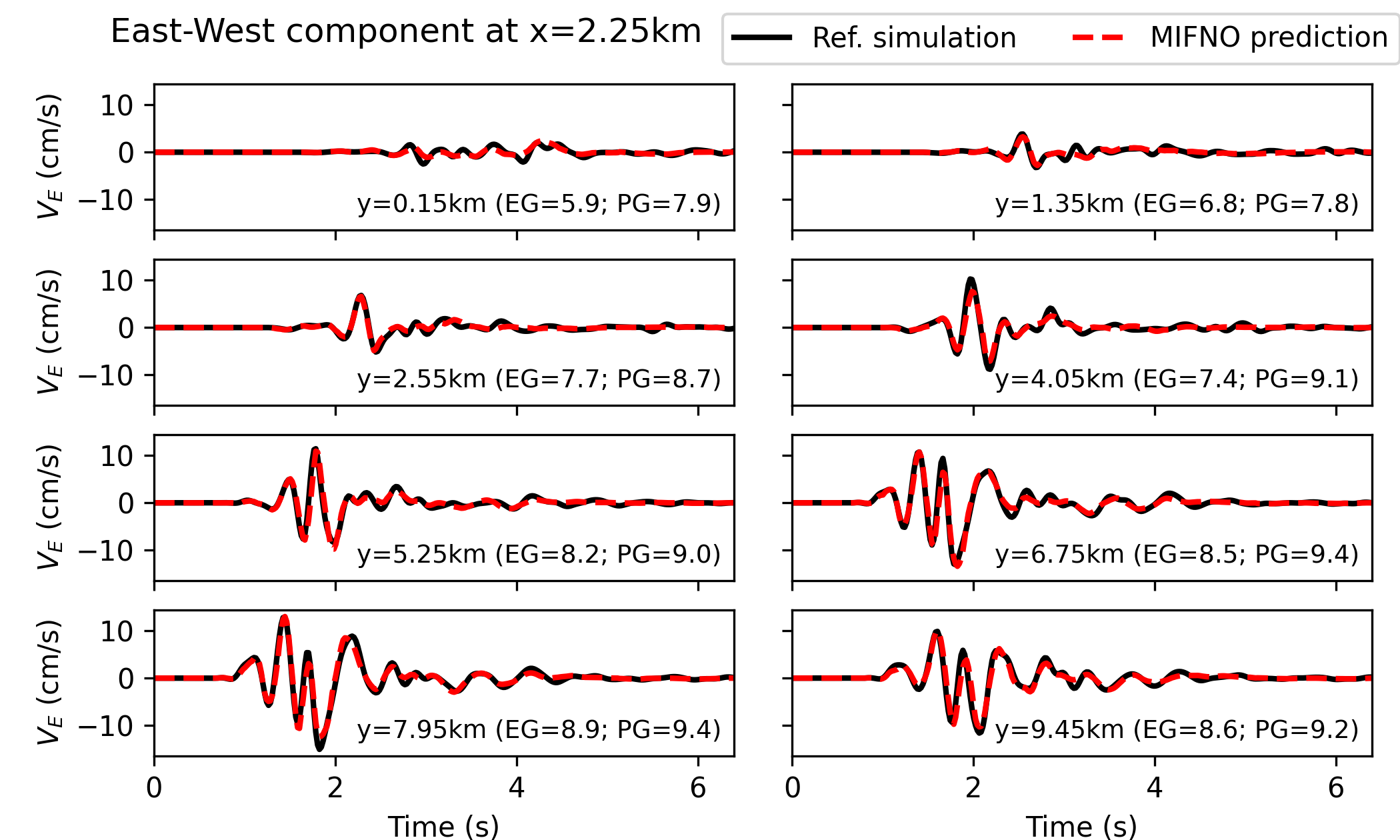}
		\caption{East-West velocity time series at 8 sensors aligned at $x$=\SI{2.25}{km} for a test sample. Predictions (red dashed lines) were obtained with the MIFNO fine-tuned on 2000 samples. The reference ground motion is given by the numerical simulation (black lines). For each sensor, its position along the $y$ axis is indicated, and the envelope GOF (EG) and phase GOF (PG) are also reported.}
		\label{fig:LeTeil_timeseries}
	\end{figure}

	Figure \ref{fig:LeTeil_timeseries} illustrates the velocity time series obtained for a test sample in the Le Teil database. Predictions are very close to the simulated ground truth for all wave arrivals and amplitudes. One can especially notice the numerous high fluctuations that are very well captured at all sensors. The MIFNO is also able to reproduce the late small-scale fluctuations (between \SI{3}{s} and \SI{5}{s}, Fig. \ref{fig:LeTeil_timeseries}) thanks to the transfer learning. Phase GOFs are excellent or close to excellent for all sensors, while envelope GOFs are excellent for the sensors with the largest fluctuations (two bottom rows in Fig. \ref{fig:LeTeil_timeseries}) and good for the other sensors. These metrics confirm the excellent visual agreement. 	
	
	Quantitatively, 700 predictions in the test dataset were assessed with the relative RMSE, the frequency biases and the GOFs. With only 100 transfer learning samples, the GOFs already improve by 1 to 2 units, compared to the predictions with the generic MIFNO (i.e. without transfer learning, line N\textsubscript{TL}=0 in Tab. \ref{tab:metrics_MIFNO_LeTeil}). This illustrates the considerable benefits of transfer learning, even with a limited number of data. After even a light transfer learning (N\textsubscript{TL}=100), \SI{68}{\%} predictions reach a good envelope GOF and \SI{61}{\%} an excellent phase GOF. 
	
	\begin{table}[h]
		\footnotesize
		\begin{center}
			Le Teil database with a random source along the fault plane
			\begin{tabular}{|l|c|c|c|c|c|c|c|}
				\hline
				\# samples & rMAE & rRMSE & rFFT\textsubscript{low} & rFFT\textsubscript{mid} & rFFT\textsubscript{high} & EG & PG \\
				\hline
				N\textsubscript{train}=3000 & 0.17 $\pm$ 0.06 & 0.25 $\pm$ 0.07 & -0.33 $\pm$ 0.23 & -0.48 $\pm$ 0.22 & -0.6 $\pm$ 0.21 & 6.45 $\pm$ 0.88 & 7.98 $\pm$ 0.7 \\
				\hline
				N\textsubscript{TL}=0 & 0.21 $\pm$ 0.04 & 0.34 $\pm$ 0.05 & -0.21 $\pm$ 0.41 & -0.37 $\pm$ 0.46 & -0.41 $\pm$ 0.45 & 5.9 $\pm$ 0.7 & 6.78 $\pm$ 0.53 \\
				N\textsubscript{TL}=100 & 0.17 $\pm$ 0.06 & 0.25 $\pm$ 0.07 & -0.15 $\pm$ 0.32 & -0.32 $\pm$ 0.28 & -0.45 $\pm$ 0.25 & 6.68 $\pm$ 0.89 & 8.14 $\pm$ 0.6 \\
				N\textsubscript{TL}=250 & 0.16 $\pm$ 0.06 & 0.24 $\pm$ 0.07 & -0.12 $\pm$ 0.35 & -0.28 $\pm$ 0.3 & -0.4 $\pm$ 0.26 & 6.84 $\pm$ 0.97 & 8.32 $\pm$ 0.59 \\
				N\textsubscript{TL}=500 & 0.16 $\pm$ 0.06 & 0.23 $\pm$ 0.07 & -0.11 $\pm$ 0.3 & -0.25 $\pm$ 0.27 & -0.36 $\pm$ 0.25 & 7.0 $\pm$ 0.96 & 8.46 $\pm$ 0.59 \\
				N\textsubscript{TL}=1000 & 0.15 $\pm$ 0.06 & 0.22 $\pm$ 0.07 & -0.13 $\pm$ 0.28 & -0.26 $\pm$ 0.26 & -0.36 $\pm$ 0.24 & 7.11 $\pm$ 0.96 & 8.55 $\pm$ 0.59 \\
				N\textsubscript{TL}=2000 & 0.15 $\pm$ 0.06 & 0.21 $\pm$ 0.07 & -0.14 $\pm$ 0.26 & -0.26 $\pm$ 0.25 & -0.35 $\pm$ 0.24 & 7.2 $\pm$ 0.96 & 8.66 $\pm$ 0.57 \\
				N\textsubscript{TL}=3000 & 0.14 $\pm$ 0.06 & 0.21 $\pm$ 0.07 & -0.11 $\pm$ 0.27 & -0.23 $\pm$ 0.25 & -0.32 $\pm$ 0.24 & 7.26 $\pm$ 0.95 & 8.71 $\pm$ 0.56 \\
				\hline
			\end{tabular}
		\end{center}
		\caption{Mean and standard deviation} of the metrics computed on 700 test samples specific to the Le Teil region. (upper row): training with only 3000 specific data. In other experiments, transfer learning was used with 100 to 3000 samples (N\textsubscript{TL} = number of transfer learning samples). rRMSE: relative RMSE (0 is best), rFFT\textsubscript{low}: relative frequency bias 0-1Hz (0 is best), rFFT\textsubscript{mid}: relative frequency bias 1-2Hz (0 is best), rFFT\textsubscript{high}: relative frequency bias 2-5Hz (0 is best), EG: envelope Goodness-of-Fit (10 is best), PG: phase Goodness-of-Fit (10 is best). For frequency biases, negative values indicate underestimation.
		\label{tab:metrics_MIFNO_LeTeil}
	\end{table}

	For comparison purposes, the MIFNO was trained solely on 3000 samples from the Le Teil database (line N\textsubscript{TL}=\SI{3000}{} in Tab. \ref{tab:metrics_MIFNO_LeTeil}). However, the predictions accuracy was worse than the predictions obtained with only 100 transfer learning samples. This shows the major benefits of initializing the weights with the pre-trained MIFNO. 
	
	Table \ref{tab:metrics_MIFNO_LeTeil} also shows that the metrics continuously improve when the number of transfer learning samples increases. The high frequency bias improves by \SI{29}{\%} between N\textsubscript{TL}=100 and N\textsubscript{TL}=3000. The GOFs also benefit from larger transfer learning databases and the improvement is mainly visible on the first quartile, meaning that good predictions with N\textsubscript{TL}=100 tend towards excellent with N\textsubscript{TL}=\SI{3000}. This is illustrated by comparing the MIFNO predictions obtained with 2000 transfer learning samples in Fig. \ref{fig:LeTeil_timeseries} and 500 samples in Fig. \ref{fig:LeTeil_timeseries_TL500}. While the main fluctuations are already accurate with 500 transfer learning samples, increasing the number of samples allows to better capture the late fluctuations with a small amplitude.

\section{Conclusion}
\label{sec:conclusion}

	We introduced a Multiple-Input Fourier Neural Operator (MIFNO) that takes as input a 3D geological domain and a vector describing the source position and orientation. The structured geology is processed with factorized Fourier layers while the source parameters are transformed via convolutional layers while maintaining resolution invariance properties. The MIFNO was trained on \SI{30 000}{} earthquake simulations from the HEMEW\textsuperscript{S}-3D database covering a large variety of heterogeneous geological models, source positions and source orientations. It predicts 3-component surface ground motion between \SI{0}{s} and \SI{6.4}{s} for frequencies up to \SI{5}{Hz}. 
	
	The MIFNO performs on-par with the reference F-FNO model, as was verified on three datasets: i) fixed sources with fixed orientation, ii) randomly located sources with a fixed orientation, iii) random sources with random orientations. This shows that the flexibility offered by our proposed architecture preserves the accuracy of the F-FNO.
	
	MIFNO predictions are considered as good to excellent for a vast majority of sensors, following the common understanding of Goodness-Of-Fit (GOF) criteria. GOFs are better for the phase than for the envelope, which indicates that the MIFNO is well suited to predict wave arrival times and the spatial propagation of wavefronts. With \SI{87}{\%} of predictions having a good envelope GOF, the main peaks are correctly predicted, which is of crucial importance for seismological applications. Complex wave propagation phenomena due to heterogeneities yield small-scale fluctuations with small amplitude that are difficult to predict accurately, thereby hindering the envelope GOF. However, signals with a high energy content tend to have a better accuracy, which is of paramount importance for many real-life applications.
	
	The difficulties related to small-scale fluctuations are exacerbated by the design choices of the HEMEW\textsuperscript{S}-3D database that purposely contains geologies with strong heterogeneities (mean coefficient of variation of \SI{20}{\%}). However, such large coefficients of variations are rarely found in reality for large correlation lengths. They are included in the HEMEW\textsuperscript{S}-3D database to enhance its variability in a Scientific Machine Learning perspective. Therefore, predictions on common 1D velocity models augmented by moderate heterogeneities will fall on the most accurate range of the MIFNO predictions. 
	
	We found that the MIFNO tends to underestimate the frequency content of the ground motion fluctuations. These fluctuations are created by the seismic wave reflections and refractions on geological heterogeneities. They are stronger in the high-frequency components of the signals and they are also more present when ground motion originates from deep sources. This explains why we found that the frequency underestimation is more pronounced in high-frequency components and for deep sources.
	
	Using a more complex MIFNO (i.e. with more layers and more channels) is a privileged approach to increase the accuracy, as illustrated by the difference between the 8-layer and 16-layer MIFNO, as well as the influence of the number of channels. This confirms previous investigations on the F-FNO, which also showed that this architecture scales well when the number of parameters increases \citep{lehmann3DElasticWave2024}. Leveraging rotation invariance to increase the size of the dataset yields more samples to adequately train deeper models. As a consequence, the MIFNO becomes more accurate.
	
	The MIFNO is also able to generalize well to different out-of-distribution data. The source was located out of the training domain with only a moderate degradation of the accuracy. Although it could also be placed outside the propagation domain, this large distribution shift worsened the predictions. The MIFNO shows a significant advantage over the F-FNO on out-of-distribution sources, thereby emphasizing the benefits of the specific \textit{source branch}.
	In addition, the MIFNO can be applied to geologies that have a different resolution than the training resolution. Although the average metrics degrade with increased resolution, the spatial propagation of wavefields was preserved and the predictions of peaks amplitudes could improve at higher resolution in some cases. These results show the robustness of the MIFNO when applied on out-of-distribution data, which is crucial to move towards real-case applications. 
	
	Generalization to geologies that are far from the training database is challenging but the MIFNO predicts accurate ground motion for a set of overthrust geological models with thin folded layers. There is a good visual agreement between the main predicted wavefronts and the simulations, and the GOF distributions with the overthrust geologies follow the distributions as the HEMEW\textsuperscript{S}-3D database. This illustrates the good generalization ability of the MIFNO for real complex geologies. However, the thin layers in the overthrust geology create many wave reflections, which are challenging to predict with a generic surrogate model. 
	
	When aiming for more accurate predictions in complex settings, transfer learning should be used to specialize the MIFNO on a smaller database tailored to the region under study. This approach was adopted for the Le Teil earthquake and it showed that only 250 transfer learning samples lead to good accuracy. The fine-tuned MIFNO surpasses predictions of the generic MIFNO and a MIFNO trained solely on the Le Teil database. When increasing the number of transfer learning samples, predictions reach excellent accuracy and reproduce even the small-amplitude fluctuations.
	
	In conclusion, the MIFNO is the first surrogate model providing the flexibility of an earthquake emulator while reducing significantly the computational costs. Once duly trained, MIFNO predictions provide \SI{6.4}{s} of ground motion in around \SI{10}{ms} (with one GPU) while numerical simulations take around \SI{43}{min} (with 32 CPUs). This massive speed-up paves the way for 3D uncertainty quantification analysis, design optimization, and inverse problems that are beyond reach with traditional methods. 
	
	We believe that the methodology presented in this work can be applied to other challenges related to propagation phenomena. Indeed, most wave Partial Differential Equations (PDEs) involve a space-dependent parameter describing the material properties and source parameters. Since both types of parameters are handled by the MIFNO, it can certainly benefit several scientific communities.

\clearpage
	
\appendix
\renewcommand{\thefigure}{S\arabic{figure}}
\renewcommand{\thetable}{S\arabic{table}}
\setcounter{figure}{0}
\setcounter{table}{0}

\section{Supplementary material}
\subsection{Supplementary material for prediction results}
	\begin{figure}[h]
		\centering
		\includegraphics[width=0.6\textwidth]{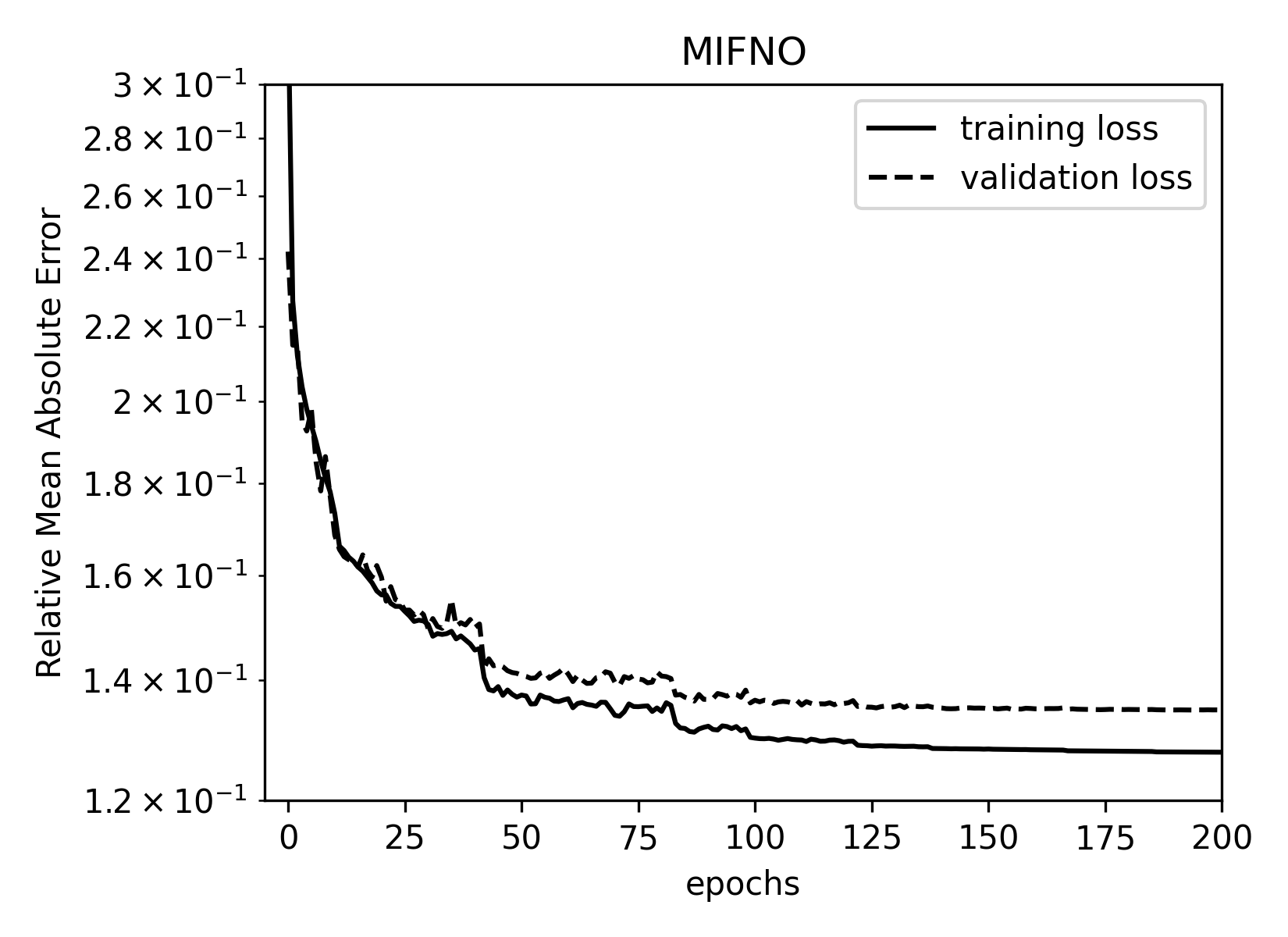}
		\caption{Evolution of the training loss (solid line) and validation loss (dashed line) as a function of the number of epochs.}
		\label{fig:loss_history}
	\end{figure}

	\begin{figure}[h]
		\centering
		\includegraphics[width=\textwidth]{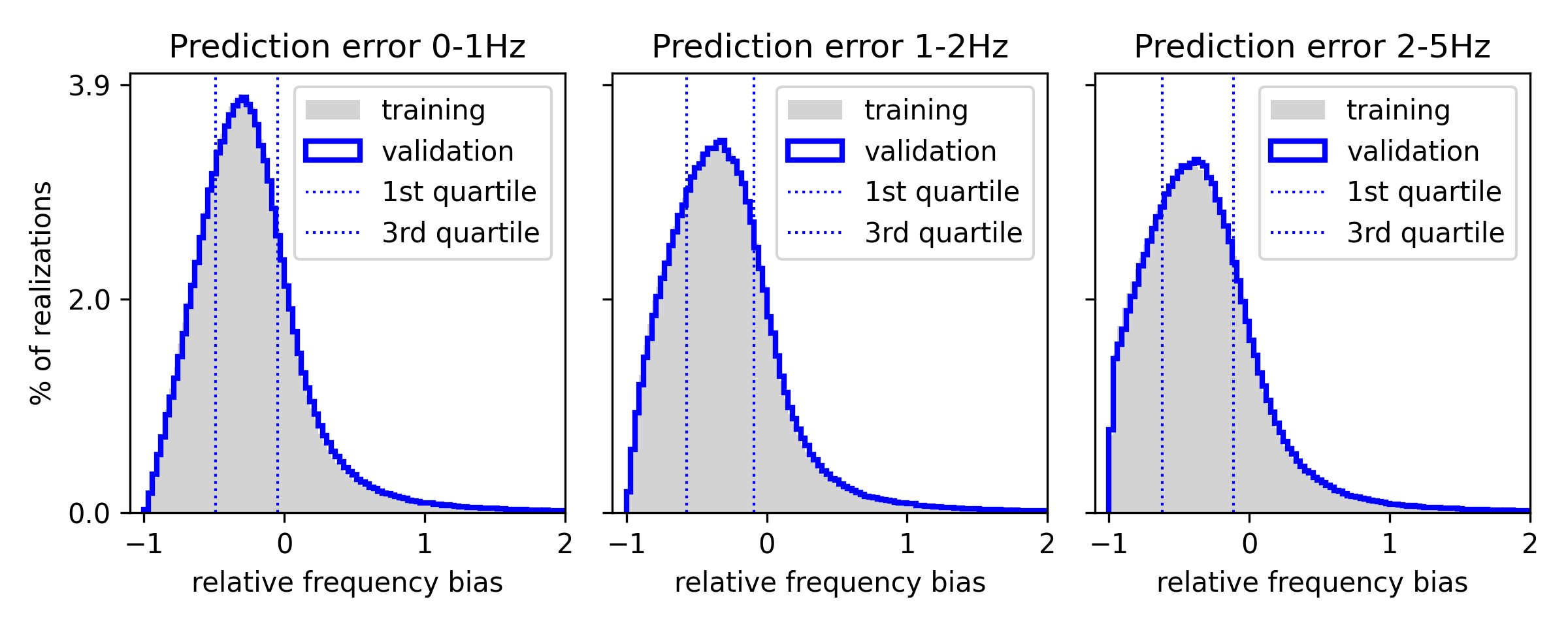}
		\caption{Distribution of the frequency bias for each sensor and each sample in the training (grey area) and validation (blue line) dataset.}
		\label{fig:error_distrib}
	\end{figure}

	\begin{figure}[h]
		\centering
		\includegraphics[width=\textwidth]{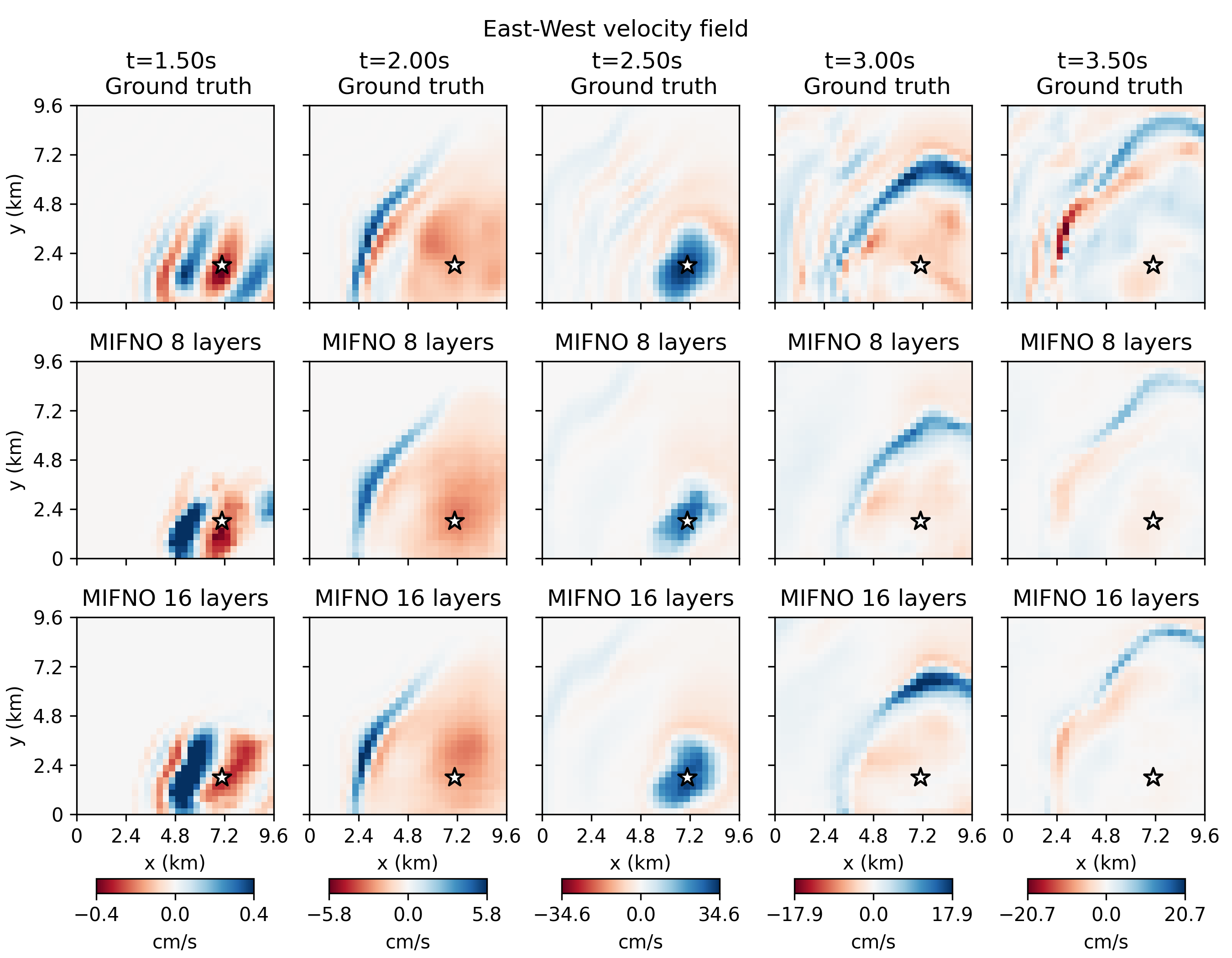}
		\caption{East-West component of the velocity fields from the reference simulation (upper row), prediction with a 8-layer MIFNO (middle row), and prediction with a 16-layer MIFNO (lower row). The white star denotes the epicenter.}
		\label{fig:snapshots_8_16layers}
	\end{figure}

	\begin{figure}[h]
		\centering
		\includegraphics[width=0.48\textwidth]{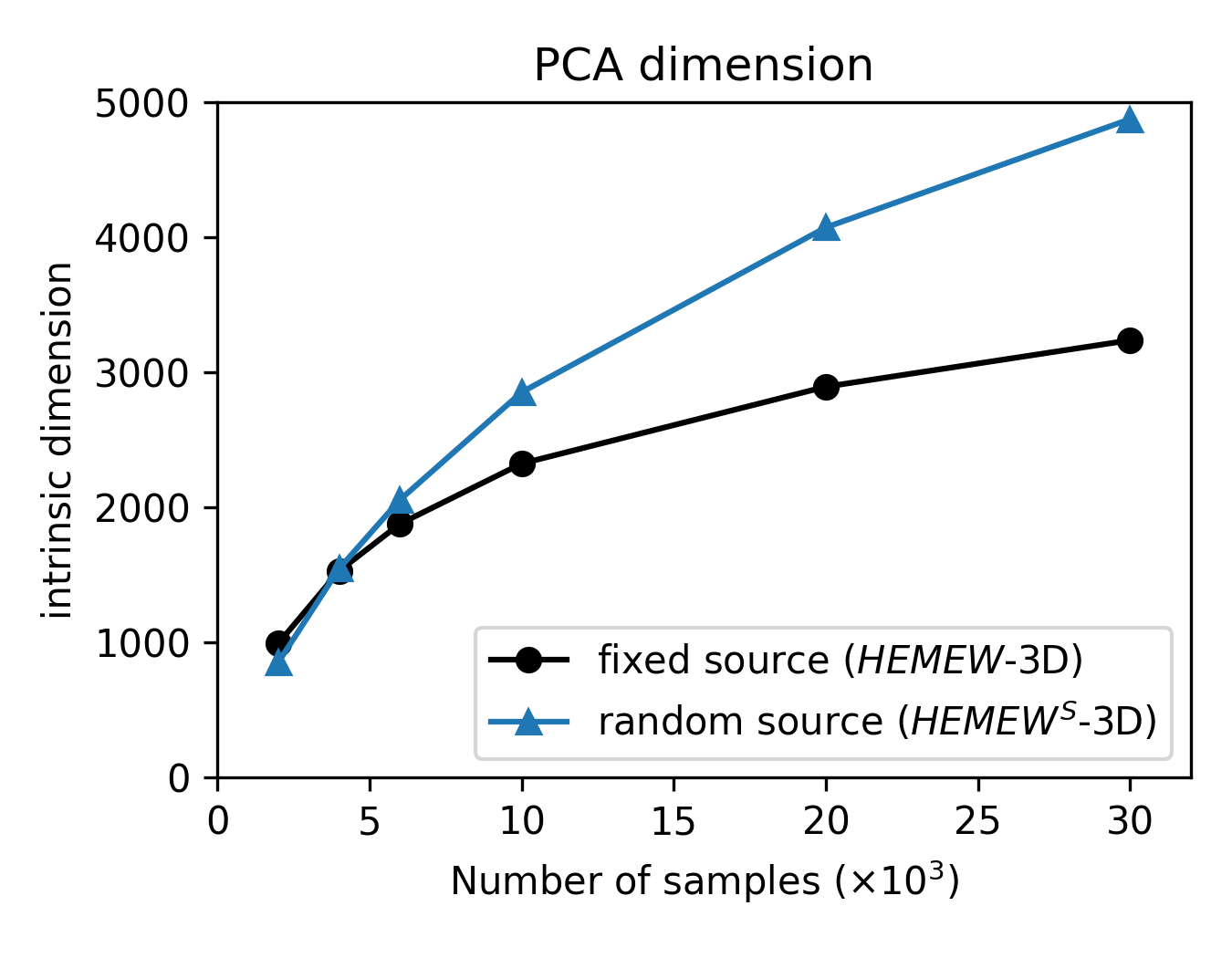}
		\includegraphics[width=0.48\textwidth]{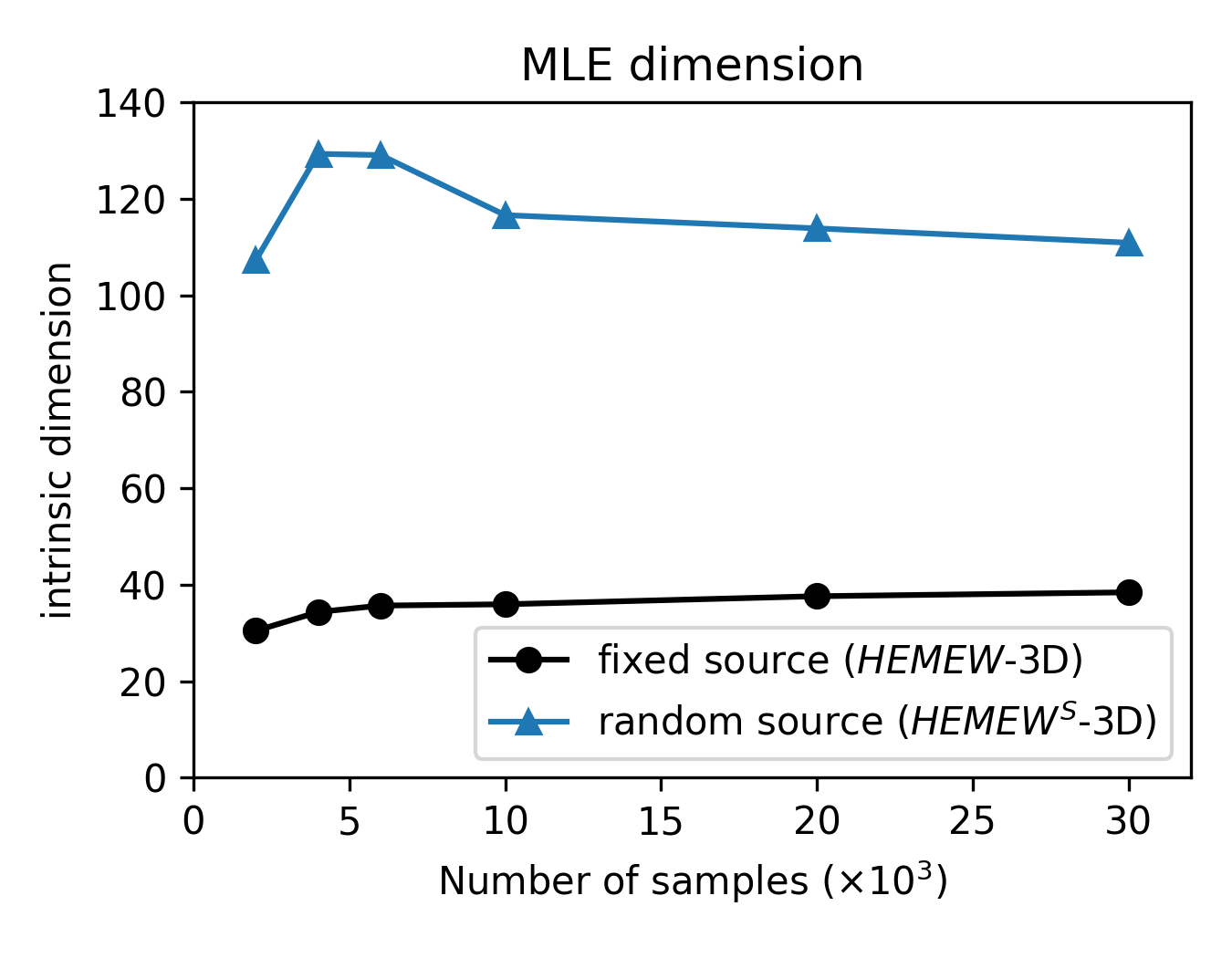}
		\caption{Intrinsic dimension estimated with a linear Principal Component Analysis (PCA, left) and a non-linear Maximum Likelihood Estimator (MLE, right). For each method, the intrinsic dimension is estimated on the HEMEW-3D database where the source is fixed and the HEMEW\textsuperscript{S}-3D database with a random source used to train the MIFNO}
		\label{fig:intrinsic_dimension}
	\end{figure}

	Table \ref{tab:metrics_MIFNO_angle_moment} compares the 8-layer MIFNO taking as input the 6 components of the moment tensor ($M_{xx}$, $M_{yy}$, $M_{zz}$, $M_{xy}$, $M_{xz}$, $M_{yz}$) or the equivalent representation with 3 angles (strike, dip, rake).

	\begin{table}[h]
		\footnotesize
		\begin{center}
			\begin{tabular}{|c|c|c|c|c|c|c|c|}
				\hline
				Model & rMAE & rRMSE & rFFT\textsubscript{low} & rFFT\textsubscript{mid} & rFFT\textsubscript{high} & EG & PG \\
				\hline
				angle & 0.14 $\pm$ 0.05 & 0.21 $\pm$ 0.06 & -0.33 $\pm$ 0.18 & -0.43 $\pm$ 0.19 & -0.51 $\pm$ 0.2 & 6.88 $\pm$ 0.75 & 8.39 $\pm$ 0.58 \\
				moment & 0.14 $\pm$ 0.05 & 0.21 $\pm$ 0.06 & -0.3 $\pm$ 0.18 & -0.41 $\pm$ 0.19 & -0.5 $\pm$ 0.21 & 6.98 $\pm$ 0.77 & 8.46 $\pm$ 0.58 \\
				\hline
			\end{tabular}
		\end{center}
		\caption{Mean and standard deviation of the metrics computed on 1000 test samples. rRMSE: relative RMSE (0 is best), rFFT\textsubscript{low}: relative frequency bias 0-1Hz (0 is best), rFFT\textsubscript{mid}: relative frequency bias 1-2Hz (0 is best), rFFT\textsubscript{high}: relative frequency bias 2-5Hz (0 is best), EG: envelope Goodness-of-Fit (10 is best), PG: phase Goodness-of-Fit (10 is best). For frequency biases, negative values indicate underestimation.}
		\label{tab:metrics_MIFNO_angle_moment}
	\end{table}

\clearpage
	
\subsection{Supplementary material for comparison with baseline models}
	Table \ref{tab:metrics_FNO_fixedsource} compares a 16-layer F-FNO with a 16-layer MIFNO predicting surface velocity wavefields when the source has a fixed position and orientation. It should be noted that the MIFNO was not specifically trained on this database.
	
	\begin{table}[h]
		\begin{center}
			Dataset with a fixed source position and fixed source orientation
			\begin{tabular}{|c|c|c|c|c|c|c|}
				\hline
				Model & rRMSE & rFFT\textsubscript{low} & rFFT\textsubscript{mid} & rFFT\textsubscript{high} & EG & PG \\
				\hline
				F-FNO & 0.14 ; 0.25 & -0.30 ; 0.03 & -0.39 ; 0.01 & -0.44 ; 0.01 & 7.09 ; 8.37 & 8.49 ; 9.32 \\
				MIFNO & 0.20 ; 0.31 & -0.40 ; -0.03 & -0.51 ; -0.06 & -0.56 ; -0.08 & 6.36 ; 7.65 & 7.71 ; 8.83 \\
				%MIFNO + TL & 0.18 ; 0.28 & -0.40 ; -0.02 & -0.52 ; -0.08 & -0.59 ; -0.11 & nan ; nan & nan ; nan \\ 
				%MIFNO + $N_{TL}=500$ & 0.17 ; 0.27 & -0.39 ; -0.05 & -0.50 ; -0.10 & -0.56 ; -0.11 & nan ; nan & nan ; nan \\
				\hline
			\end{tabular}
		\end{center}
		\caption{1st and 3rd quartiles of the metrics computed on 1000 validation samples. rRMSE: relative RMSE (0 is best), rFFT\textsubscript{low}: relative frequency bias 0-1Hz (0 is best), rFFT\textsubscript{mid}: relative frequency bias 1-2Hz (0 is best), rFFT\textsubscript{high}: relative frequency bias 2-5Hz (0 is best), EG: envelope Goodness-of-Fit (10 is best), PG: phase Goodness-of-Fit (10 is best). For frequency biases, negative values indicate underestimation.}
		\label{tab:metrics_FNO_fixedsource}
	\end{table}
	
\begin{figure}[h]
		\centering
		\includegraphics[width=0.75\textwidth]{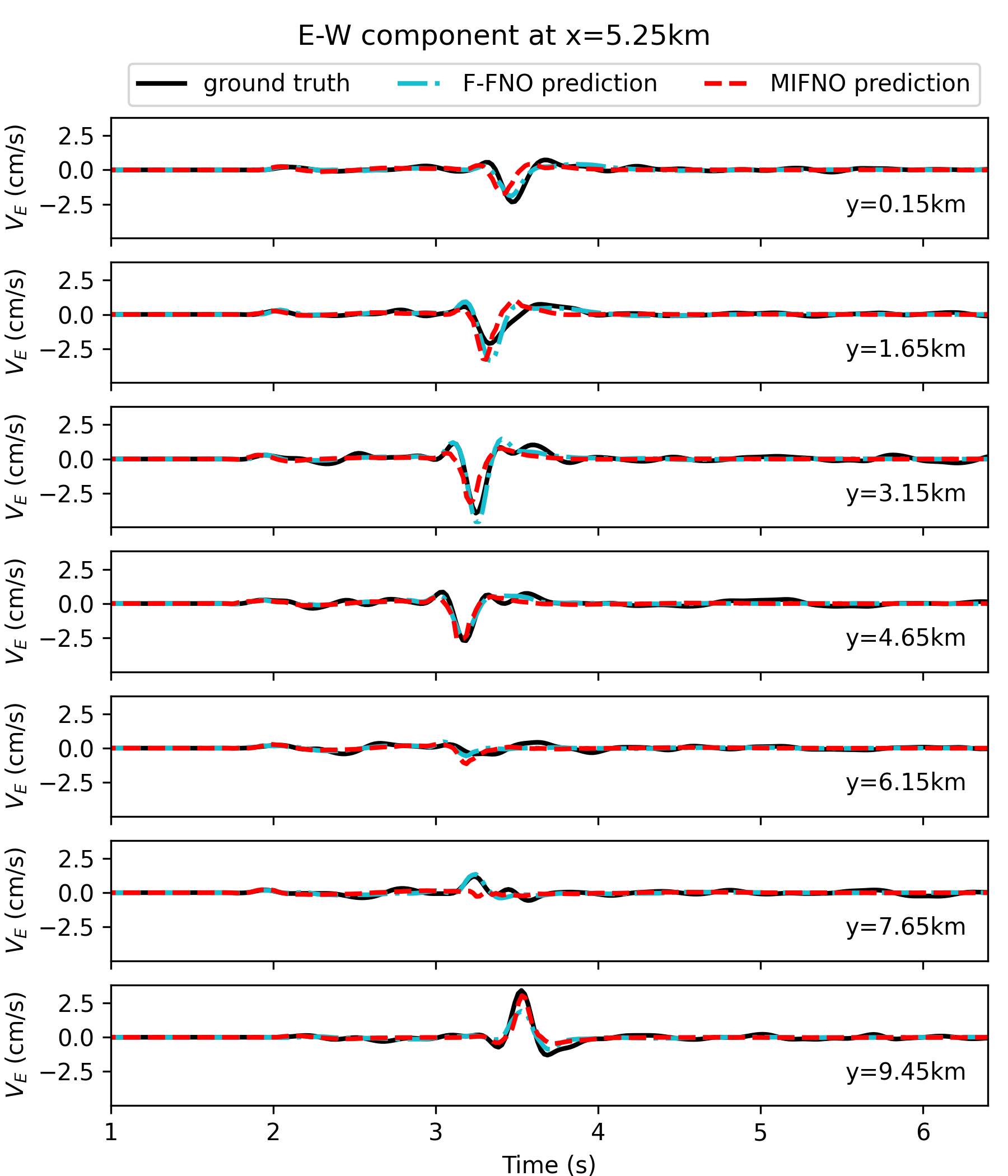}
		\caption{East-West component of ground motion from simulations (black solid line), F-FNO predictions (blue dashed line), and MIFNO predictions (red dashed line)}
		\label{fig:line_timeseries_MIFNO_vs_FNO}
	\end{figure}

	Figure \ref{fig:line_timeseries_MIFNO_vs_FNO} illustrates the F-FNO and MIFNO predictions when the source has a fixed position and orientation. Envelope GOFs of the F-FNO are between \SI{8.3}{} and \SI{8.7}{}, MIFNO ones are between \SI{7.5}{} and \SI{8.7}{}. Phase GOFs of the F-FNO are between \SI{9.2}{} and \SI{9.6}{}, MIFNO ones between \SI{8.2}{} and \SI{9.4}{}. 
	
	\begin{table}[h]
		\begin{center}
			Dataset with a random source position and fixed source orientation
			\begin{tabular}{|c|c|c|c|c|c|c|}
				\hline
				Model & rRMSE & rFFT\textsubscript{low} & rFFT\textsubscript{mid} & rFFT\textsubscript{high} & EG & PG \\
				\hline
				F-FNO 8 layers & 0.14 ; 0.24 & -0.31 ; 0.01 & -0.41 ; -0.03 & -0.48 ; -0.04 & 6.71 ; 8.05 & 8.26 ; 9.19 \\
				MIFNO 8 layers & 0.13 ; 0.23 & -0.27 ; 0.05 & -0.37 ; -0.00 & -0.43 ; -0.00 & 6.93 ; 8.24 & 8.40 ; 9.26 \\
				\hline
			\end{tabular}
		\end{center}
		\caption{1st and 3rd quartiles of the metrics computed on 1000 validation samples. rRMSE: relative RMSE (0 is best), rFFT\textsubscript{low}: relative frequency bias 0-1Hz (0 is best), rFFT\textsubscript{mid}: relative frequency bias 1-2Hz (0 is best), rFFT\textsubscript{high}: relative frequency bias 2-5Hz (0 is best), EG: envelope Goodness-of-Fit (10 is best), PG: phase Goodness-of-Fit (10 is best). For frequency biases, negative values indicate underestimation. Both models were trained with \SI{20000}{} samples for 300 epochs.}
		\label{tab:metrics_FNO_movingsource_8layers}
	\end{table}		
	
\clearpage
	
\subsection{Supplementary material for influence of the source parameters}
	
	\begin{figure}[h]
		\centering
		\includegraphics[width=0.8\textwidth]{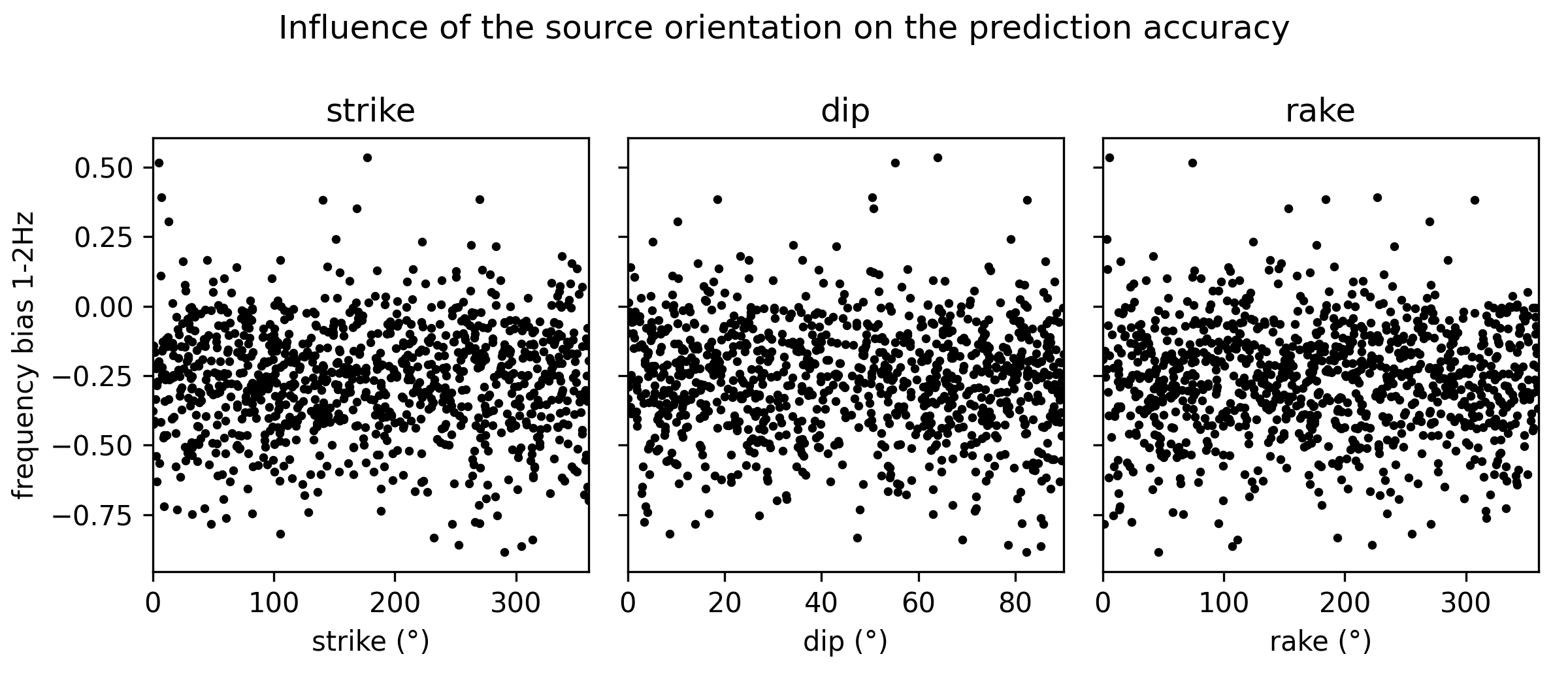}
		\caption{For 1000 samples, the 1-2 Hz frequency bias is shown against the source orientation (strike, dip, rake).}
		\label{fig:error_vs_source_orientation}
	\end{figure}
	
	\begin{figure}[h]
		\centering
		\includegraphics[width=0.7\textwidth]{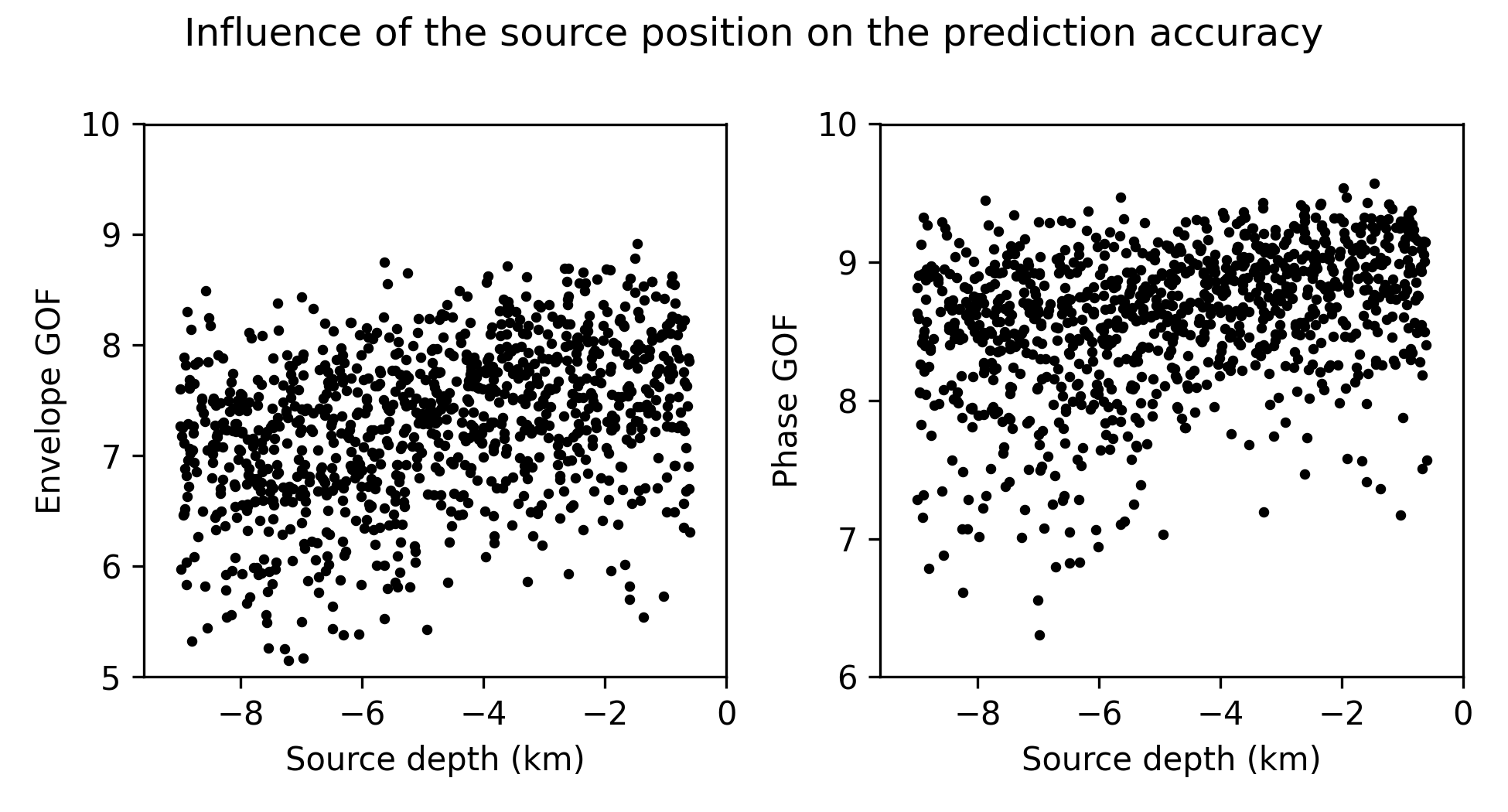}
		\caption{For 1000 samples, the envelope and phase GOF of MIFNO predictions is shown against the source depth.}
		\label{fig:GOF_vs_source_depth}
	\end{figure}

\clearpage

\subsection{Supplementary material for out-of-distribution data}

	\begin{figure}[h]
		\centering
		\includegraphics[width=0.65\textwidth]{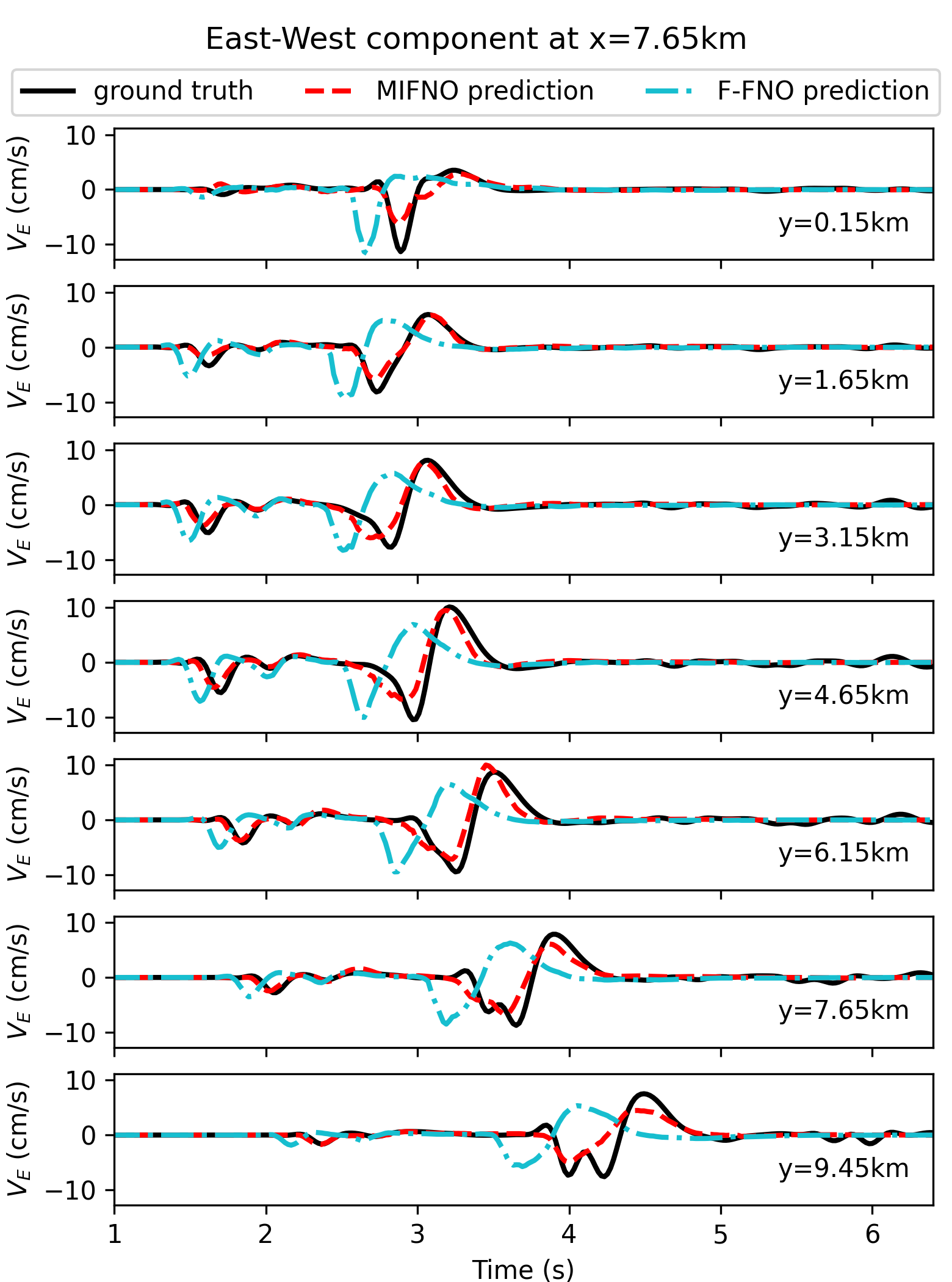}
		\caption{East-West component of ground motion from simulations (black solid line), MIFNO predictions (red dashed line), and F-FNO predictions (blue dashed line) for a source located outside the training domain at (5.1km, 12.9km, -2.5km).}
		\label{fig:timeseries_source_outside}
	\end{figure}

	\begin{figure}[h]
		\centering
		\includegraphics[width=\textwidth]{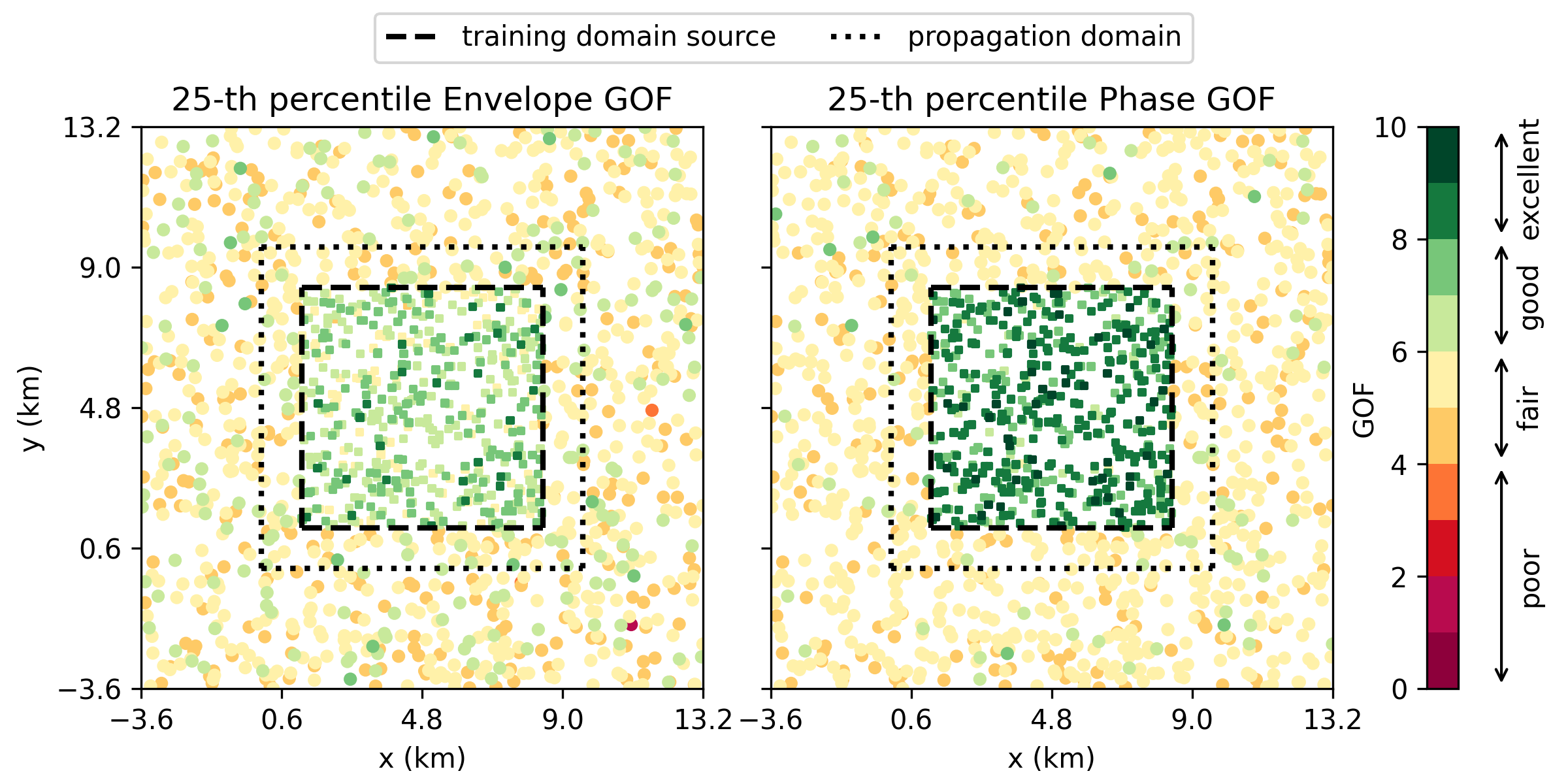}
		\caption{25-th percentile of envelope GOF (left) and phase GOF (right) obtained with the F-FNO for 1000 samples where the source is located outside the training domain (dashed square) or even outside the physical domain (dotted square). For reference, 1000 test samples with a source inside the training domain are shown.}
		\label{fig:GOF_source_outside_FFNO}
	\end{figure}
	
	\begin{figure}[h]
		\centering
		\includegraphics[width=0.85\textwidth]{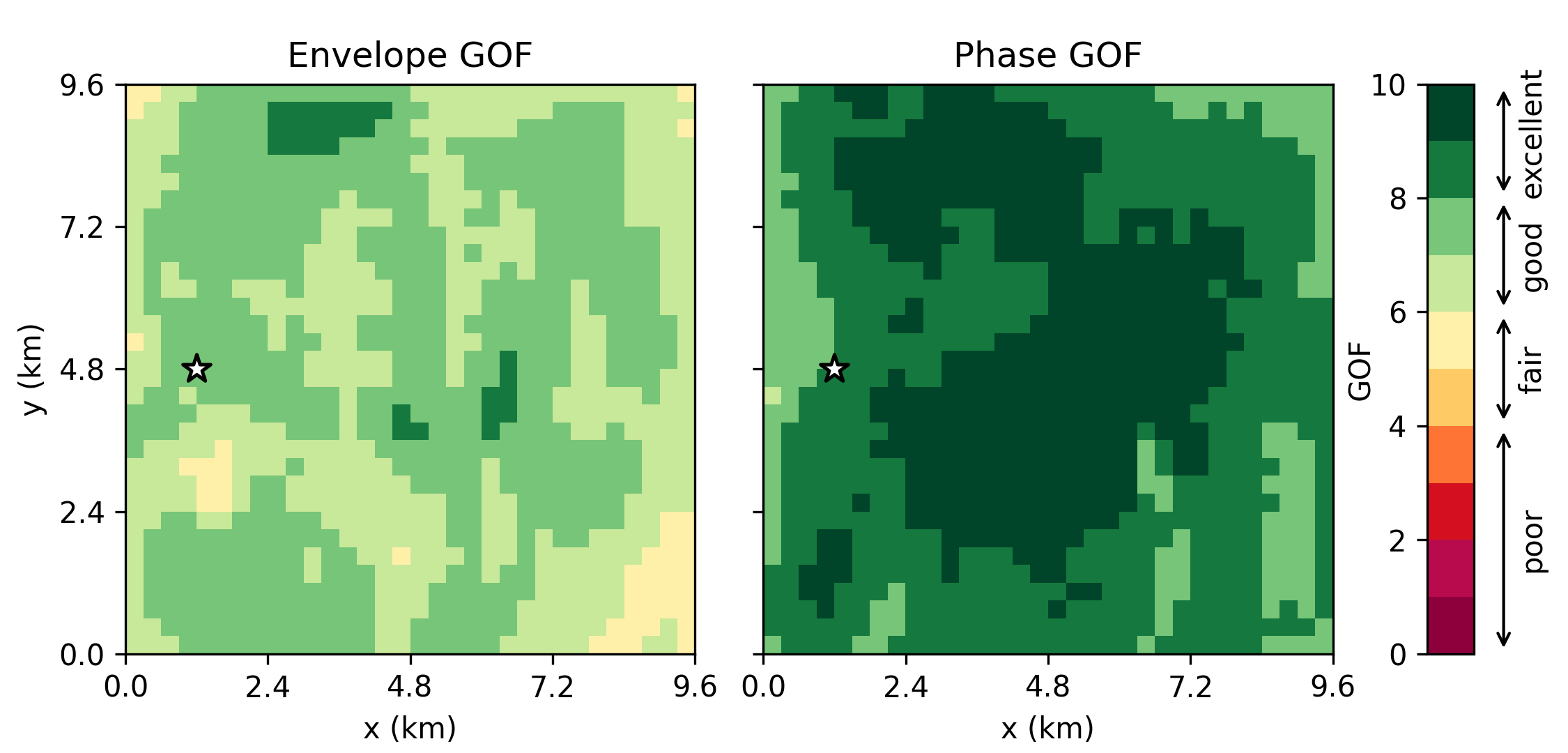}
		\caption{Envelope and phase GOF for the geology depicted in Fig. \ref{fig:overthrust}. The white star denotes the epicenter.}
		\label{fig:gof_overthrust}
	\end{figure}

	\begin{figure}[h]
		\centering
		\includegraphics[width=0.75\textwidth]{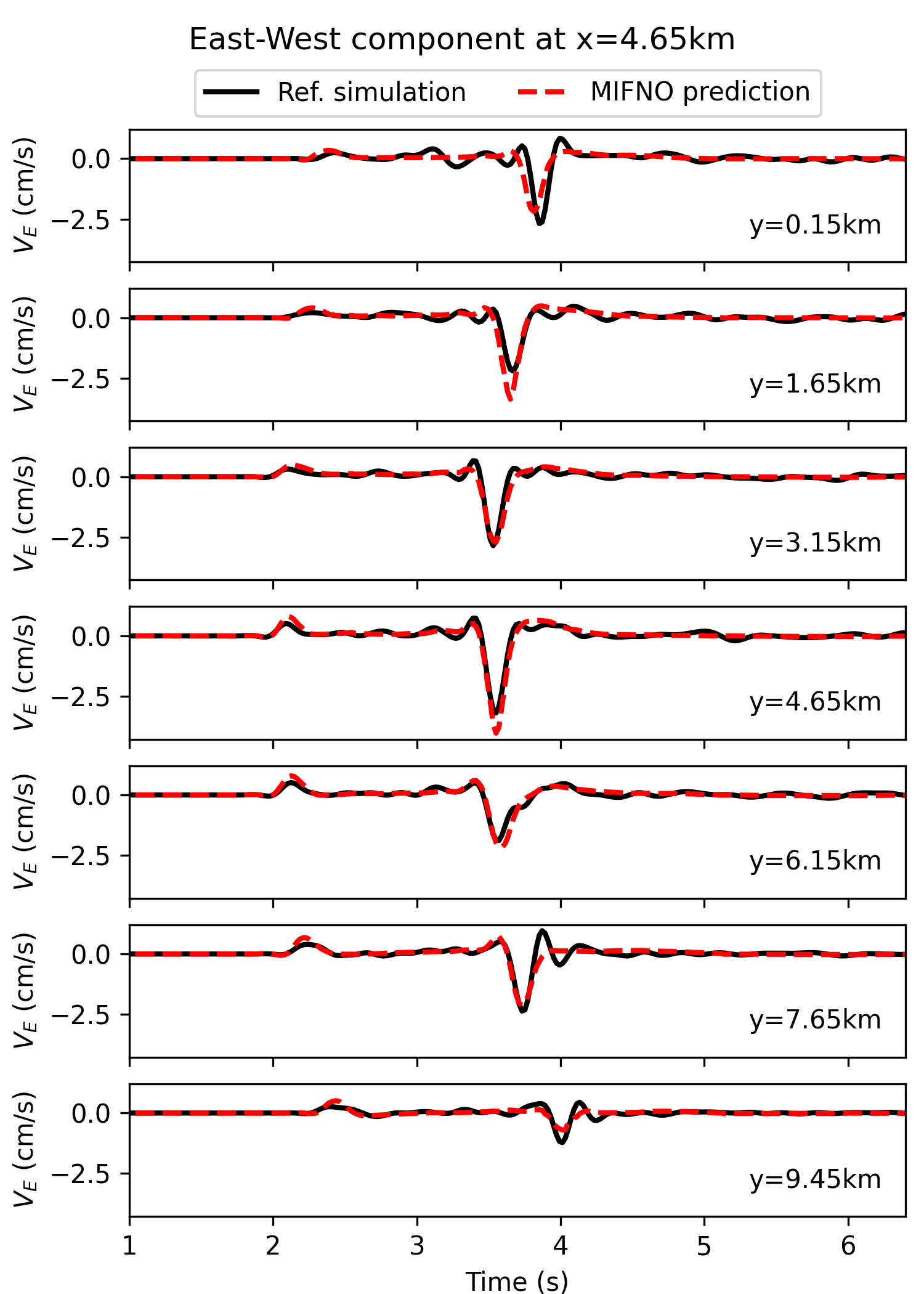}
		\caption{East-West component of ground motion from simulations (black solid line) and MIFNO predictions (red dashed line) for the overthrust geology depicted in Fig. \ref{fig:overthrust}}
		\label{fig:line_timeseries_overthrust}
	\end{figure}

	\begin{figure}[h]
		\centering
		\includegraphics[width=0.75\textwidth]{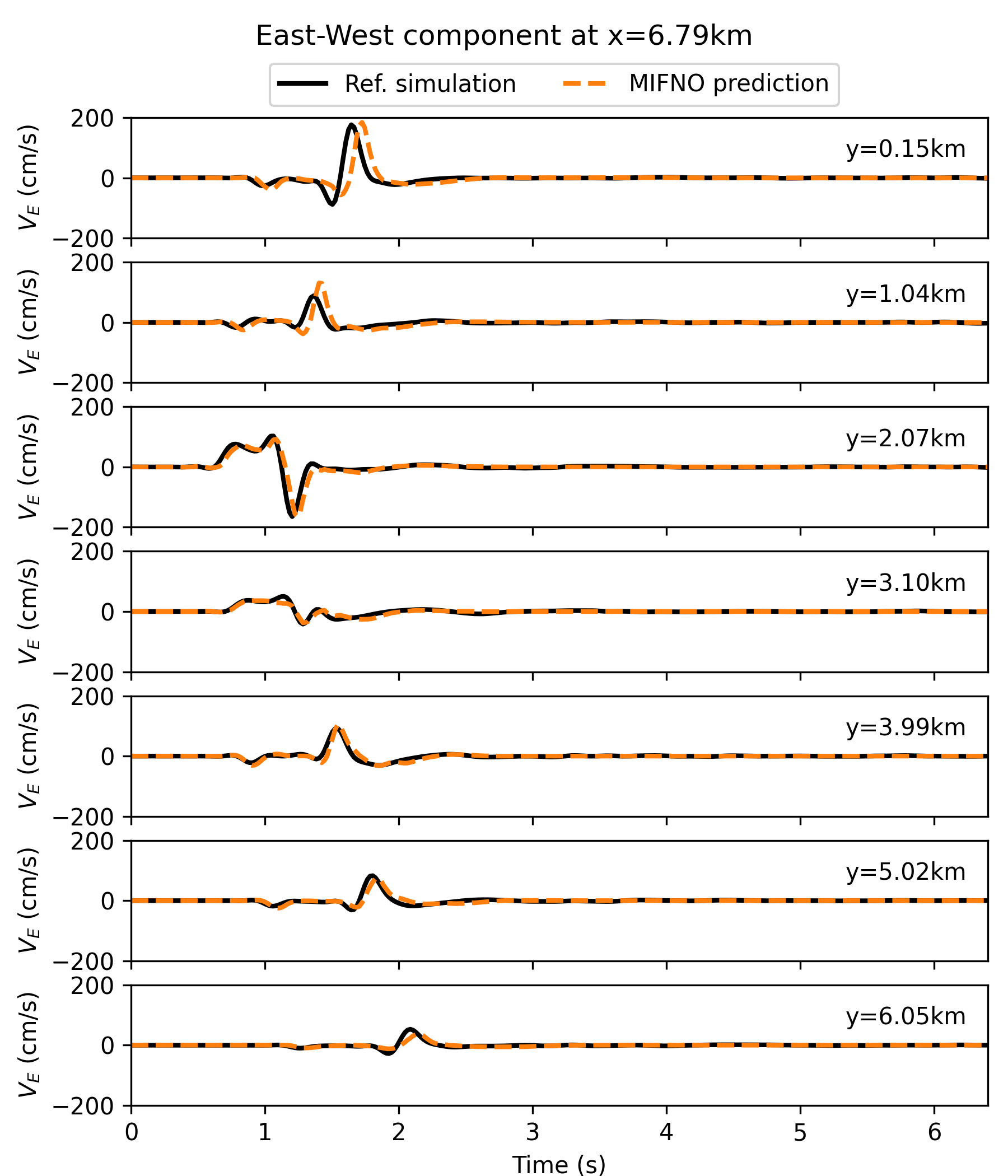}
		\caption{East-West component of ground motion from simulations (black solid line) and MIFNO predictions with resolution 64 (orange dashed line)}
		\label{fig:line_timeseries_res64}
	\end{figure}

\clearpage
\section{Supplementary material for transfer learning}
	\begin{figure}[h]
		\begin{minipage}[b]{.35\textwidth}
			\centering
			\includegraphics[width=0.8\textwidth]{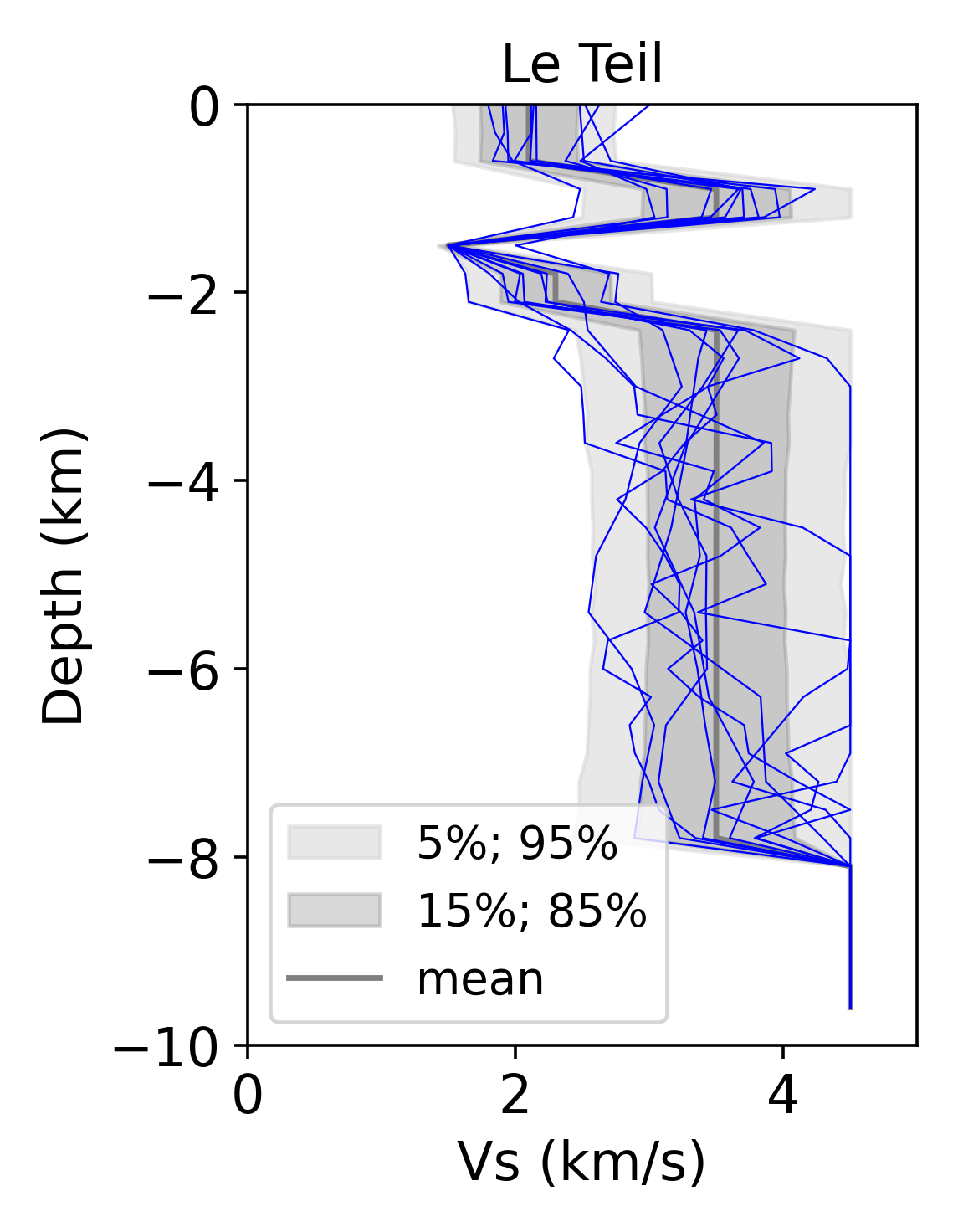}
			\captionof{figure}{V\textsubscript{S} distribution in the Le Teil database. For each of the 32 vertical levels, the mean and percentiles are computed over all horizontal points and samples (shaded areas). Blue lines show some individual vertical profiles.}
			\label{fig:Vs_profile}
		\end{minipage}
		\hfill
		\begin{minipage}[b]{.62\textwidth}
			\centering
			\begin{tabular}{|l|lll|}
				\hline 
				Thickness (m) & V\textsubscript{S} (m/s) & V\textsubscript{P} (m/s) & $\rho$ (kg/m\textsuperscript{3})\\
				\hline 
				600 & 2100 & 3570 & 2329 \\ 
				600 & 3500 & 5950 & 2706 \\ 
				300 & 1200 & 2040 & 1923 \\ 
				600 & 2300 & 3910 & 2380 \\ 
				5700 & 3500 & 5950 & 2706 \\ 
				1800 & 4500 & 7650 & 3170 \\ 
				\hline
			\end{tabular}
			\captionof{table}{Reference 1D geological model for the LeTeil region. Each layer from top to bottom is described by its thickness, S-wave velocity (V\textsubscript{S}), P-wave velocity (V\textsubscript{P}), and density $\rho$ (adapted from \cite{smerziniRegionalPhysicsbasedSimulation2023}).}
			\label{tab:velocity_LeTeil}
		\end{minipage}
	\end{figure}

	\begin{figure}[h]
		\centering
		\includegraphics[width=0.8\textwidth]{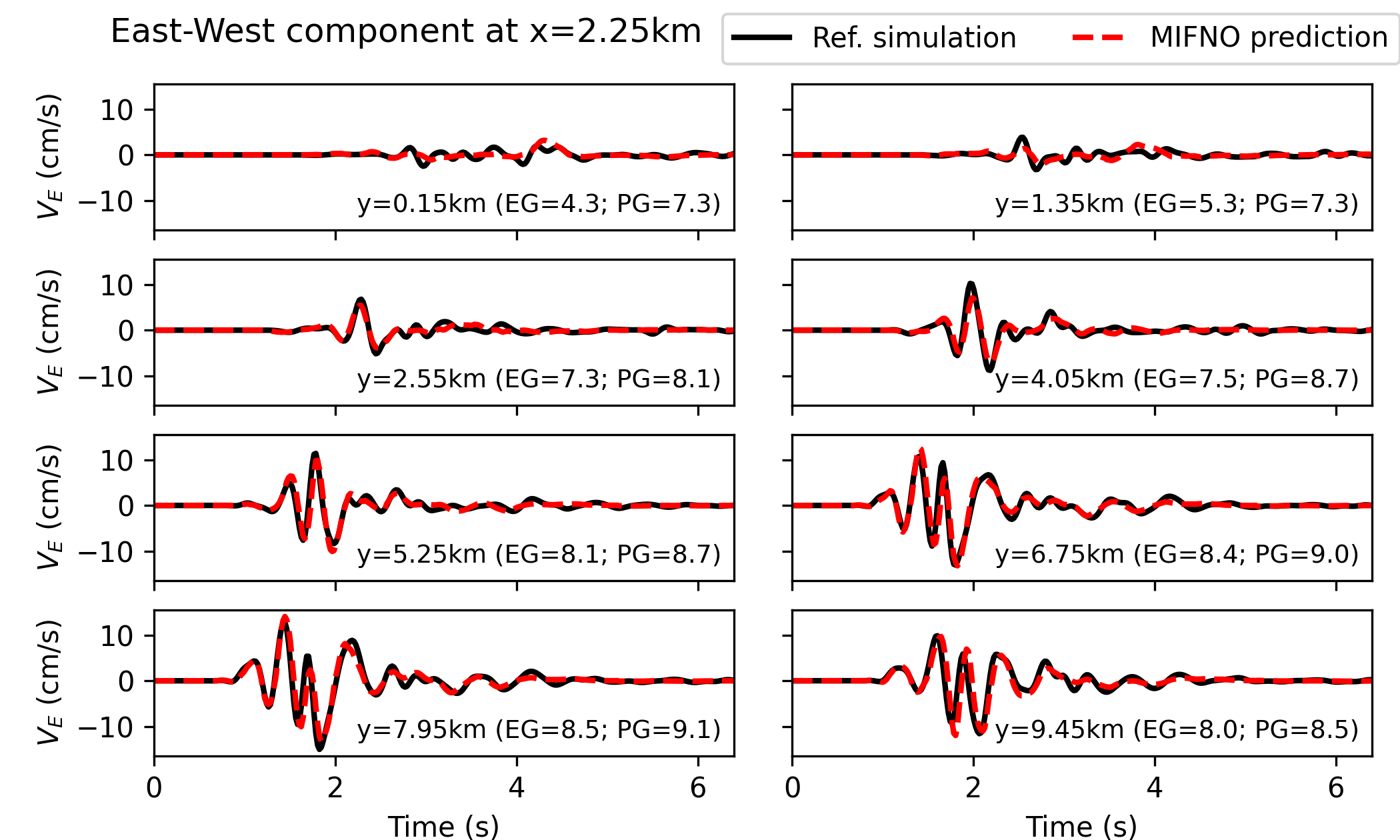}
		\caption{Same as Figure \ref{fig:LeTeil_timeseries} but only 500 transfer learning samples}
		\label{fig:LeTeil_timeseries_TL500}
	\end{figure}

\clearpage
\section*{Data availability statement}
	The HEMEW\textsuperscript{S}-3D database is available at \url{https://doi.org/10.57745/LAI6YU}. The code to train and evaluate the MIFNO can be found at \url{https://github.com/lehmannfa/MIFNO}.

\section*{Disclosure of interest}
	The authors have no conflict of interest to declare.

\clearpage

\bibliographystyle{elsarticle-harv} 
\bibliography{biblio.bib}

\end{document}